\documentclass[journal]{IEEEtran}

\usepackage{amsmath,amsfonts}
\usepackage{algorithmic}
\usepackage{algorithm}
\usepackage{array}
\usepackage[caption=false,font=scriptsize,labelfont=sf,textfont=sf]{subfig}
\usepackage{textcomp}
\usepackage{stfloats}
\usepackage{url}
\usepackage{verbatim}
\usepackage{graphicx}
\usepackage[nocompress]{cite}
\usepackage{amsthm}
\usepackage[hidelinks,breaklinks=true]{hyperref}
\usepackage[noabbrev]{cleveref}
\usepackage[most]{tcolorbox}
\usepackage{xcolor}
\usepackage{bbm}
\usepackage{booktabs}
\usepackage{siunitx}
\usepackage{multirow}
\usepackage{float}

\crefname{figure}{Figure}{Figures}
\crefname{definition}{Definition}{Definitions}
\crefname{table}{Table}{Tables}
\crefname{equation}{Equation}{Equations}
\crefname{section}{Section}{Sections}
\crefname{appendix}{Appendix}{Appendices}

\newtheoremstyle{my_definition}
  {}
  {}
  {\itshape}
  {}
  {\bfseries}
  {.}
  { }
  {\thmname{#1} \thmnumber{#2} (\thmnote{#3})}
\theoremstyle{my_definition}
\newtheorem{definition}{Definition}

\newcommand{\multirowoffset}{-0.5\dimexpr \aboverulesep + \belowrulesep + \cmidrulewidth}
\newcolumntype{C}[1]{>{\centering\arraybackslash}p{#1}}

\begin{document}

\title{Capturing Context-Aware Route Choice Semantics for Trajectory Representation Learning}

\author{
  Ji~Cao$^{\ast}$,
  Yu~Wang$^{\ast}$,
  Tongya~Zheng,
  Jie~Song,
  Qinghong~Guo,
  Zujie~Ren$^{\dag}$,\\
  Canghong~Jin,
  Gang~Chen,
  Mingli~Song

  \thanks{This work was supported by the National Key Research and Development Program of China (No. 2019YFB2102101), the Starry Night Science Fund of Zhejiang University Shanghai Institute for Advanced Study (Grant No. SN-ZJU-SIAS-001), the Hangzhou Joint Fund of the Zhejiang Provincial Natural Science Foundation of China under Grant (No. LHZSD24F020001), and the Fundamental Research Funds for the Central Universities (No. 226-2025-00057).}
  \thanks{Ji Cao, Yu Wang, Zujie Ren, Gang Chen, and Mingli Song are with the College of Computer Science and Technology, Zhejiang University, Hangzhou 310027, China; Ji Cao and Zujie Ren are also with the Zhejiang Lab, Hangzhou 311121, China (e-mail: \mbox{caoj25@zju.edu.cn}; \mbox{yu.wang@zju.edu.cn}; \mbox{renzju@zju.edu.cn}; \mbox{cg@zju.edu.cn}; \mbox{brooksong@zju.edu.cn}).}
  \thanks{Tongya Zheng and Canghong Jin are with the Zhejiang Provincial Engineering Research Center for Real-Time SmartTech in Urban Security Governance, Hangzhou City University, Hangzhou 310015, China (e-mail: \mbox{doujiang\_zheng@163.com}; \mbox{jinch@hzcu.edu.cn}).}
  \thanks{Jie Song is with the School of Software Technology, Zhejiang University, Ningbo 315100, China (e-mail: \mbox{sjie@zju.edu.cn}).}
  \thanks{Qinghong Guo is with the Polytechnic Institute of Zhejiang University, Hangzhou 310015, China (e-mail: \mbox{q.h\_guo@zju.edu.cn}).}
  \thanks{$^{\ast}$ Ji Cao and Yu Wang contributed equally to this work.}
  \thanks{$^{\dag}$ Zujie Ren is the corresponding author.}
}

\maketitle

\begin{abstract}
Trajectory representation learning (TRL) aims to encode raw trajectory data into low-dimensional embeddings for downstream tasks such as travel time estimation, mobility prediction, and trajectory similarity analysis. From a behavioral perspective, a trajectory reflects a sequence of route choices within an urban environment. However, most existing TRL methods ignore this underlying decision-making process and instead treat trajectories as static, passive spatiotemporal sequences, thereby limiting the semantic richness of the learned representations. To bridge this gap, we propose \textbf{CORE}, a TRL framework that integrates context-aware route choice semantics into trajectory embeddings. CORE first incorporates a multi-granular Environment Perception Module, which leverages large language models (LLMs) to distill environmental semantics from point of interest (POI) distributions, thereby constructing a context-enriched road network. Building upon this backbone, CORE employs a Route Choice Encoder with a mixture-of-experts (MoE) architecture, which captures route choice patterns by jointly leveraging the context-enriched road network and navigational factors. Finally, a Transformer encoder aggregates the route-choice-aware representations into a global trajectory embedding. Extensive experiments on 4 real-world datasets across 6 downstream tasks demonstrate that CORE consistently outperforms 15 state-of-the-art TRL methods, achieving an average improvement of 9.20\% over the best-performing baseline. Our code is available at \url{https://github.com/caoji2001/CORE}.

\end{abstract}


\section{Introduction}

\IEEEPARstart{T}{he} proliferation of GPS-enabled devices facilitates the large-scale collection of spatiotemporal trajectories, which record detailed movement patterns of vehicles~\cite{zheng2015trajectory, wang2021survey, wang2022deep}. These trajectories encode rich spatiotemporal semantics that support a wide range of downstream applications, including travel time estimation~\cite{james2023citywide}, mobility prediction~\cite{zhang2023beyond}, and trajectory similarity analysis~\cite{hu2024spatio}.

\emph{Trajectory representation learning} (TRL), which encodes raw trajectory data into low-dimensional embeddings, receives considerable attention for its ability to extract informative spatiotemporal patterns~\cite{mao2022jointly, yang2023lightpath, jiang2023self}. Early TRL methods~\cite{fu2020trembr, yang2021unsupervised} primarily employ LSTMs~\cite{hochreiter1997long} or GRUs~\cite{cho2014learning} as sequence backbones. For example, Trembr~\cite{fu2020trembr} uses an RNN-based encoder-decoder to reconstruct trajectories, thereby capturing their spatiotemporal characteristics. Motivated by the strong sequence-modeling capability of the Transformer architecture~\cite{vaswani2017attention}, recent studies~\cite{chen2021robust, mao2022jointly, jiang2023self} increasingly adopt it as the backbone. For instance, START~\cite{jiang2023self} integrates a graph attention network (GAT)~\cite{velivckovic2018graph} with a time-aware Transformer encoder to capture travel semantics and temporal regularities, whereas GREEN~\cite{zhou2025grid} leverages a Transformer architecture to jointly model road- and grid-level trajectories. Despite these advances, most existing TRL methods still treat trajectories as static, passive spatiotemporal sequences, neglecting the underlying behavioral mechanisms that govern trajectory formation.

\begin{figure}[t]
    \centering
    \includegraphics[width=1.0\linewidth]{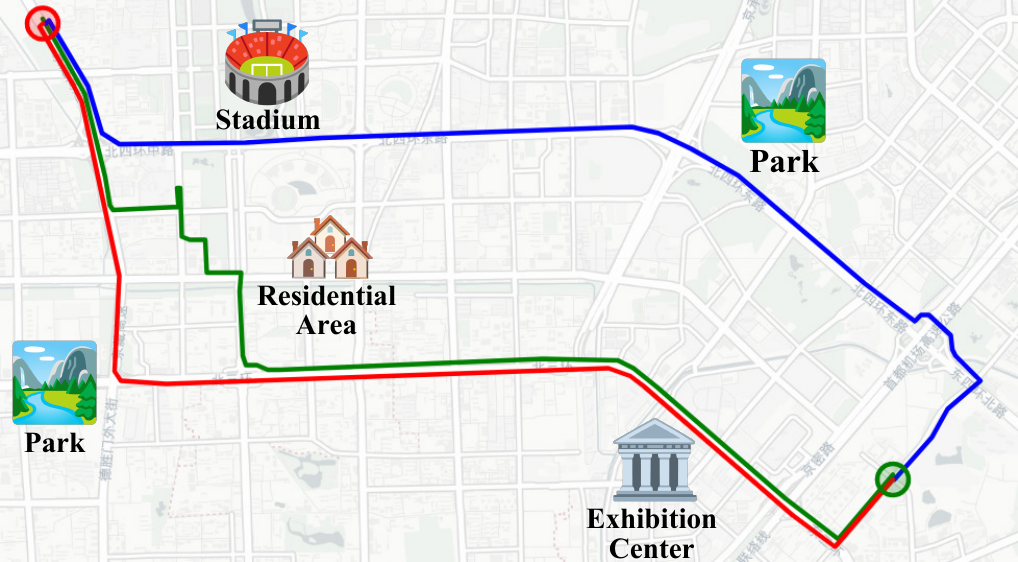}
    \caption{Even with the same OD pair, different drivers may choose substantially different routes, reflecting complex route choice behaviors.}
    \label{fig:intro_route_choice_behavior}
\end{figure}

In reality, a trajectory is the outcome of a sequence of route choices, which are intrinsically contextualized by the surrounding urban environment~\cite{prato2009route, lima2016understanding}. As illustrated in \cref{fig:intro_route_choice_behavior}, three trajectories share the same origin--destination (OD) pair: the blue trajectory follows urban expressways with light traffic (passing a park and a stadium), like prioritizing efficiency and minimal congestion; the red trajectory also uses expressways but experiences moderate traffic (passing an exhibition center and a park), reflecting a trade-off between travel time and the views along the route; in contrast, the green trajectory partially diverts into a heavily congested residential area, indicating a stopover or a preference for familiar roads. These divergences result from a sequence of latent route choices within the urban context. Consequently, modeling trajectories solely as static, passive spatiotemporal sequences, while ignoring the underlying route choices, produces representations with limited behavioral fidelity, which in turn hinders performance on downstream tasks.

However, capturing route choice semantics from trajectory data involves two closely interrelated challenges: (1) \textbf{External Urban Context Perception:} Route choices essentially manifest as sequences of decisions within an urban environment. Consequently, accurately characterizing such behavioral patterns necessitates a precise perception of the environmental context. (2) \textbf{Internal Route Choice Complexity:} Route choice logic varies substantially across different intersection contexts, which makes it non-trivial for a unified model to capture such heterogeneous behavioral patterns.

To address these challenges, we propose \textbf{CORE}, a framework for capturing \textbf{CO}ntext-aware \textbf{R}oute choic\textbf{E} semantics for effective trajectory representation learning. CORE treats trajectories as sequences of context-aware route choices and consists of two key components. First, to achieve context awareness, we develop a multi-granular Environment Perception Module. This module leverages the world knowledge encoded in large language models (LLMs) to distill environmental semantics from point of interest (POI) distributions at both road-segment and urban-functional levels, yielding a context-aware road network that serves as a semantic backbone. Second, to capture the latent route choices governing observed trajectories, we introduce a Route Choice Encoder. It integrates the aforementioned context-aware road network with navigational factors, and employs a mixture-of-experts (MoE)~\cite{dai2024deepseekmoe} architecture to capture heterogeneous routing semantics across diverse intersection contexts. A Transformer encoder then aggregates the resulting route-choice-aware representations into a global trajectory embedding. The entire framework is end-to-end trainable and utilizes a contrastive learning objective~\cite{chen2020simple} for pretraining.

To evaluate CORE, we conduct extensive experiments on 4 real-world trajectory datasets across 6 downstream tasks. The results demonstrate that CORE consistently outperforms 15 \emph{state-of-the-art} TRL methods, achieving an average performance improvement of 9.20\% over the best-performing baseline. In addition, an ablation study shows that each module of CORE contributes to performance, while further analysis confirms that CORE exhibits superior data efficiency in few-shot scenarios and competitive computational efficiency.

To summarize, we make the following three contributions:
\begin{itemize}
  \item We identify a critical oversight in existing TRL methods: trajectories are treated as simple sequences of road segments, rather than as sequences of route choices within an urban environment, and this fundamentally limits these models' ability to capture the intrinsic decision-making semantics embedded in trajectories.
  \item We propose CORE as a more effective TRL framework. CORE learns trajectory embeddings by incorporating context-aware route choice semantics as an inductive bias, using a multi-granular Environment Perception Module, an MoE-based Route Choice Encoder, and a Transformer-based aggregation layer.
  \item We conduct extensive experiments on 4 real-world trajectory datasets to evaluate CORE, and compare it with 15 \emph{state-of-the-art} baselines. The results show that CORE consistently outperforms existing TRL methods.
\end{itemize}

\section{Related Work}

\subsection{Trajectory Representation Learning}

TRL aims to map raw trajectory data into low-dimensional embeddings that can be leveraged by a wide range of spatiotemporal analysis tasks. Early TRL methods predominantly learn task-specific trajectory embeddings. For example, t2vec~\cite{li2018deep}, NEUTRAJ~\cite{yao2019computing}, and GTS~\cite{han2021graph} are specifically designed for trajectory similarity computation, whereas DeepTTE~\cite{wang2018will}, WDR~\cite{wang2018learning}, and ConSTGAT~\cite{fang2020constgat} primarily focus on travel time estimation. However, such task-specific representations often exhibit limited generalization ability across downstream tasks. To overcome this limitation, subsequent research shifts toward general-purpose pretraining, aiming to learn versatile, task-agnostic trajectory representations that are effective across a wide range of downstream applications. For instance, Trembr~\cite{fu2020trembr} and RED~\cite{zhou2025red} employ sequence-to-sequence models~\cite{cho2014learning} to reconstruct trajectories, while JGRM~\cite{ma2024more} adopts a masked language modeling (MLM) objective~\cite{devlin2019bert} to capture intrinsic spatiotemporal characteristics within trajectories. Inspired by the success of contrastive learning in computer vision~\cite{chen2020simple}, a series of subsequent studies introduce contrastive objectives into TRL: PIM~\cite{yang2021unsupervised} maximizes mutual information between global and local views of trajectories, JCLRNT~\cite{mao2022jointly} aligns trajectory and road network features through cross-scale contrastive learning, and START~\cite{jiang2023self} performs contrastive learning across trajectories to capture travel semantics and temporal regularities. Recent multimodal TRL methods further incorporate external semantics from grids, traffic states, POIs, road attributes, and mobility patterns~\cite{zhou2025grid, han2025bridging, zhou2025trajcogn, liu2025trajmamba, wei2025path}. Although these methods enrich trajectory embeddings with contextual signals, they still primarily learn representations of observed spatiotemporal sequences, largely overlooking the underlying route choices that reflect travelers' decision-making processes.

\subsection{Road Network Representation Learning}

Road network representation learning aims to learn latent representations of road segments or intersections in an urban network. Early methods such as IRN2Vec~\cite{wang2019learning} and SRN2Vec~\cite{wang2020representation} extend node2vec~\cite{grover2016node2vec} by incorporating spatial locality and mobility patterns to obtain road network representations. Building on this line of work, Toast~\cite{chen2021robust} jointly models traffic patterns and mobility semantics to enhance embedding robustness, while DyToast~\cite{chen2025semantic} further introduces a unified temporal encoding to learn time-aware representations. Given the strong capability of graph neural networks (GNNs)~\cite{wu2020comprehensive} in modeling non-Euclidean structures, subsequent studies primarily adopt GNN-based architectures. For example, RFN~\cite{jepsen2020relational} utilizes a graph convolutional network (GCN)~\cite{kipf2017semi} to model road segments and intersections jointly. HRNR~\cite{wu2020learning} constructs a three-level hierarchical GNN to learn road network embeddings that integrate both structural and functional properties, while HyperRoad~\cite{zhang2023road} and DST~\cite{guo2026dual} leverage hypergraphs~\cite{antelmi2023survey} to capture higher-order relationships among road segments. SARN~\cite{chang2023spatial} further integrates graph contrastive learning (GCL)~\cite{zhu2021graph} to strengthen spatial modeling and improve representation robustness. However, the incorporation of rich urban environmental semantics into road network embeddings remains underexplored. To address this limitation, we leverage LLMs to distill human-centric semantics from multi-granular POI distributions and inject these semantics into road network embeddings, thereby yielding context-aware road network representations.

\subsection{LLMs for Spatiotemporal Modeling}

Driven by the progress of LLMs in natural language processing~\cite{ouyang2022training}, recent work explores their potential in spatiotemporal data mining. In mobility modeling, LLMob~\cite{jiawei2024large}, CoPB~\cite{shao2024chain}, and AgentMove~\cite{feng2025agentmove} formulate mobility generation as an LLM-based planning problem, where agents reason over individual routines and urban context to synthesize daily activities. For POI recommendation, Mobility-LLM~\cite{gong2024mobility}, LLM4POI~\cite{li2024large}, POI-Enhancer~\cite{cheng2025poi}, M3ob~\cite{dai2025learning}, and ALOHA~\cite{wang2026adaptive} convert user check-in sequences and POI attributes into natural language descriptions and, through prompt engineering, inject LLM knowledge into spatiotemporal encodings to better capture user intent and preferences, thereby improving next-POI prediction. In traffic forecasting, PromptST~\cite{zhang2023promptst}, UrbanGPT~\cite{li2024urbangpt}, and UniST~\cite{yuan2024unist} adopt LLM paradigms such as pretraining and instruction tuning to build general-purpose spatiotemporal predictors.

More recently, TrajCogn~\cite{zhou2025trajcogn}, TrajMamba~\cite{liu2025trajmamba}, and Path-LLM~\cite{wei2025path} explore integrating LLMs into TRL by leveraging textual semantics from POIs, road attributes, or mobility patterns. These studies demonstrate the potential of LLM-derived semantic knowledge for trajectory modeling. However, their semantic signals are mainly used as trajectory-level textual augmentations, without explicitly grounding POI-derived environmental semantics into road network representations across multiple spatial granularities. In contrast, CORE uses LLMs to derive multi-granular environmental semantics from POI distributions and injects them into road network embeddings, thereby aligning heterogeneous POI and road network data within a unified, context-aware framework to support subsequent route choice encoding.

\section{Preliminaries}

\subsection{Definitions}

\begin{definition}[GPS Trajectory]
A GPS trajectory is defined as a sequence of points $\mathcal{T}^{\text{GPS}} = \{ \boldsymbol{\tau}^{\text{GPS}}_1, \boldsymbol{\tau}^{\text{GPS}}_2, \ldots, \boldsymbol{\tau}^{\text{GPS}}_n \}$, where each $\boldsymbol{\tau}^{\text{GPS}}_i=(\text{lat}_i, \text{lon}_i, t_i)$ consists of the latitude, longitude, and timestamp, respectively.

\end{definition}

\begin{definition}[Road Network]
A road network is represented as a directed graph $\mathcal{G} = \langle \mathcal{V}, \mathcal{E} \rangle$, where $\mathcal{V}$ is the set of road segments and $\mathcal{E} \subseteq \mathcal{V} \times \mathcal{V}$ denotes directed connectivity. A directed edge $(r,r') \in \mathcal{E}$ indicates that segment $r'$ is directly reachable from segment $r$.
\end{definition}

\begin{definition}[Road Network Constrained Trajectory]
A road network constrained trajectory\footnote{For convenience, ``trajectory'' will refer to a ``road network constrained trajectory'' in unambiguous contexts throughout this paper.} $\mathcal{T} = \{ \boldsymbol{\tau}_1, \boldsymbol{\tau}_2, \ldots, \boldsymbol{\tau}_n \}$ is defined as a timestamped sequence of road segments obtained by map matching a raw GPS trajectory to the underlying road network. Each element $\boldsymbol{\tau}_i=(r_i,t_i)$ consists of the matched road segment $r_i\in\mathcal{V}$ and its timestamp $t_i$.
\end{definition}

\begin{definition}[Point of Interest]
\label{def:poi}
A point of interest (POI) is formalized as a geospatial entity tuple $\boldsymbol{p}$ = (latitude, longitude, category, subcategory, name), where $\text{latitude}$ and $\text{longitude}$ denote the geographic coordinates, $\text{category}$ and $\text{subcategory}$ define a two-level functional taxonomy (e.g., category ``Shopping Services'' and subcategory ``Convenience Stores''), and $\text{name}$ specifies the entity's proper name (e.g., ``7-Eleven'').
\end{definition}

\subsection{Problem Statement}

Given a trajectory dataset $\mathcal{D} = \{ \mathcal{T}_1, \mathcal{T}_2, \ldots, \mathcal{T}_{\lvert \mathcal{D} \rvert} \}$, our objective is to learn an encoding function:
\begin{equation}
F: \mathcal{T} \to \mathbf{z} \in \mathbb{R}^d,
\end{equation}
which maps a trajectory $\mathcal{T}$ to a $d$-dimensional representation $\mathbf{z}$. We expect the mapping function $F(\cdot)$ to exhibit the following two core capabilities: (1) \textbf{Spatiotemporal Pattern Capture:} effective encoding of the intrinsic spatiotemporal patterns of trajectories into a low-dimensional representation space. (2) \textbf{Task Versatility:} flexible adaptation to diverse downstream applications, such as destination prediction, travel time estimation, and similar trajectory retrieval.

\section{Methodology}

\begin{figure*}[!t]
    \centering
    \includegraphics[width=1.0\linewidth]{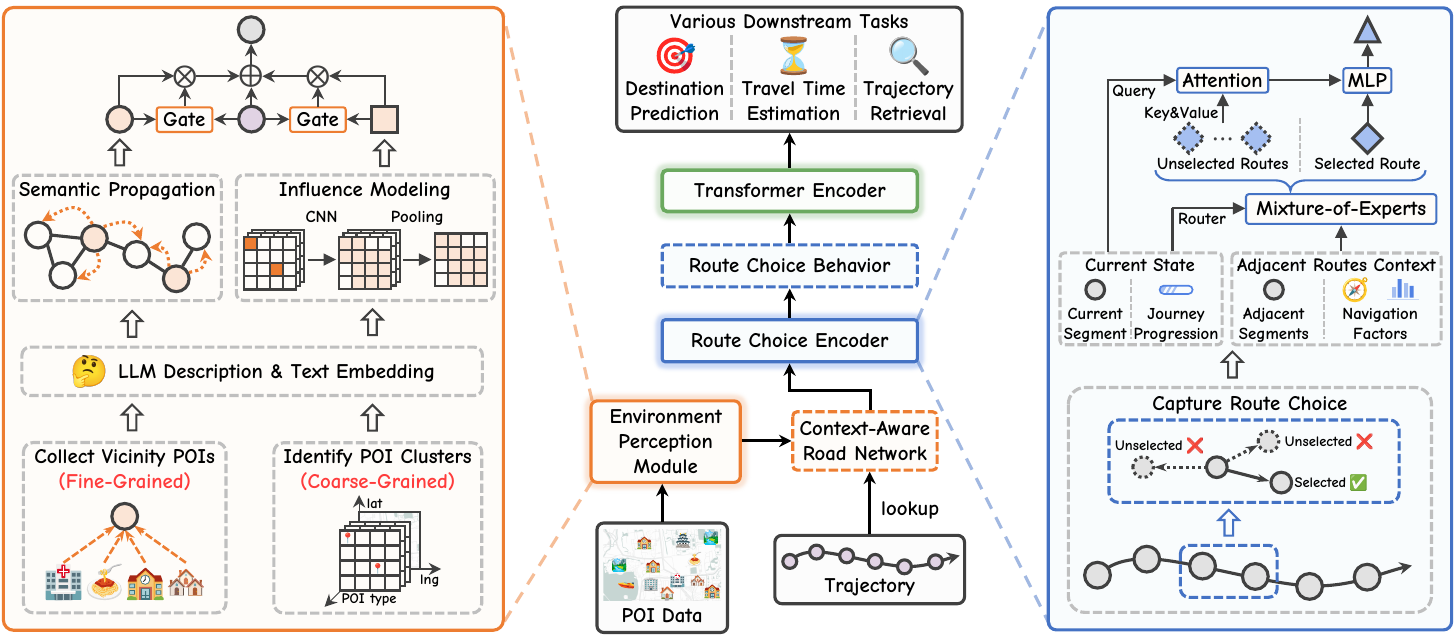}
    \caption{Overview of CORE. The Environment Perception Module first constructs a context-aware road network based on the environmental semantics implied by POI distributions. Building on this backbone, the Route Choice Encoder models route choices by jointly leveraging the context-enriched road network and navigational factors. A Transformer encoder finally aggregates the resulting route-choice-aware representations into global trajectory embeddings.}
    \label{fig:framework}
\end{figure*}

In this section, we introduce CORE, a TRL framework that treats trajectories as observable traces of sequential route choice decisions. As illustrated in \cref{fig:framework}, CORE comprises two key components: (1) an Environment Perception Module, which constructs a context-aware road network from multi-granular POI distributions; and (2) a Route Choice Encoder, which models route choices by integrating the context-aware road network with navigational factors via an MoE architecture. Finally, a Transformer encoder aggregates these sequential behavioral representations into a global trajectory embedding. The entire framework is end-to-end trainable and is pretrained with a contrastive learning objective.

\subsection{Environment Perception Module}

To effectively model route choices, it is essential to capture the underlying environmental semantics that constitute the decision-making context. Although transportation network semantics are jointly shaped by intrinsic network attributes and the surrounding urban context (e.g., land use, commercial vitality)~\cite{hillier1989social}, the latter is often difficult to quantify directly. Because POIs provide a rich characterization of urban functions, we use them as observable proxies for environmental semantics and design a multi-granular modeling scheme: a fine-grained level that characterizes road-segment-level context, and a coarse-grained level that quantifies the influence of the broader urban functional structure. Together, these two levels yield a context-aware road network representation that serves as the backbone for subsequent route choice encoding.

\subsubsection{Fine-Grained Semantic Modeling}

At the fine-grained level, we characterize the local urban context of each road segment by aggregating POIs within a radius $\delta$ (set to $\delta=\SI{100}{m}$ in our experiments). However, bridging the semantic gap between heterogeneous POI attributes and latent urban functions remains non-trivial; naive aggregation of POI categories fails to capture the emergent semantics arising from their spatial co-occurrence (e.g., reasoning that a cluster of cafes, bookstores, and galleries signifies an artistic and cultural district). To bridge this gap, we leverage the extensive world knowledge embedded in LLMs to interpret POI distributions. Specifically, we employ Qwen3-8B~\cite{yang2025qwen3} to distill high-level textual descriptions from raw POI sequences, which are subsequently encoded by Qwen3-Embedding-8B~\cite{zhang2025qwen3} into dense environmental vectors $\mathbf{e}_i^{\text{fine}} \in \mathbb{R}^d$ (please refer to \cref{subsec:appendix_prompt_fine} for prompt templates and illustrative examples).

\begin{figure}[!t]
  \centering
  \subfloat[Beijing]{\includegraphics[width=0.475\linewidth]{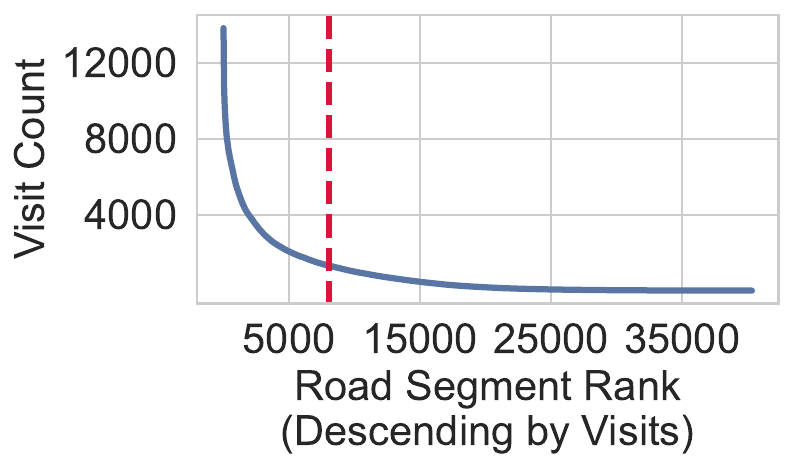}}\hfil
  \subfloat[Chengdu]{\includegraphics[width=0.475\linewidth]{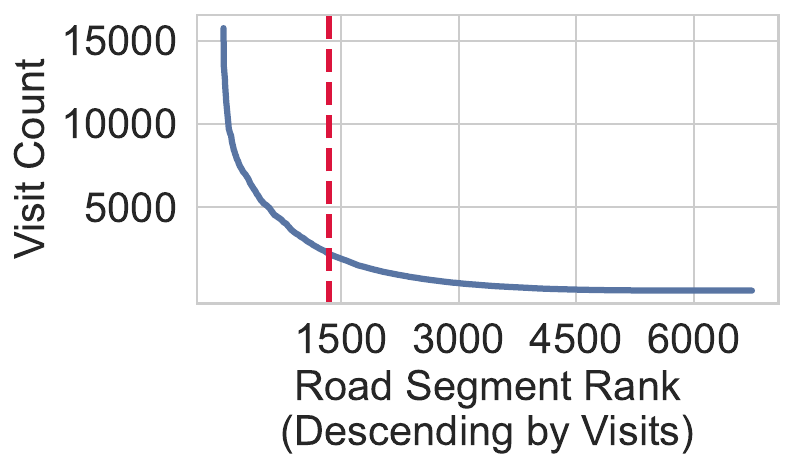}}
  \caption{Visit count distributions over road segments in Beijing and Chengdu. Segments to the left of the red dashed line are the top 20\% most visited.}
  \label{fig:road_vis_cnt}
\end{figure}

Nevertheless, applying LLM-based descriptions to every road segment in a large-scale network is computationally prohibitive. For instance, the main urban area of Beijing contains approximately \num{40000} road segments, and segment-wise processing leads to severe scalability issues. Fortunately, as shown in \cref{fig:road_vis_cnt}, visitation is strongly long-tailed: a small set of \emph{critical segments} carries the majority of traffic, while most segments are only sparsely visited.

Motivated by these observations, we design an efficient semantic representation scheme. Specifically, we select the top 20\% of segments with the highest traffic volume as \emph{critical segments} and compute their $\mathbf{e}_i^{\text{fine}}$ using the above LLM-based pipeline; all remaining segments are initialized to zero. Thus, the initial node features of the subsequent GAT are defined as:
\begin{equation}
\mathbf{h}_i^{(0)} =
\begin{cases}
\mathbf{e}_i^{\text{fine}}, & \text{if } i \text{ is in the top }20\% \text{ by volume},\\
\mathbf{0}, & \text{otherwise}.
\end{cases}
\end{equation}
Starting from this initialization, we use a three-layer GAT to propagate fine-grained semantic signals over the road network. The feature propagation and aggregation at each layer $\ell \in \{0, 1, 2\}$ is:
\begin{equation}
\mathbf{h}_i^{(\ell+1)} = \sigma\Big( \sum_{j \in \mathcal{N}(i) \cup \{i\}} \alpha_{ij}^{(\ell)} \mathbf{W}^{(\ell)} \mathbf{h}_j^{(\ell)} \Big),
\end{equation}
where the attention coefficient $\alpha_{ij}^{(\ell)}$ is computed as:
\begin{equation}
\resizebox{0.89\linewidth}{!}{
\begin{math}
\alpha_{ij}^{(\ell)} = \frac
{\exp \Big( \sigma\big({\mathbf{a}^{(\ell)}}^{\mathrm{T}} [\mathbf{W}^{(\ell)} \mathbf{h}_i^{(\ell)} \Vert \mathbf{W}^{(\ell)} \mathbf{h}_j^{(\ell)}] \big) \Big)}
{\sum_{k \in \mathcal{N}(i) \cup \{ i \}} \exp \Big( \sigma\big({\mathbf{a}^{(\ell)}}^{\mathrm{T}} [\mathbf{W}^{(\ell)} \mathbf{h}_i^{(\ell)} \Vert \mathbf{W}^{(\ell)} \mathbf{h}_k^{(\ell)}]\big) \Big)}
.
\end{math}
}
\end{equation}
Here, $\mathcal{N}(i)$ denotes the outgoing neighboring road segments of segment $i$, $\mathbf{a}^{(\ell)} \in \mathbb{R}^{2d}$ and $\mathbf{W}^{(\ell)} \in \mathbb{R}^{d \times d}$ are learnable parameters, and $\sigma$ is a non-linear activation function. The choice of the 20\% selection ratio is empirically validated in \cref{subsec:hyperparameter_sensitivity}, which demonstrates that this sparse initialization strategy achieves performance nearly identical to full-coverage LLM reasoning (covering 100\% of segments), while effectively reducing computational overhead. After three layers, the propagation expands the semantic coverage from the initial 20\% to the majority of the road network, yielding a final representation $\tilde{\mathbf{e}}_i^{\text{fine}} = \mathbf{h}_i^{(3)}$ enriched with local environmental context. For the remaining uncovered segments, valid representations are still effectively characterized by the subsequent coarse-grained semantics and basic road attributes.

\subsubsection{Coarse-Grained Semantic Modeling}

While fine-grained semantic modeling captures local environmental attributes at the level of individual road segments, it does not fully characterize each segment's role within the city's broader functional structure, which in turn shapes route choice behavior. To address this limitation, we introduce coarse-grained environmental semantic modeling, which encodes the urban functional structure and complements the fine-grained features.

Given that urban functions (e.g., commercial, residential) are unevenly distributed and exhibit spatial clustering~\cite{harris1945nature, yuan2012discovering}, we model these concentrations as \emph{functional hotspots} to characterize the urban functional structure. Let the urban domain be $\Omega=[0,W)\times[0,H)$ with horizontal and vertical extents $W$ and $H$. Then the $(i,j)$-th grid is defined as:
\begin{equation}
\resizebox{0.89\linewidth}{!}{
\begin{math}
g_{(i, j)} = \big[ (i-1)L, \min\{iL, W\} \big) \times \big[ (j-1)L, \min \{jL, H\} \big),
\end{math}
}
\end{equation}
for $i = 1, \ldots, \lceil \frac{W}{L} \rceil$ and $j = 1, \ldots, \lceil \frac{H}{L} \rceil$, where $L$ is the grid side length (set to \SI{1000}{m}). Subsequently, for each POI category $c$, we quantify the POI density within each grid and identify \emph{functional hotspots} $\mathcal{H}_c$ as the top~10\% of grids with the highest counts. For any hotspot grid $g_{(i, j)}$ of category $c$, we follow the fine-grained semantic pipeline: the grid's category-$c$ POI set is summarized with Qwen3-8B~\cite{yang2025qwen3} and the generated text is embedded by Qwen3-Embedding-8B~\cite{zhang2025qwen3} to obtain a $d$-dimensional vector (prompt templates and illustrative examples in \cref{subsec:appendix_prompt_coarse}). Grids that are not hotspots are assigned the zero vector. Formally, we define:
\begin{equation}
\resizebox{0.89\linewidth}{!}{
\begin{math}
\mathbf{e}_{(i, j), c}^{\text{coarse}} = \left\{
\begin{aligned}
&\mathcal{F} \big(\{ \boldsymbol{p} \mid \boldsymbol{p} \ \text{in} \ g_{(i, j)}, \boldsymbol{p}.{\text{category}} = c \}\big), && g_{(i, j)} \in \mathcal{H}_c \\
&\mathbf{0}, && g_{(i, j)} \notin \mathcal{H}_c
\end{aligned}
\right.
\end{math}
}
\end{equation}
where $\mathcal{F}(\cdot)$ denotes the composition of an LLM-based description step followed by a text embedding encoder.

To capture spillover effects, in which \emph{functional hotspots} influence adjacent grids (e.g., commercial centers substantially influence the spatiotemporal patterns of surrounding traffic), we apply a $3 \times 3$ convolution kernel to the feature maps of each POI category $c$:
\begin{equation}
\tilde{\mathbf{e}}_{(i, j), c}^{\text{coarse}} = \sum_{a=-1}^1 \sum_{b=-1}^1 \mathbf{W}_{(a, b)} \mathbf{e}_{(i+a, j+b), c}^{\text{coarse}},
\end{equation}
with learnable weights $\mathbf{W}_{(a, b)} \in \mathbb{R}^{d \times d}$ for each relative position $(a, b)$, and zero-padding applied at grid boundaries.

Finally, we perform average pooling along the POI category dimension within each grid, yielding $\tilde{\mathbf{e}}_{(i, j)}^{\text{coarse}}$ as the representation of grid $(i, j)$.

\subsubsection{Multi-Granular Semantic Fusion}

To construct a unified, context-aware representation of urban road networks, we integrate the aforementioned multi-granular environmental semantics with basic segment attributes, where the attribute vector $\mathbf{r}_i \in \mathbb{R}^d$ for segment $i$ is the element-wise sum of the embeddings of road type, length, in-degree, and out-degree:
\begin{equation}
\mathbf{r}_i = \mathbf{r}_i^\text{type} + \mathbf{r}_i^\text{length} + \mathbf{r}_i^\text{in} + \mathbf{r}_i^\text{out}.
\label{eq:road_emb}
\end{equation}
We then fuse these attributes with the fine- and coarse-grained environmental semantics through a gating mechanism, yielding the final context-aware road segment representation $\tilde{\mathbf{r}}_i$:
\begin{equation}
\resizebox{0.89\linewidth}{!}{
\begin{math}
\tilde{\mathbf{r}}_i = \mathbf{r}_i + \tilde{\mathbf{e}}_i^{\text{fine}} \odot \mathrm{Gate}(\tilde{\mathbf{e}}_i^{\text{fine}} \! \parallel \! \mathbf{r}_i) + \tilde{\mathbf{e}}_{\text{grid}(i)}^{\text{coarse}} \odot \mathrm{Gate}(\tilde{\mathbf{e}}_{\text{grid}(i)}^{\text{coarse}} \! \parallel \! \mathbf{r}_i).
\end{math}
}
\end{equation}
Here, $\mathrm{Gate}(\cdot)$ denotes a linear projection from $\mathbb{R}^{2d}$ to $\mathbb{R}^{d}$ followed by a $\tanh$ activation, applied to the concatenated context--road representation in each branch.

\subsection{Route Choice Encoder}

Building on the context-aware road network, the Route Choice Encoder is designed to characterize the sequential route choice process within trajectories. Given a trajectory as a sequence of road segments $(r_1,\ldots,r_n)$, each non-terminal step $i<n$ selects the next segment $r_{i+1}$ from the outgoing neighbor set $\mathcal{N}(r_i)$. We represent this step using two feature groups: a \emph{current state} $\mathbf{s}_{\text{current}} \in \mathbb{R}^{2d}$ and an \emph{adjacent route context} $\mathbf{s}_{r_c}\in \mathbb{R}^{3d}$:
\begin{itemize}
  \item \textbf{Current State.} For the current segment $r_i$, let $\tilde{\mathbf r}_{r_i}\in\mathbb R^{d}$ be its context-aware representation. We augment it with a \emph{journey progression} scalar $\rho_i\in[0,1]$ to obtain:
  \begin{equation}
  \mathbf{s}_{\text{current}} = \tilde{\mathbf{r}}_{r_i} \parallel \mathbf{w}_{\rho} \rho_i,
  \end{equation}
  where $\mathbf{w}_{\rho} \in \mathbb{R}^{d}$ is a learnable vector, and
  \begin{equation}
  \rho_i = \frac{\sum_{j=1}^i r_j.{\text{length}}}{\sum_{j=1}^n r_j.{\text{length}}} \in [0, 1].
  \end{equation}

  \item \textbf{Adjacent Route Context.} For each candidate $r_c\in \mathcal{N}(r_i)$, let $\tilde{\mathbf{r}}_{r_c}\in\mathbb R^{d}$ be its context-aware representation. We further encode two navigational factors: the \emph{historical transition likelihood} $P(r_c \mid r_i)$ and the \emph{destination-oriented directional deviation} $\Delta\theta_{r_c}$, forming:
  \begin{equation}
  \mathbf{s}_{r_c} = \tilde{\mathbf{r}}_{r_c} \parallel \mathbf{w}_P P(r_c \mid r_i) \parallel \mathbf{w}_\Delta \Delta\theta_{r_c},
  \end{equation}
  where $\mathbf{w}_P, \mathbf{w}_\Delta \in \mathbb{R}^d$ are learnable parameters. The transition likelihood $P(r_c \mid r_i)$ is computed from historical counts $N(r_i \to r_c)$ over the outgoing neighbor set:
  \begin{equation}
  P(r_c \mid r_i) = \frac{N(r_i \to r_c)}{\sum_{r_c^{\prime} \in \mathcal{N}(r_i)} N(r_i \to r_c^{\prime})}.
  \end{equation}
  The directional deviation $\Delta\theta_{r_c}$ measures how much turning toward $r_c$ deviates from the direct bearing to the destination segment $r_n$:
  \begin{equation}
  \Delta\theta_{r_c} = \angle (\overrightarrow{r_i r_c}, \overrightarrow{r_i r_n}).
  \end{equation}
\end{itemize}

Given the substantial variability in route choice logic across different intersections (e.g., highway interchanges vs. multifunctional commercial streets), we employ DeepSeekMoE~\cite{dai2024deepseekmoe} to capture these complex patterns. Specifically, the \emph{current state} $\mathbf{s}_{\text{current}}$ is fed to the MoE router, which dynamically selects an expert ensemble tailored to the current route choice context. Subsequently, the MoE model transforms the \emph{adjacent route context} $\mathbf{s}_{r_c}$ for all candidates $r_c \in \mathcal{N}(r_i)$, yielding contextual representations $\tilde{\mathbf{s}}_{r_c}$ by the expert networks:
\begin{align}
  & \tilde{\mathbf{s}}_{r_c} = \text{FFN}(\mathbf{s}_{r_c}) + \sum_{i=1}^{n_e} \big( g_i \text{FFN}_i (\mathbf{s}_{r_c}) \big),\\
  & g_i = \left\{
    \begin{aligned}
    & s_i, && s_i \in \text{Top-k}(\{ s_j \mid 1 \leq j \leq n_e \}, k) \\
    & 0, && \text{otherwise}
    \end{aligned}
  \right.,\\
  & s_i = \text{Softmax} (\mathbf{s}_{\text{current}} \textbf{w}_i^{\mathrm{T}}),
\end{align}
where $n_e$ denotes the number of experts and $k$ is the number of activated experts (in our implementation, we set $n_e=8$ and $k=2$). $\mathrm{FFN}(\cdot)$ is a shared feed-forward network, and $\mathrm{FFN}_i(\cdot)$ is the $i$-th expert network, both yielding $d$-dimensional outputs. $\textbf{w}_i \in \mathbb{R}^{2d}$ denotes the router's learnable weight vector for the $i$-th expert.

Subsequently, to capture route choice semantics, we compute the behavior representation at $r_i$ by synthesizing information from both the \emph{selected} segment and its \emph{unselected} counterparts. Specifically, we treat the next segment $r_{i+1}$ as \emph{selected}, while the remaining outgoing neighbors $\mathcal{N}(r_i) \setminus \{r_{i+1}\}$ are regarded as \emph{unselected} (for the terminal segment $r_n$, all outgoing neighbors $\mathcal{N}(r_n)$ are considered \emph{unselected}). To aggregate the representations of all unselected segments, we employ an attention mechanism:
\begin{align}
  & \tilde{\mathbf{s}}_{\mathcal{N}(r_i) \setminus \{r_{i+1}\}} = \sum_{r \in \mathcal{N}(r_i) \setminus \{r_{i+1}\}} \alpha_{r} \tilde{\mathbf{s}}_r \textbf{W}_v,\\
  & \alpha_{r} = \text{Softmax} \big( (\textbf{W}_q \mathbf{s}_{\text{current}})(\textbf{W}_k \tilde{\mathbf{s}}_r)^{\mathrm{T}} / \sqrt{d} \big),
\end{align}
where $\textbf{W}_q \in \mathbb{R}^{d \times 2d}$, $\textbf{W}_k, \textbf{W}_v \in \mathbb{R}^{d \times d}$ are learnable parameters for query, key, and value, respectively. Finally, we encode the decision logic by concatenating the representation of the \emph{selected} segment $\tilde{\mathbf{s}}_{r_{i+1}}$ with the aggregated \emph{unselected} context $\tilde{\mathbf{s}}_{\mathcal{N}(r_i) \setminus \{r_{i+1}\}}$, and feeding them into an MLP to obtain the route choice behavior representation $\mathbf{o}_i$:
\begin{equation}
  \mathbf{o}_i = \text{MLP}(\tilde{\mathbf{s}}_{r_{i+1}} \parallel \tilde{\mathbf{s}}_{\mathcal{N}(r_i) \setminus \{r_{i+1}\}}).
\end{equation}

\subsection{Trajectory Embedding and Pretraining}

To obtain a global trajectory embedding, we first enrich the route choice behavior representation $\mathbf{o}_i$ for each spatiotemporal point $(r_i, t_i)$ with time-of-day and day-of-week context before sequence aggregation:
\begin{equation}
  \mathbf{x}_i = \mathbf{o}_{i} + \mathbf{t}_{m(t_i)}^{\mathrm{m}} + \mathbf{t}_{d(t_i)}^{\mathrm{d}}.
\end{equation}
Here, $\mathbf{t}_{m(t_i)}^{\text{m}}$ and $\mathbf{t}_{d(t_i)}^{\text{d}}$ are learnable embeddings for minute-of-day (1 to 1440) and day-of-week (1 to 7) of the timestamp $t_i$. We then prepend a learnable classification token \texttt{[CLS]} and pass the sequence to a Transformer encoder~\cite{vaswani2017attention}:
\begin{equation}
\resizebox{0.89\linewidth}{!}{
\begin{math}
(\mathbf{z}, \tilde{\mathbf{x}}_1, \ldots, \tilde{\mathbf{x}}_n) = \mathrm{TransformerEncoder}([\texttt{CLS}], \mathbf{x}_1, \ldots, \mathbf{x}_n),
\end{math}
}
\end{equation}
where $\mathbf{z}$ serves as the global trajectory embedding.

Following prior TRL studies~\cite{jiang2023self, han2025bridging}, we then adopt a contrastive learning objective for pretraining to learn robust and discriminative representations. Specifically, we utilize the following two data augmentation strategies:
\begin{itemize}
  \item \textbf{Trajectory Cropping.} To effectively capture spatial characteristics, we randomly crop the start or end of a trajectory. The crop ratio is sampled from $U(0.05, 0.15)$.
  
  \item \textbf{Temporal Perturbation.} To enhance temporal modeling, we perturb travel times for $15\%$ of the segments: for a segment with $\Delta t$, set $\Delta t^{\prime}=\Delta t-r(\Delta t - \Delta t_{\text{avg}})$, where $\Delta t_{\text{avg}}$ is the segment-wise average and $r \sim U(0.15,0.30)$.
\end{itemize}
Subsequently, for each trajectory in a batch $\mathcal{B}$, we apply both augmentations to obtain $2|\mathcal{B}|$ views and optimize with the NT-Xent loss~\cite{chen2020simple}, which promotes agreement between positive pairs while separating them from negatives:
\begin{equation}
\resizebox{0.89\linewidth}{!}{
\begin{math}
  \mathcal{L} = -\frac{1}{2 \vert \mathcal{B} \vert} \sum_{i=1}^{2\vert\mathcal{B}\vert}
  \bigg[
    \log{
      \frac
      {\exp{\big(\text{sim}(\mathbf{z}_i, \mathbf{z}_{\text{pos}(i)}) / \tau\big)}}
      {\sum_{j=1}^{2\vert\mathcal{B}\vert}{\mathbbm{1}_{[j \neq i]} \exp{\big(\text{sim}(\mathbf{z}_i, \mathbf{z}_j)/\tau\big)}}}
    }
  \bigg],
\end{math}
}
\end{equation}
where $\text{sim}(\cdot, \cdot)$ denotes cosine similarity, $\tau = 0.07$ is the temperature, and $\text{pos}(i)$ is the positive sample index for $\mathbf{z}_i$.

\section{Experiments}

To assess the effectiveness of the CORE framework, we conduct comprehensive experiments on four real-world trajectory datasets and seek to answer the following research questions:

\begin{itemize}
  \item \emph{\textbf{RQ1}: How does CORE compare with \emph{state-of-the-art} TRL methods on downstream tasks?}

  \item \emph{\textbf{RQ2}: How does each module in CORE contribute to the overall model performance?}

  \item \emph{\textbf{RQ3}: How efficient is CORE in terms of data utilization and computational costs?}

  \item \emph{\textbf{RQ4}: What insights do qualitative visualizations provide into the representations learned by CORE?}

  \item \emph{\textbf{RQ5}: How robust is CORE to variations in environmental perception hyperparameters, Transformer encoder capacity, LLM backbone, and POI data completeness?}

\end{itemize}

\subsection{Experimental Setup}

\subsubsection{Datasets}

\begin{figure}[htbp]
  \centering
  \subfloat[Beijing]{\includegraphics[width=0.19\linewidth]{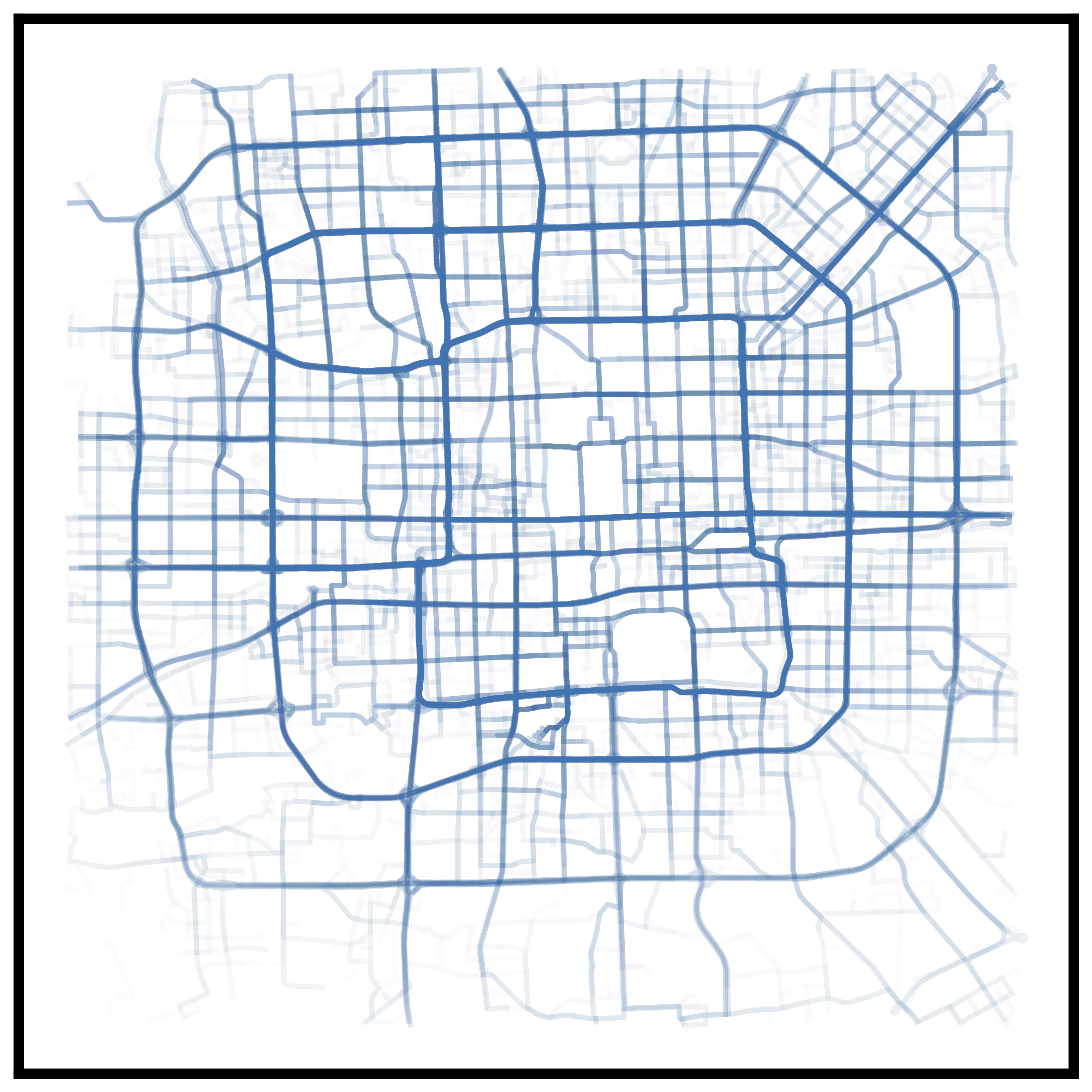}}\hfil
  \subfloat[Chengdu]{\includegraphics[width=0.19\linewidth]{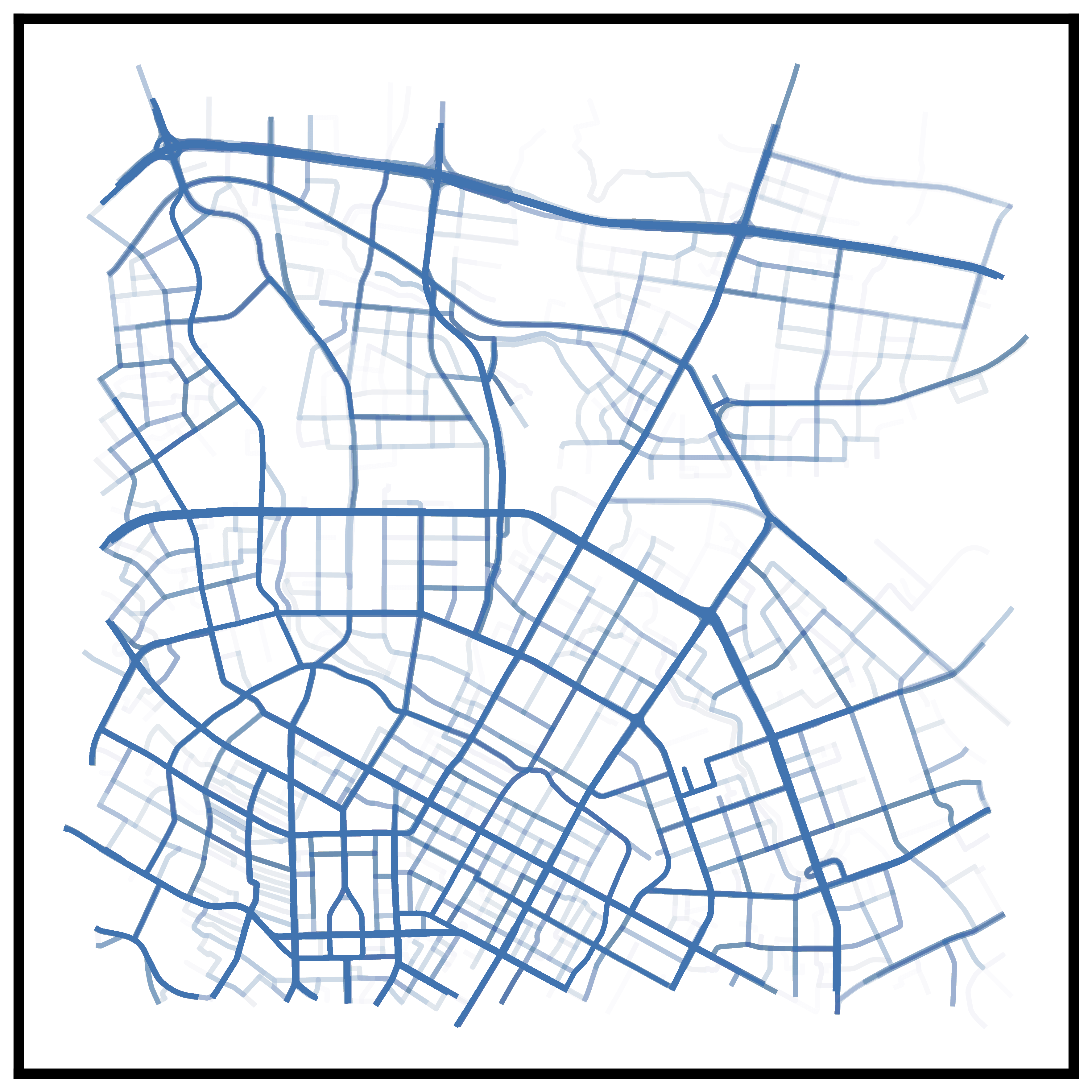}}\hfil
  \subfloat[Xi'an]{\includegraphics[width=0.19\linewidth]{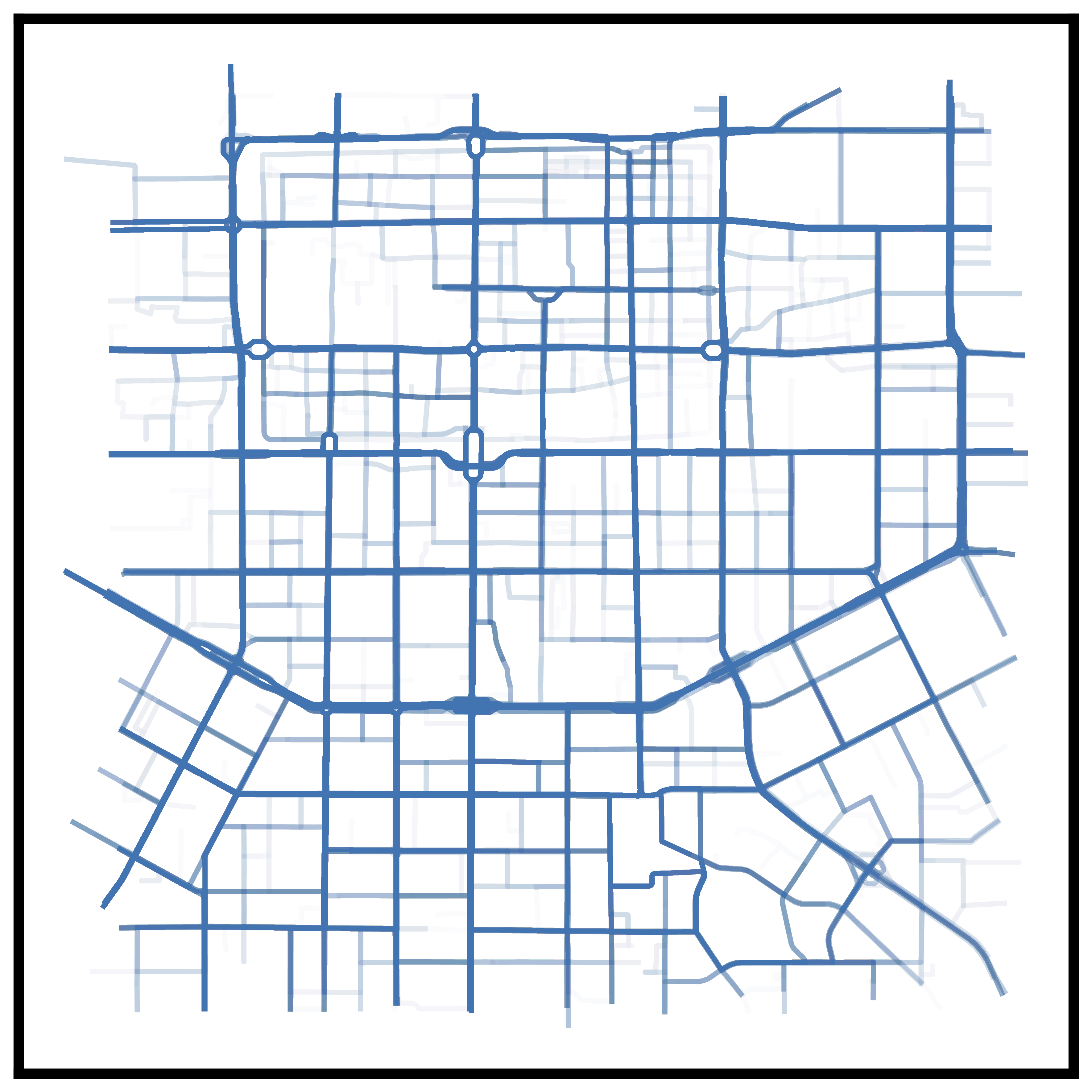}}\hfil
  \subfloat[Porto]{\includegraphics[width=0.38\linewidth]{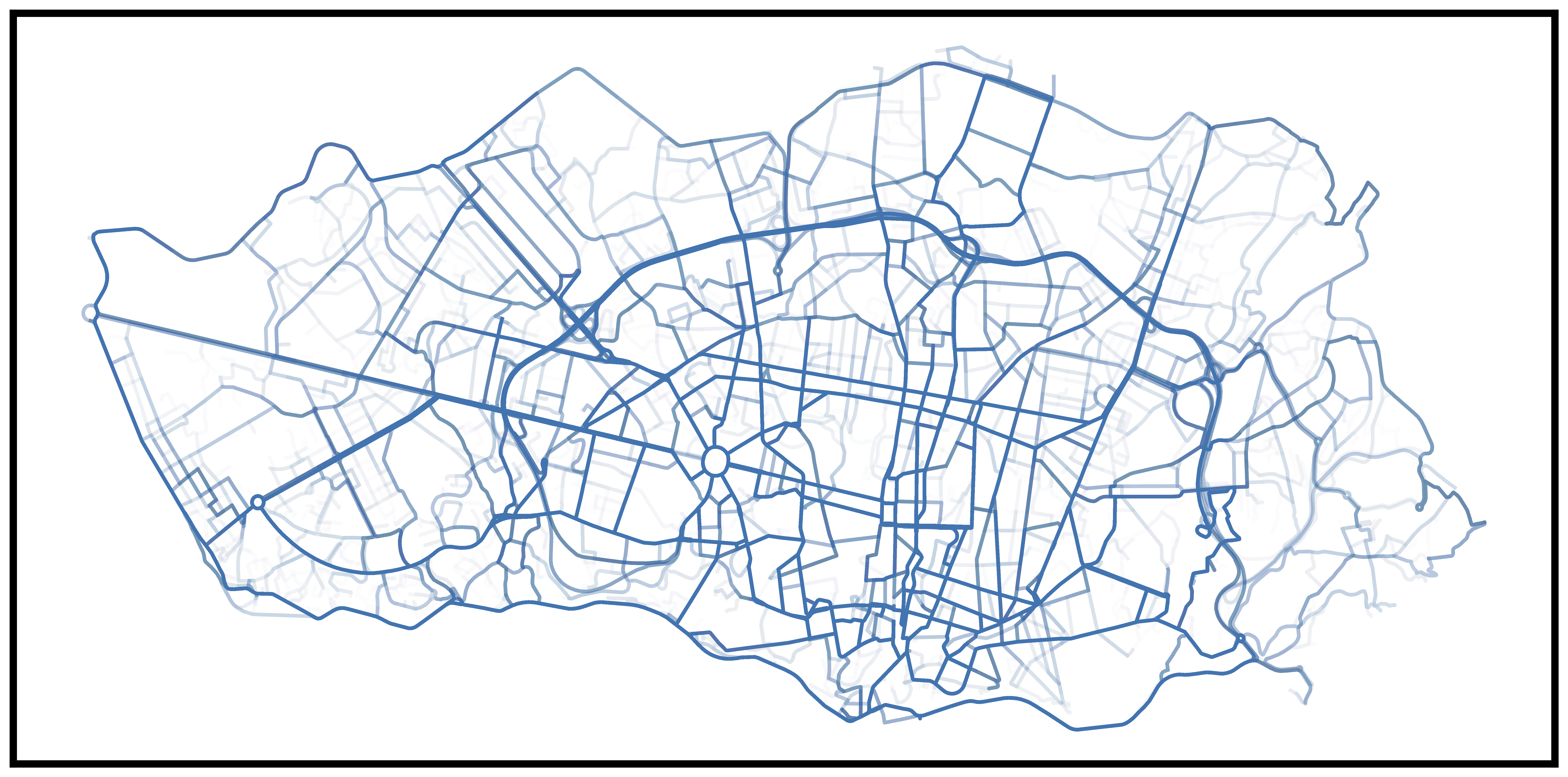}}
  \caption{Spatial distributions of the four trajectory datasets.}
  \label{fig:traj_dist}
\end{figure}

We evaluate CORE across four real-world trajectory datasets: Beijing, Chengdu, Xi'an, and Porto. The corresponding road networks are extracted from OpenStreetMap\footnote{\url{https://www.openstreetmap.org}} using OSMnx~\cite{boeing2025modeling}, and all trajectories are map-matched with FMM~\cite{yang2018fast}. Each trajectory dataset is chronologically partitioned into training, validation, and test sets in a 7:1:2 ratio. The dataset statistics are summarized in \cref{tab:traj_stats}, and their spatial distributions are visualized in \cref{fig:traj_dist}.

\begin{table}[htbp]
\caption{Statistics of the four real-world trajectory datasets.}
\label{tab:traj_stats}
\centering
\resizebox{1.0\linewidth}{!}{
\begin{tabular}{lcccc}
  \toprule
  \textbf{Dataset} & \textbf{\# Trajectories} & \textbf{\# Segments} & \textbf{Avg. Length} & \textbf{Avg. Time} \\
  \midrule
  Beijing & \num{1010325} & \num{40306} & \SI{6.60}{\kilo\meter} & \SI{15.57}{\minute} \\
  Chengdu & \num{431581} & \num{6741} & \SI{4.10}{\kilo\meter} & \SI{8.77}{\minute} \\
  Xi'an & \num{407181} & \num{6795} & \SI{4.54}{\kilo\meter} & \SI{11.59}{\minute} \\
  Porto & \num{695085} & \num{11095} & \SI{4.08}{\kilo\meter} & \SI{8.02}{\minute} \\
  \bottomrule
\end{tabular}
}
\end{table}

In addition, we collect POI data for Beijing, Chengdu, and Xi'an using the Amap API\footnote{\url{https://lbs.amap.com}}. For Porto, where the Amap API is unavailable due to geographic restrictions, we obtain POI data from OpenStreetMap via the Overpass API\footnote{\url{https://overpass-api.de}}. The statistics of the POI data are reported in \cref{tab:poi_stats}.

\begin{table}[htbp]
\caption{Statistics of the POI data.}
\label{tab:poi_stats}
\centering
\begin{tabular}{lccc}
  \toprule
  \textbf{Dataset} & \textbf{\# POIs} & \textbf{\# Categories} & \textbf{\# Subcategories} \\
  \midrule
  Beijing & \num{706115} & \num{14} & \num{128} \\
  Chengdu & \num{141918} & \num{14} & \num{126} \\
  Xi'an & \num{87309} & \num{14} & \num{124} \\
  Porto & \num{8652} & \num{22} & \num{264} \\
  \bottomrule
\end{tabular}
\end{table}

\subsubsection{Baselines}

We compare CORE against the following 15 \emph{state-of-the-art} baselines, including 12 general-purpose TRL methods and 3 trajectory
generation models (TS-TrajGen, STEGA, and HOSER):
\begin{itemize}
  \item \textbf{HRNR}~\cite{wu2020learning} employs a hierarchical GNN to jointly model both structural and functional properties of road networks.
  \item \textbf{Toast}~\cite{chen2021robust} combines a context-aware skip-gram module for modeling traffic patterns with a trajectory-enhanced Transformer for capturing travel semantics.
  \item \textbf{PIM}~\cite{yang2021unsupervised} employs mutual information maximization to learn trajectory representations.
  \item \textbf{DyToast}~\cite{chen2025semantic} extends Toast~\cite{chen2021robust} by employing trigonometric functions to better capture temporal patterns.
  \item \textbf{JCLRNT}~\cite{mao2022jointly} jointly learns road network and trajectory representations via contrastive learning.
  \item \textbf{START}~\cite{jiang2023self} employs GAT to encode road representations enriched with travel semantics, and Transformer~\cite{vaswani2017attention} to capture periodic temporal regularities.
  \item \textbf{TS-TrajGen}~\cite{jiang2023continuous} generates trajectories by combining a two-stage GAN with the A* search algorithm.
  \item \textbf{STEGA}~\cite{wang2024spatiotemporal} performs traffic trajectory generation via semantic-aware graph learning.
  \item \textbf{JGRM}~\cite{ma2024more} jointly encodes GPS and road trajectories using a Transformer-based architecture, and is pretrained with self-supervised objectives.
  \item \textbf{HOSER}~\cite{cao2025holistic} holistically fuses semantics from multi-level road networks, multi-granular trajectories, and destination information for trajectory generation.
  \item \textbf{TrajCogn}~\cite{zhou2025trajcogn} adapts LLMs for trajectory learning. Specifically, it maps trajectory-specific spatio-temporal features into a textual space, and it uses prompts to fuse trajectory movement patterns with travel purposes.
  \item \textbf{TrajMamba}~\cite{liu2025trajmamba} leverages the Mamba architecture~\cite{dao2024transformers} for trajectory encoding, and integrates travel purposes via contrastive learning. Moreover, it utilizes knowledge distillation pre-training for trajectory compression.
  \item \textbf{Path-LLM}~\cite{wei2025path} employs LLMs to learn multi-modal trajectory representations via the fusion of topological and textual modalities, and uses two-stage overlapping curriculum learning to optimize LLM training.
  \item \textbf{TRACK}~\cite{han2025bridging} bridges traffic states and trajectory data to jointly learn dynamic representations.
  \item \textbf{GREEN}~\cite{zhou2025grid} employs CNN and GNN encoders for grid- and road-level features, respectively, and fuses them via cross-attention to construct trajectory embeddings.
\end{itemize}

\subsubsection{Implementation Details}

All experiments are conducted on a single NVIDIA RTX A6000 GPU. For a fair comparison, all baselines use the same embedding dimension as CORE ($d=128$) and the same Transformer depth (6 layers). Road segment embeddings are computed via \cref{eq:road_emb}. For pretraining and fine-tuning, all TRL methods are trained for 50 epochs using the AdamW~\cite{loshchilov2019decoupled} optimizer with a batch size of 64. The learning rate is initialized to \num{2e-4}, linearly warmed up over the first 5 epochs, and then cosine-decayed to \num{1e-6}. All reported results are averaged over five independent trials to ensure statistical reliability. For CORE's MoE-based Route Choice Encoder, we employ Loss-Free Balancing~\cite{wang2024auxiliary} to achieve load balancing among experts.

\subsection{Overall Performance (RQ1)}

We evaluate the learned representations on six downstream tasks: \emph{road label prediction} (RLP), \emph{destination prediction} (DP), \emph{travel time estimation} (TTE), \emph{similar trajectory retrieval} (STR), \emph{path ranking} (PR), and \emph{trajectory generation} (TG). These tasks assess our model's capabilities at two distinct levels: RLP evaluates intermediate road segment representations, whereas the other five evaluate final trajectory-level representations. For STR, similarity is computed directly from learned trajectory embeddings, whereas for the remaining tasks (RLP, DP, TTE, PR, and TG), we attach a task-specific MLP head and fine-tune the model end-to-end.

\subsubsection{Road Label Prediction}

Following prior work~\cite{mao2022jointly, chen2025semantic}, we evaluate the learned road segment representations by predicting the number of lanes, an attribute withheld from the inputs in \cref{eq:road_emb}. The task is optimized using cross-entropy loss and is evaluated using Macro-F1 and Micro-F1. The results in \cref{tab:rlp_result} indicate that CORE achieves the best performance across all four datasets, primarily due to its Environment Perception Module, which captures real-world environmental semantics. Path-LLM also delivers competitive performance because it dynamically fuses road topology with textual attributes, while HRNR remains competitive by leveraging hierarchical representation learning to capture both structural and functional road-network characteristics.

\begin{table}[!t]
\caption{Evaluation on Road Label Prediction Task.}
\label{tab:rlp_result}
\centering
\resizebox{1.0\linewidth}{!}{
\begin{tabular}{l *{4}{C{1.71cm}}}
  \toprule
  \multirow{2}{*}[\multirowoffset]{\textbf{Methods}} & \multicolumn{2}{c}{\textbf{Beijing}} & \multicolumn{2}{c}{\textbf{Chengdu}} \\
  \cmidrule(lr){2-3} \cmidrule(lr){4-5}
  & \textbf{Macro-F1}$\uparrow$ & \textbf{Micro-F1}$\uparrow$ & \textbf{Macro-F1}$\uparrow$ & \textbf{Micro-F1}$\uparrow$ \\
  \midrule
  HRNR & 0.7912 & 0.7990 & 0.7054 & 0.7903 \\
  Toast & 0.5868 & 0.6285 & 0.4200 & 0.7508 \\
  PIM & 0.6284 & 0.6519 & 0.4684 & 0.7569 \\
  DyToast & 0.6263 & 0.6488 & 0.4511 & 0.7548 \\
  JCLRNT & 0.6479 & 0.6751 & 0.6376 & 0.7754 \\
  START & 0.7574 & 0.7776 & 0.6535 & 0.7805 \\
  JGRM & 0.7822 & 0.7903 & 0.6819 & 0.7892 \\
  TrajCogn & 0.7341 & 0.7496 & 0.6102 & 0.7644 \\
  TrajMamba & 0.7188 & 0.7327 & 0.7044 & 0.8116 \\
  Path-LLM & \underline{0.7984} & \underline{0.8133} & \underline{0.7268} & \underline{0.8275} \\
  TRACK & 0.6744 & 0.6948 & 0.6107 & 0.7590 \\
  GREEN & 0.6705 & 0.7006 & 0.6182 & 0.7692 \\
  \cmidrule(lr){1-5}
  CORE & \textbf{0.9303} & \textbf{0.9408} & \textbf{0.8574} & \textbf{0.9009} \\
  Improvement(\%) & 16.52 & 15.68 & 17.97 & 8.87 \\
  \midrule
  \multirow{2}{*}[\multirowoffset]{\textbf{Methods}} & \multicolumn{2}{c}{\textbf{Xi'an}} & \multicolumn{2}{c}{\textbf{Porto}} \\
  \cmidrule(lr){2-3} \cmidrule(lr){4-5}
  & \textbf{Macro-F1}$\uparrow$ & \textbf{Micro-F1}$\uparrow$ & \textbf{Macro-F1}$\uparrow$ & \textbf{Micro-F1}$\uparrow$ \\
  \midrule
  HRNR & 0.6459 & 0.6651 & \underline{0.6408} & \underline{0.7172} \\
  Toast & 0.4308 & 0.5129 & 0.3779 & 0.6380 \\
  PIM & 0.4357 & 0.5155 & 0.4207 & 0.6056 \\
  DyToast & 0.4327 & 0.5149 & 0.4315 & 0.6267 \\
  JCLRNT & 0.4582 & 0.5076 & 0.4936 & 0.6312 \\
  START & 0.6225 & 0.6363 & 0.6136 & 0.6900 \\
  JGRM & 0.6376 & 0.6355 & 0.6322 & 0.7086 \\
  TrajCogn & 0.5914 & 0.6047 & 0.5091 & 0.6290 \\
  TrajMamba & 0.6374 & 0.6713 & 0.6139 & 0.7134 \\
  Path-LLM & \underline{0.6980} & \underline{0.7215} & 0.6372 & 0.7039 \\
  TRACK & 0.4652 & 0.5190 & 0.5493 & 0.6607 \\
  GREEN & 0.5290 & 0.5545 & 0.5799 & 0.6833 \\
  \cmidrule(lr){1-5}
  CORE & \textbf{0.8456} & \textbf{0.8367} & \textbf{0.8073} & \textbf{0.8551} \\
  Improvement(\%) & 21.15 & 15.97 & 25.98 & 19.23 \\
  \bottomrule
\end{tabular}
}
\end{table}

\subsubsection{Travel Time Estimation}

This task aims to predict the time a vehicle needs to complete a given path. We formulate it as a regression problem and optimize with mean squared error (MSE) loss, and we quantify the target travel time in minutes. We input only the departure time to avoid potential leakage. We report mean absolute error (MAE), root mean square error (RMSE), and mean absolute percentage error (MAPE) for evaluation. The results in \cref{tab:tte_result} show that CORE achieves the best performance, attributable to its multi-granular urban semantic perception and route choice encoding. Among the baselines, TRACK and GREEN provide competitive accuracy: TRACK incorporates historical traffic states to capture dynamic road conditions, whereas GREEN fuses complementary grid- and road-level expressions to encode regional timing and road-network structure. The LLM-enhanced baselines TrajCogn, Path-LLM, and TrajMamba also perform well, mainly benefiting from LLMs' pretrained semantic priors that help capture travel semantics predictive of trip duration.

\begin{table}[!t]
\caption{Evaluation on Travel Time Estimation Task.}
\label{tab:tte_result}
\centering
\resizebox{1.0\linewidth}{!}{
\begin{tabular}{l *{6}{C{1.0cm}}}
  \toprule

  \multirow{2}{*}[\multirowoffset]{\textbf{Methods}} & \multicolumn{3}{c}{\textbf{Beijing}} & \multicolumn{3}{c}{\textbf{Chengdu}} \\
  \cmidrule(lr){2-4} \cmidrule(lr){5-7}
  & \textbf{MAE}$\downarrow$ & \textbf{RMSE}$\downarrow$ & \textbf{MAPE}$\downarrow$ & \textbf{MAE}$\downarrow$ & \textbf{RMSE}$\downarrow$ & \textbf{MAPE}$\downarrow$ \\
  \midrule

  HRNR & 5.1072 & 9.4108 & 0.3388 & 1.7796 & 2.7743 & 0.2308 \\
  Toast & 5.1042 & 9.2255 & 0.3423 & 1.7804 & 2.7730 & 0.2301 \\
  PIM & 5.1622 & 9.3977 & 0.3456 & 1.7877 & 2.7778 & 0.2292 \\
  DyToast & 4.2883 & 8.3221 & 0.2863 & 1.5071 & 2.3719 & 0.2025 \\
  JCLRNT & 5.0191 & 9.1380 & 0.3281 & 1.7959 & 2.7878 & 0.2343 \\
  START & 5.0103 & 9.1894 & 0.3376 & 1.4901 & 2.3406 & 0.2010 \\
  JGRM & 4.4038 & 8.5129 & 0.2954 & 1.4523 & 2.3005 & 0.1899 \\
  TrajCogn & 4.1578 & 8.2276 & 0.2799 & 1.4436 & 2.2701 & 0.1890 \\
  TrajMamba & 4.0791 & 8.2476 & 0.2863 & 1.4415 & 2.2547 & 0.1876 \\
  Path-LLM & 4.0830 & 8.1891 & 0.2845 & 1.4281 & 2.2316 & 0.1880 \\
  TRACK & \underline{4.0375} & \underline{8.0509} & \underline{0.2773} & 1.4307 & 2.2401 & \underline{0.1864}  \\
  GREEN & 4.1657 & 8.2444 & 0.2915 & \underline{1.4259} & \underline{2.2245} & 0.1882 \\
  \cmidrule(lr){1-7}
  CORE & \textbf{3.6703} & \textbf{6.9956} & \textbf{0.2551} & \textbf{1.3681} & \textbf{2.1719} & \textbf{0.1801} \\
  Improvement(\%) & 9.09 & 13.11 & 8.01 & 4.05 & 2.36 & 3.38 \\
  \midrule

  \multirow{2}{*}[\multirowoffset]{\textbf{Methods}} & \multicolumn{3}{c}{\textbf{Xi'an}} & \multicolumn{3}{c}{\textbf{Porto}} \\
  \cmidrule(lr){2-4} \cmidrule(lr){5-7}
  & \textbf{MAE}$\downarrow$ & \textbf{RMSE}$\downarrow$ & \textbf{MAPE}$\downarrow$ & \textbf{MAE}$\downarrow$ & \textbf{RMSE}$\downarrow$ & \textbf{MAPE}$\downarrow$ \\
  \midrule

  HRNR & 2.9449 & 4.7241 & 0.2835 & 1.5942 & 2.3751 & 0.2504 \\
  Toast & 2.9350 & 4.6983 & 0.2824 & 1.6225 & 2.4240 & 0.2541 \\
  PIM & 2.9540 & 4.7358 & 0.2786 & 1.6647 & 2.4892 & 0.2587 \\
  DyToast & 2.2856 & 3.7257 & 0.2282  & 1.4581 & 2.2460 & 0.2406 \\
  JCLRNT & 2.9791 & 4.7513 & 0.2869 & 1.6093 & 2.4102 & 0.2544 \\
  START & 2.3700 & 3.7994 & 0.2482  & 1.6077 & 2.4095 & 0.2536 \\
  JGRM & 2.3021 & 3.8054 & 0.2291 & 1.4238 & 2.2529 & 0.2270 \\
  TrajCogn & 2.2945 & 3.7688 & 0.2283 & 1.4431 & 2.3142 & 0.2302 \\
  TrajMamba & 2.2806 & 3.7123 & 0.2112 & 1.4158 & 2.1972 & 0.2268 \\
  Path-LLM & 2.2675 & 3.6407 & 0.2092 & 1.4207 & 2.2297 & 0.2274 \\
  TRACK & \underline{2.2514} & \underline{3.5998} & \underline{0.2047} & \underline{1.4001} & \underline{2.1877} & \underline{0.2247} \\
  GREEN & 2.2743 & 3.7211 & 0.2056 & 1.4056 & 2.1930 & 0.2283 \\
  \cmidrule(lr){1-7}

  CORE & \textbf{2.1498} & \textbf{3.5783} & \textbf{0.1951} & \textbf{1.3519} & \textbf{2.1151} & \textbf{0.2198} \\
  Improvement(\%) & 4.51 & 0.60 & 4.69 & 3.44 & 3.32 & 2.18 \\
  \bottomrule
\end{tabular}
}
\end{table}

\subsubsection{Similar Trajectory Retrieval}

Following prior work~\cite{mao2022jointly}, we sample two disjoint trajectory subsets from the test set, denoted $\mathcal{Q}$ and $\mathcal{D}$. For each trajectory in $\mathcal{Q}$, we apply a detour method to construct a corresponding perturbed trajectory in $\mathcal{Q}^{\prime}$, ensuring high trajectory-level similarity between $\mathcal{Q}$ and $\mathcal{Q}^{\prime}$. The objective of this task is to accurately retrieve the detoured counterpart from $\mathcal{Q}^{\prime} \cup \mathcal{D}$ for each query trajectory in $\mathcal{Q}$. We set $|\mathcal{Q}| = 5000$ and $|\mathcal{D}| = 50000$, and compute similarity in the learned embedding space via cosine similarity. We evaluate with HR@$k$ (HR@1, HR@5) and mean reciprocal rank (MRR). As shown in \cref{tab:str_result}, CORE achieves the best performance, as its capture of route choice semantics sharpens trajectory discrimination. Additionally, GREEN and START perform well because their contrastive objectives align similar trajectories and separate dissimilar ones. However, these objectives do not explicitly capture route choice semantics or multi-granular urban context, leaving their embeddings less discriminative than CORE.

\begin{table}[!t]
\caption{Evaluation on Similar Trajectory Retrieval Task.}
\label{tab:str_result}
\centering
\resizebox{1.0\linewidth}{!}{
\begin{tabular}{l *{6}{C{1.0cm}}}
  \toprule
  \multirow{2}{*}[\multirowoffset]{\textbf{Methods}} & \multicolumn{3}{c}{\textbf{Beijing}} & \multicolumn{3}{c}{\textbf{Chengdu}} \\
  \cmidrule(lr){2-4} \cmidrule(lr){5-7}
  & \textbf{HR@1}$\uparrow$ & \textbf{HR@5}$\uparrow$ & \textbf{MRR}$\uparrow$ & \textbf{HR@1}$\uparrow$ & \textbf{HR@5}$\uparrow$ & \textbf{MRR}$\uparrow$ \\
  \midrule
  HRNR & 0.8653 & 0.9226 & 0.8914 & 0.6069 & 0.7705 & 0.6782 \\
  Toast & 0.8532 & 0.9112 & 0.8792 & 0.5739 & 0.7180 & 0.6100 \\
  PIM & 0.8060 & 0.8615 & 0.8323 & 0.5498 & 0.7012 & 0.6025 \\
  DyToast & 0.8773 & 0.9259 & 0.8998 & 0.6256 & 0.7811 & 0.6962 \\
  JCLRNT & 0.8985 & 0.9369 & 0.9166 & 0.7521 & 0.8706 & 0.8069 \\
  START & 0.9056 & \underline{0.9442} & \underline{0.9231} & 0.7859 & 0.8911 & 0.8337 \\
  JGRM & 0.8991 & 0.9412 & 0.9190 & 0.7903 & 0.9006 & 0.8357 \\
  TrajCogn & 0.8530 & 0.9048 & 0.8622 & 0.7574 & 0.8620 & 0.8087 \\
  TrajMamba & 0.8788 & 0.9118 & 0.8765 & 0.7758 & 0.8506 & 0.7965 \\
  Path-LLM & 0.8816 & 0.9374 & 0.9019 & 0.7378 & 0.8759 & 0.8266 \\
  TRACK & 0.8941 & 0.9387 & 0.9158 & 0.7712 & 0.8831 & 0.8221 \\
  GREEN & \underline{0.9078} & 0.9437 & 0.9207 & \underline{0.7992} & \underline{0.9044} & \underline{0.8394} \\
  \cmidrule(lr){1-7}
  CORE & \textbf{0.9611} & \textbf{0.9911} & \textbf{0.9729} & \textbf{0.9402} & \textbf{0.9746} & \textbf{0.9554} \\
  Improvement(\%) & 5.87 & 4.97 & 5.39 & 17.64 & 7.76 & 13.82 \\
  \midrule
  \multirow{2}{*}[\multirowoffset]{\textbf{Methods}} & \multicolumn{3}{c}{\textbf{Xi'an}} & \multicolumn{3}{c}{\textbf{Porto}} \\
  \cmidrule(lr){2-4} \cmidrule(lr){5-7}
  & \textbf{HR@1}$\uparrow$ & \textbf{HR@5}$\uparrow$ & \textbf{MRR}$\uparrow$ & \textbf{HR@1}$\uparrow$ & \textbf{HR@5}$\uparrow$ & \textbf{MRR}$\uparrow$ \\
  \midrule
  HRNR & 0.6641 & 0.8062 & 0.7251 & 0.7856 & 0.9128 & 0.8423 \\
  Toast & 0.6618 & 0.7838 & 0.6902 & 0.6520 & 0.8342 & 0.7280 \\
  PIM & 0.6188 & 0.7453 & 0.6685 & 0.6437 & 0.8165 & 0.7221 \\
  DyToast & 0.6900 & 0.8187 & 0.7488 & 0.7528 & 0.8874 & 0.8138 \\
  JCLRNT & 0.8092 & 0.8895 & 0.8451 & 0.8064 & 0.9135 & 0.8405 \\
  START & 0.8645 & 0.9214 & 0.8724 & 0.8153 & 0.9197 & 0.8459 \\
  JGRM & 0.8699 & 0.9280 & 0.8859 & 0.8104 & 0.9154 & 0.8420 \\
  TrajCogn & 0.8476 & 0.9121 & 0.8758 & 0.7834 & 0.8919 & 0.8081 \\
  TrajMamba & 0.8616 & 0.9298 & 0.8860 & 0.7998 & 0.9227 & 0.8752 \\
  Path-LLM & 0.8717 & 0.9323 & 0.8912 & 0.8089 & 0.9088 & 0.8388 \\
  TRACK & 0.8613 & 0.9255 & 0.8793 & 0.8122 & 0.9178 & 0.8446 \\
  GREEN & \underline{0.8862} & \underline{0.9403} & \underline{0.9123} & \underline{0.8501} & \underline{0.9398} & \underline{0.9145} \\
  \cmidrule(lr){1-7}
  CORE & \textbf{0.9540} & \textbf{0.9858} & \textbf{0.9685} & \textbf{0.9746} & \textbf{0.9930} & \textbf{0.9820} \\
  Improvement(\%) & 7.65 & 4.84 & 6.16 & 14.65 & 5.66 & 7.38 \\
  \bottomrule
\end{tabular}
}
\end{table}

\subsubsection{Destination Prediction}

This task aims to predict a trajectory's destination from partial observations~\cite{wu2020learning}. Specifically, given the first 50\% of a trajectory, the model predicts the terminal road segment with learning guided by a cross-entropy loss. To avoid information leakage, destination-dependent terms (i.e., \emph{journey progression} and navigational factors) are computed with respect to the endpoint of the observed 50\% prefix. As shown in \cref{tab:tdp_result}, CORE achieves the best performance. Moreover, CORE yields significant gains in Beijing compared to the best-performing baseline, while improvements in the other three cities are modest. We attribute this to road network complexity: Beijing's more intricate road network topology makes destination prediction harder, whereas simpler networks elsewhere allow baselines to approach performance ceilings. Path-LLM also performs strongly, mainly benefiting from its LLM-enhanced alignment and fusion of topological and textual cues.

\begin{table}[!t]
\caption{Evaluation on Destination Prediction Task.}
\label{tab:tdp_result}
\centering
\resizebox{1.0\linewidth}{!}{
\begin{tabular}{l *{6}{C{1.0cm}}}
  \toprule
  \multirow{2}{*}[\multirowoffset]{\textbf{Methods}} & \multicolumn{3}{c}{\textbf{Beijing}} & \multicolumn{3}{c}{\textbf{Chengdu}} \\
  \cmidrule(lr){2-4} \cmidrule(lr){5-7}
  & \textbf{Acc@1}$\uparrow$ & \textbf{Acc@5}$\uparrow$ & \textbf{Acc@10}$\uparrow$ & \textbf{Acc@1}$\uparrow$ & \textbf{Acc@5}$\uparrow$ & \textbf{Acc@10}$\uparrow$ \\
  \midrule
  HRNR & 0.0789 & 0.2019 & 0.2613 & 0.4168 & 0.6215 & 0.6995 \\
  Toast & 0.0694 & 0.1809 & 0.2374 & 0.3954 & 0.5940 & 0.6667 \\
  PIM & 0.0629 & 0.1614 & 0.2100 & 0.4073 & 0.6129 & 0.6858 \\
  DyToast & 0.0772 & 0.1846 & 0.2415 & 0.4007 & 0.6019 & 0.6703 \\
  JCLRNT & 0.0782 & 0.1879 & 0.2538 & 0.4072 & 0.6154 & 0.6932 \\
  START & 0.0856 & 0.2153 & 0.2876 & 0.4128 & 0.6231 & 0.7033 \\
  JGRM & 0.0865 & 0.2180 & 0.2905 & 0.4150 & 0.6261 & 0.7044 \\
  TrajCogn & 0.0861 & 0.2184 & 0.2910 & 0.4096 & 0.6157 & 0.6940 \\
  TrajMamba & 0.0871 & 0.2256 & 0.3009 & 0.4161 & 0.6244 & 0.6972 \\
  Path-LLM & \underline{0.0925} & \underline{0.2294} & \underline{0.3072} & \underline{0.4175} & \underline{0.6271} & \underline{0.7058} \\
  TRACK & 0.0873 & 0.2198 & 0.2939 & 0.4159 & 0.6267 & 0.7041 \\
  GREEN & 0.0913 & 0.2285 & 0.3062 & 0.4060 & 0.6137 & 0.6933 \\
  \cmidrule(lr){1-7}
  CORE & \textbf{0.1076} & \textbf{0.2647} & \textbf{0.3499} & \textbf{0.4296} & \textbf{0.6411} & \textbf{0.7189} \\
  Improvement(\%) & 16.32 & 15.39 & 13.90 & 2.90 & 2.23 & 1.86 \\
  \midrule
  \multirow{2}{*}[\multirowoffset]{\textbf{Methods}} & \multicolumn{3}{c}{\textbf{Xi'an}} & \multicolumn{3}{c}{\textbf{Porto}} \\
  \cmidrule(lr){2-4} \cmidrule(lr){5-7}
  & \textbf{Acc@1}$\uparrow$ & \textbf{Acc@5}$\uparrow$ & \textbf{Acc@10}$\uparrow$ & \textbf{Acc@1}$\uparrow$ & \textbf{Acc@5}$\uparrow$ & \textbf{Acc@10}$\uparrow$ \\
  \midrule
  HRNR & 0.3511 & 0.5800 & 0.6624 & 0.1659 & 0.3624 & 0.4540 \\
  Toast & 0.3260 & 0.5421 & 0.6247 & 0.1490 & 0.3366 & 0.4214 \\
  PIM & 0.3396 & 0.5629 & 0.6440 & 0.1378 & 0.3190 & 0.4002 \\
  DyToast & 0.3304 & 0.5451 & 0.6222 & 0.1557 & 0.3367 & 0.4254 \\
  JCLRNT & 0.3430 & 0.5687 & 0.6566 & 0.1558 & 0.3508 & 0.4450 \\
  START & 0.3499 & 0.5844 & 0.6742 & 0.1627 & 0.3649 & 0.4598 \\
  JGRM & 0.3512 & 0.5834 & 0.6721 & 0.1620 & 0.3644 & 0.4593 \\
  TrajCogn & 0.3430 & 0.5768 & 0.6631 & 0.1596 & 0.3576 & 0.4511 \\
  TrajMamba & 0.3509 & 0.5792 & 0.6650 & 0.1628 & 0.3644 & 0.4601 \\
  Path-LLM & 0.3503 & 0.5788 & 0.6641 & \underline{0.1674} & \underline{0.3696} & 0.4619 \\
  TRACK & 0.3503 & 0.5824 & 0.6705 & 0.1608 & 0.3631 & 0.4582 \\
  GREEN & \underline{0.3550} & \underline{0.5889} & \underline{0.6773} & 0.1650 & 0.3693 & \underline{0.4630} \\
  \cmidrule(lr){1-7}
  CORE & \textbf{0.3631} & \textbf{0.5977} & \textbf{0.6845} & \textbf{0.1734} & \textbf{0.3750} & \textbf{0.4694} \\
  Improvement(\%) & 2.28 & 1.49 & 1.06 & 3.58 & 1.46 & 1.38 \\
  \bottomrule
\end{tabular}
}
\end{table}

\subsubsection{Path Ranking}

This task assigns each candidate trajectory between a fixed OD pair a score in the range $[0, 1]$, where higher scores indicate greater similarity to the ``optimal'' path~\cite{yang2022context}. We generate the top-$k$ candidate paths (here, $k=10$) and use each path's IoU with the ground-truth path as the regression target. The model is optimized with MSE loss, and evaluated using Kendall's $\tau$~\cite{kendall1938new}, Spearman's $\rho$~\cite{spearman1961proof}, and MAE. Results in \cref{tab:pr_result} show that CORE ranks paths more accurately than the baselines, mainly because it better captures route choice semantics and thus more clearly distinguishes the optimal route from less probable alternatives. Among the baselines, START, TrajCogn, TrajMamba, and Path-LLM are generally competitive. START benefits from diverse self-supervised augmentations and contrastive learning, which help the model capture inter-trajectory discrepancies and thus discriminate candidate routes. TrajCogn, TrajMamba, and Path-LLM further leverage LLM-derived semantic priors to enrich trajectory representations with high-level mobility semantics, thereby improving the estimation of relative path quality.

\begin{table}[!t]
\caption{Evaluation on Path Ranking Task.}
\label{tab:pr_result}
\centering
\resizebox{1.0\linewidth}{!}{
\begin{tabular}{l *{6}{C{1.0cm}}}
  \toprule
  \multirow{2}{*}[\multirowoffset]{\textbf{Methods}} & \multicolumn{3}{c}{\textbf{Beijing}} & \multicolumn{3}{c}{\textbf{Chengdu}} \\
  \cmidrule(lr){2-4} \cmidrule(lr){5-7}
  & $\tau\uparrow$ & $\rho\uparrow$ & \textbf{MAE}$\downarrow$ & $\tau\uparrow$ & $\rho\uparrow$ & \textbf{MAE}$\downarrow$ \\
  \midrule
  HRNR & 0.6259 & 0.6749 & 0.1492 & 0.7489 & 0.7885 & 0.1007 \\
  Toast & 0.6298 & 0.6808 & 0.1444 & 0.7359 & 0.7779 & 0.1020 \\
  PIM & 0.6125 & 0.6627 & 0.1506 & 0.7502 & 0.7895 & 0.0969 \\
  DyToast & 0.6317 & 0.6822 & 0.1421 & 0.7456 & 0.7858 & 0.0981 \\
  JCLRNT & 0.6352 & 0.6871 & 0.1405 & 0.7526 & 0.7908 & 0.0958 \\
  START & 0.6421 & 0.6903 & 0.1397 & \underline{0.7559} & 0.7954 & \underline{0.0931} \\
  JGRM & 0.6357 & 0.6854 & 0.1416 & 0.7533 & 0.7934 & 0.0962 \\
  TrajCogn & 0.6462 & 0.6966 & 0.1401 & 0.7551 & 0.7954 & 0.0955 \\
  TrajMamba & 0.6484 & 0.6970 & 0.1390 & 0.7557 & \underline{0.7955} & 0.0967 \\
  Path-LLM & \underline{0.6504} & \underline{0.7020} & \underline{0.1386} & 0.7544 & 0.7942 & 0.0960 \\
  TRACK & 0.6455 & 0.6898 & 0.1402 & 0.7549 & 0.7951 & \textbf{0.0927} \\
  GREEN & 0.6302 & 0.6811 & 0.1438 & 0.7428 & 0.7824 & 0.1014 \\
  \cmidrule(lr){1-7}
  CORE & \textbf{0.6814} & \textbf{0.7290} & \textbf{0.1283} & \textbf{0.7619} & \textbf{0.8017} & 0.0934 \\
  Improvement(\%) & 4.77 & 3.85 & 7.43 & 0.79 & 0.78 & -0.76 \\
  \midrule
  \multirow{2}{*}[\multirowoffset]{\textbf{Methods}} & \multicolumn{3}{c}{\textbf{Xi'an}} & \multicolumn{3}{c}{\textbf{Porto}} \\
  \cmidrule(lr){2-4} \cmidrule(lr){5-7}
  & $\tau\uparrow$ & $\rho\uparrow$ & \textbf{MAE}$\downarrow$ & $\tau\uparrow$ & $\rho\uparrow$ & \textbf{MAE}$\downarrow$ \\
  \midrule
  HRNR & 0.6972 & 0.7459 & 0.1062 & 0.7932 & 0.8226 & 0.1206 \\
  Toast & 0.6894 & 0.7419 & 0.1094 & 0.7858 & 0.8173 & 0.1186 \\
  PIM & 0.7022 & \underline{0.7521} & 0.1038  & 0.7901 & 0.8206 & 0.1213 \\
  DyToast & 0.6975 & 0.7483 & 0.1058 & 0.7931 & 0.8232 & 0.1245 \\
  JCLRNT & \underline{0.7024} & 0.7518 & 0.1039 & 0.7949 & 0.8248 & 0.1190 \\
  START & 0.6985 & 0.7481 & 0.1047 & \underline{0.8036} & 0.8325 & 0.1172 \\
  JGRM & 0.6982 & 0.7467 & 0.1051 & 0.7984 & 0.8280 & 0.1183 \\
  TrajCogn & 0.7001 & 0.7505 & 0.1041 & 0.8029 & 0.8261 & 0.1195 \\
  TrajMamba & 0.6994 & 0.7516 & \underline{0.1034} & 0.8031 & 0.8265 & 0.1174 \\
  Path-LLM & 0.7015 & 0.7504 & 0.1037 & 0.8034 & \underline{0.8329} & \underline{0.1157} \\
  TRACK & 0.6998 & 0.7480 & 0.1044 & 0.8025 & 0.8319 & 0.1179 \\
  GREEN & 0.6969 & 0.7478 & 0.1041 & 0.7956 & 0.8255 & 0.1186 \\
  \cmidrule(lr){1-7}
  CORE & \textbf{0.7129} & \textbf{0.7630} & \textbf{0.1007} & \textbf{0.8267} & \textbf{0.8511} & \textbf{0.1103} \\
  Improvement(\%) & 1.49 & 1.45 & 2.61 & 2.87 & 2.19 & 4.67 \\
  \bottomrule
\end{tabular}
}
\end{table}

\subsubsection{Trajectory Generation}

This task aims to generate trajectories that resemble real-world ones~\cite{feng2020learning,wang2024star}. We adopt a search-based paradigm~\cite{cao2025holistic} and train the model with a next road segment prediction objective. We quantify similarity to real trajectories using Hausdorff~\cite{xie2017distributed}, DTW~\cite{keogh2002exact}, and EDR~\cite{chen2005robust}. As shown in \cref{tab:tg_result}, existing generation baselines can produce semantically plausible trajectories. However, they mainly learn trajectory transition patterns and overlook context-aware route choice among candidate roads. CORE instead models both selected and unselected road segments, thereby better reproducing fine-grained routing decisions.

\begin{table}[!t]
\caption{Evaluation on Trajectory Generation Task.}
\label{tab:tg_result}
\centering
\resizebox{1.0\linewidth}{!}{
\begin{tabular}{l *{6}{C{1.0cm}}}
  \toprule
  \multirow{2}{*}[\multirowoffset]{\textbf{Methods}} & \multicolumn{3}{c}{\textbf{Beijing}} & \multicolumn{3}{c}{\textbf{Chengdu}} \\
  \cmidrule(lr){2-4} \cmidrule(lr){5-7}
  & \textbf{Hausdorff}$\downarrow$ & \textbf{DTW}$\downarrow$ & \textbf{EDR}$\downarrow$ & \textbf{Hausdorff}$\downarrow$ & \textbf{DTW}$\downarrow$ & \textbf{EDR}$\downarrow$ \\
  \midrule
  TS-TrajGen & 0.9535 & 18.9531 & 0.6395 & 0.2067 & 2.5042 & 0.3651 \\
  STEGA & 0.6931 & 11.5059 & 0.5001 & 0.1597 & 1.5372 & 0.2987 \\
  HOSER & \underline{0.5623} & \underline{7.3345} & \underline{0.4568} & \underline{0.1300} & \underline{1.0188} & \underline{0.2495} \\
  \cmidrule(lr){1-7}
  CORE & \textbf{0.4395} & \textbf{5.8033} & \textbf{0.3511} & \textbf{0.1102} & \textbf{0.8371} & \textbf{0.2082} \\
  Improvement(\%) & 21.84 & 20.88 & 23.14 & 15.23 & 17.83 & 16.55 \\
  \midrule

  \multirow{2}{*}[\multirowoffset]{\textbf{Methods}} & \multicolumn{3}{c}{\textbf{Xi'an}} & \multicolumn{3}{c}{\textbf{Porto}} \\
  \cmidrule(lr){2-4} \cmidrule(lr){5-7}
  & \textbf{Hausdorff}$\downarrow$ & \textbf{DTW}$\downarrow$ & \textbf{EDR}$\downarrow$ & \textbf{Hausdorff}$\downarrow$ & \textbf{DTW}$\downarrow$ & \textbf{EDR}$\downarrow$ \\
  \midrule
  TS-TrajGen & 0.3258 & 4.8951 & 0.3689 & 0.7158 & 19.2583 & 0.6899 \\
  STEGA & 0.2557 & 3.4756 & 0.3104 & 1.0532 & 16.0175 & 0.6659 \\
  HOSER  & \underline{0.2481} & \underline{3.4005} & \underline{0.2978} & \underline{0.5697} & \underline{13.4901} & \underline{0.5395} \\
  \cmidrule(lr){1-7}
  CORE & \textbf{0.2093} & \textbf{2.8187} & \textbf{0.2477} & \textbf{0.4592} & \textbf{10.5123} & \textbf{0.4259} \\
  Improvement(\%) &  15.64 & 17.11 & 16.82 & 19.40 & 22.07 & 21.06 \\
  \bottomrule
\end{tabular}
}
\end{table}

\subsection{Ablation Studies (RQ2)}

\begin{figure}[!t]
    \centering
    \includegraphics[width=1.0\linewidth]{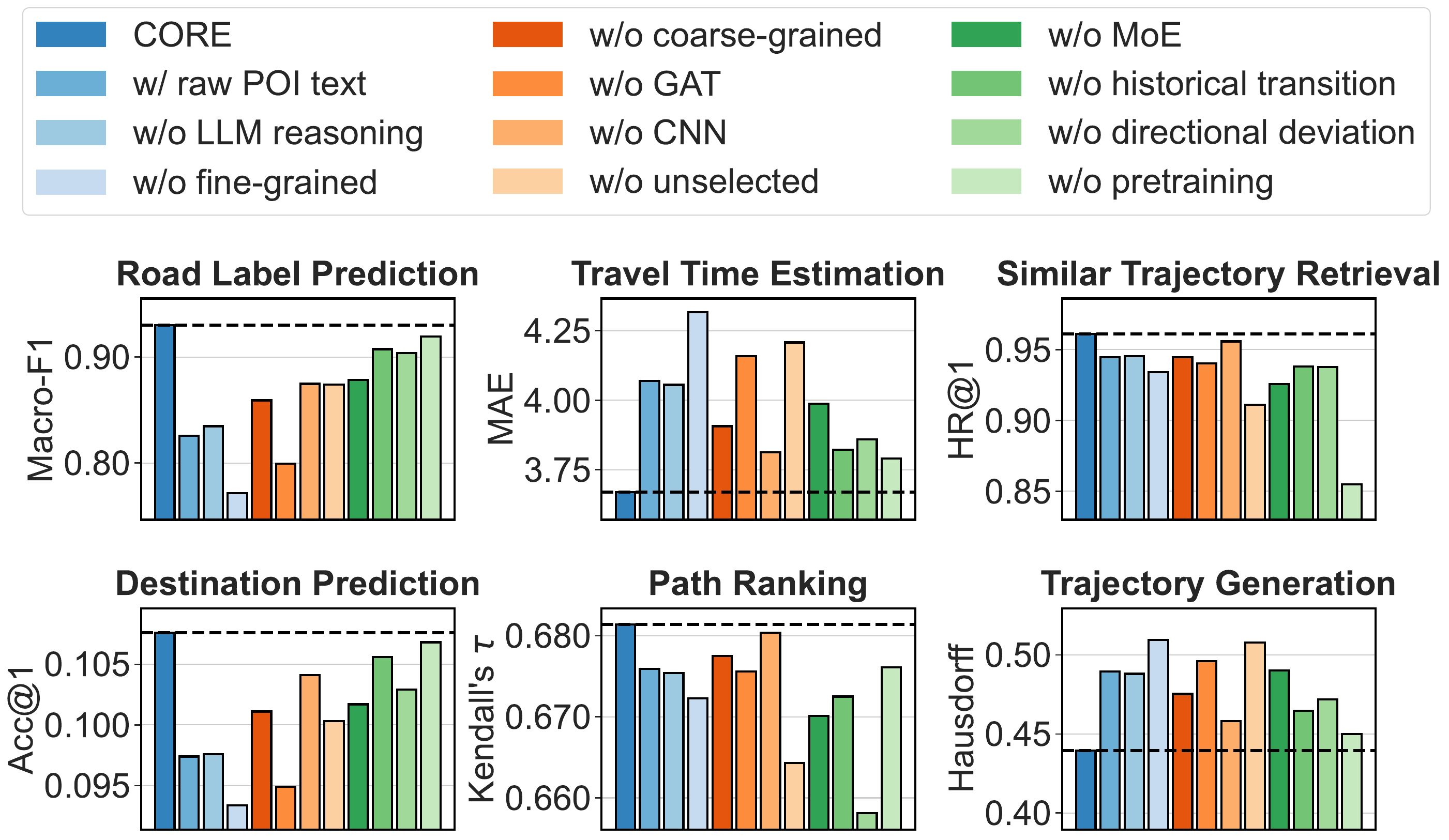}
    \caption{Performance comparison of CORE and its ablated variants on the Beijing dataset (refer to \cref{subsec:appendix_ablation} for other datasets).}
    \label{fig:ablation_beijing}
\end{figure}

To quantify the contribution of each module, we conduct a series of ablation experiments. The corresponding results are summarized in \cref{fig:ablation_beijing}. For brevity, we report a single representative metric for each task.

\subsubsection{Impact of Environment Perception Module}
(a) \emph{w/ raw POI text}: this variant directly concatenates the POI's category, subcategory, and name to derive environmental semantics via text embedding. (b) \emph{w/o LLM reasoning}: it bypasses the LLM inference process and directly inputs the prompt into the text embedding model. (c) \emph{w/o fine-grained}: this variant removes fine-grained semantic modeling. (d) \emph{w/o coarse-grained}: it removes coarse-grained environmental semantic modeling. (e) \emph{w/o GAT}: this setting ablates the GAT in fine-grained semantic modeling, so the environmental semantics of \emph{critical segments} no longer propagate. (f) \emph{w/o CNN}: it removes the convolutional layers in coarse-grained environmental semantic modeling, thereby ignoring the contextual influence of \emph{functional hotspots}. Among these variants, \emph{w/ raw POI text} and \emph{w/o LLM reasoning} incur substantial performance degradation, highlighting the LLM's pivotal role in extracting environmental semantics. Eliminating the fine-grained branch or its GAT also leads to substantial performance drops, whereas removing the coarse-grained branch or its CNN has a comparatively smaller impact. This suggests fine-grained semantics are crucial for expressive representations, while the coarse-grained branch provides useful but secondary cues.

\subsubsection{Impact of Route Choice Encoder}
(a) \emph{w/o unselected}: it does not model unselected adjacent road segments. (b) \emph{w/o MoE}: this variant replaces MoE with MLPs using the same activation parameters. (c) \emph{w/o historical transition}: this variant removes the \emph{historical transition likelihood} $P(r_c \mid r_i)$. (d) \emph{w/o directional deviation}: it removes the \emph{destination-oriented directional deviation} $\Delta\theta_{r_c}$. The results show that the variants \emph{w/o unselected} and \emph{w/o MoE} exhibit the most substantial performance degradation, indicating that jointly modeling semantics from both selected and unselected segments and employing the MoE architecture is crucial for capturing route choice patterns. Removing \emph{historical transition} or \emph{directional deviation} also consistently harms performance, highlighting the complementary importance of these navigational factors.

\subsubsection{Impact of Pretraining}
(a) \emph{w/o pretraining}: this variant removes the contrastive pretraining stage. The resulting performance drop across all tasks, particularly in \emph{similar trajectory retrieval}, confirms that the contrastive objective is critical for shaping a discriminative embedding space.

\subsection{Efficiency Analysis (RQ3)}

In this subsection, we evaluate the efficiency of CORE from three complementary perspectives: (1) \emph{Data Efficiency}, which assesses the model's performance under data-scarce conditions; (2) \emph{Training and Inference Efficiency}, which quantifies the training and inference overhead compared with competitive baselines; and (3) the \emph{Overhead of the LLM-Based Environment Perception Module}, which quantifies the preprocessing cost required for constructing the context-aware road network.

\subsubsection{Data Efficiency}

\begin{figure}[!t]
    \centering
    \includegraphics[width=1.0\linewidth]{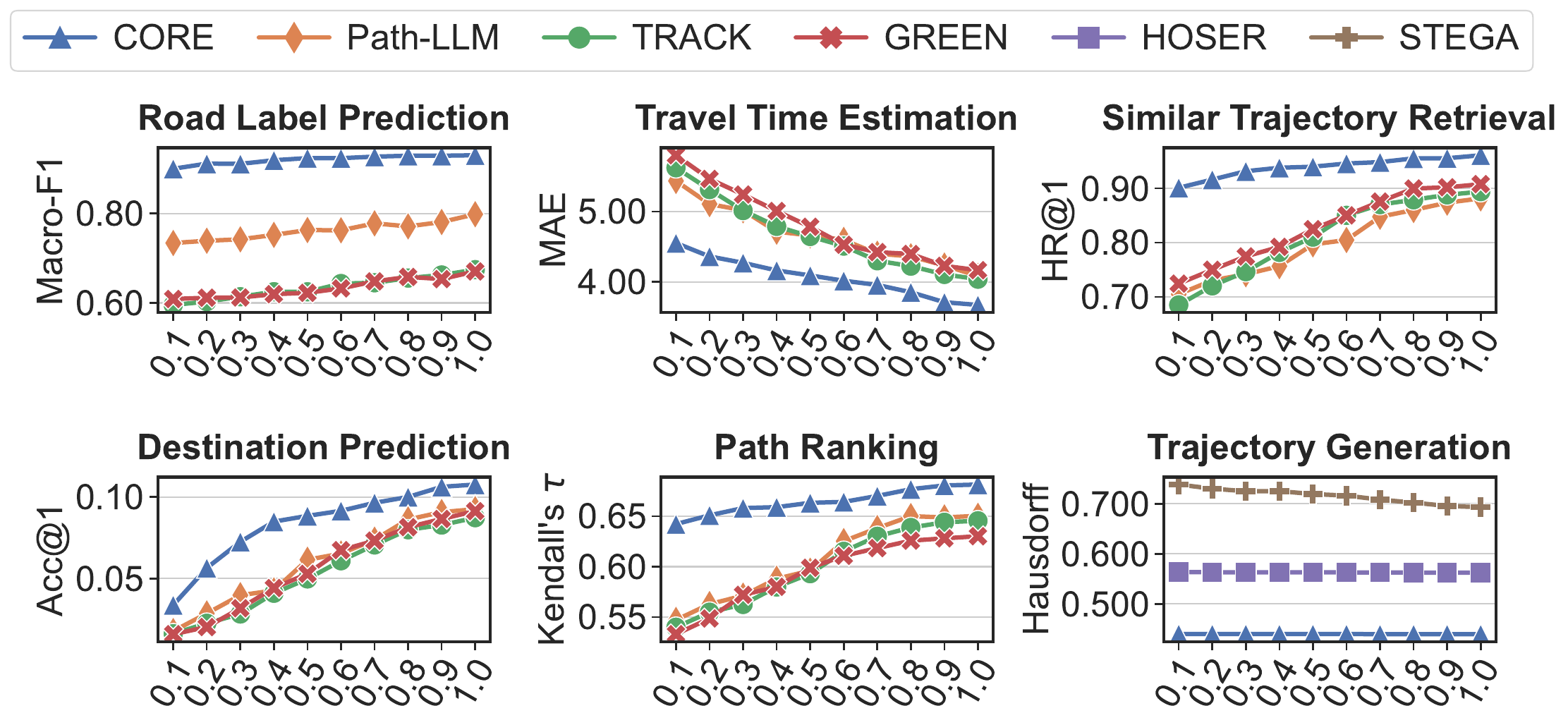}
    \caption{Results of using varying proportions of training data on the Beijing dataset (refer to \cref{subsec:appendix_data_efficiency} for other datasets).}
    \label{fig:data_efficiency_beijing}
\end{figure}

We investigate CORE's data efficiency by varying the proportion of training data used for pretraining and fine-tuning, and comparing CORE with competitive baselines. The results are shown in \cref{fig:data_efficiency_beijing}. We observe that CORE consistently outperforms baselines in the few-shot learning scenarios, demonstrating superior data efficiency. This strong few-shot learning capability primarily stems from CORE's explicit capturing of context-aware route choice semantics, which injects a powerful inductive bias into the learning process, enabling the model to effectively learn complex trajectory patterns without relying on massive amounts of data.

\begin{table}[!t]
\caption{Average training time per epoch (in hours) and inference time per trajectory (in milliseconds).}
\label{tab:training_and_inference_efficiency}
\centering
\begin{tabular}{lcccc}
  \toprule
  \multirow{2}{*}[\multirowoffset]{\textbf{Methods}} & \multicolumn{2}{c}{\textbf{Beijing}} & \multicolumn{2}{c}{\textbf{Chengdu}} \\
  \cmidrule(lr){2-3} \cmidrule(lr){4-5}
  & \textbf{Training}$\downarrow$ & \textbf{Inference}$\downarrow$ & \textbf{Training}$\downarrow$ & \textbf{Inference}$\downarrow$ \\
  \midrule
  Path-LLM & 5.0350 & 3.4363 & 0.6038 & 0.4215 \\
  TRACK & 1.6134 & 1.1778 & 0.2258 & 0.4055 \\
  GREEN & \textbf{0.5150} & \underline{0.3191} & \textbf{0.1494} & \underline{0.1897} \\
  CORE & \underline{0.8077} & \textbf{0.2785} & \underline{0.1835} & \textbf{0.1695} \\
  \midrule
  \multirow{2}{*}[\multirowoffset]{\textbf{Methods}} & \multicolumn{2}{c}{\textbf{Xi'an}} & \multicolumn{2}{c}{\textbf{Porto}} \\
  \cmidrule(lr){2-3} \cmidrule(lr){4-5}
  & \textbf{Training}$\downarrow$ & \textbf{Inference}$\downarrow$ & \textbf{Training}$\downarrow$ & \textbf{Inference}$\downarrow$ \\
  \midrule
  Path-LLM & 0.6203 & 0.4549 & 1.7498 & 1.2543 \\
  TRACK & 0.2192 & 0.4298 & 0.6207 & 0.7693 \\
  GREEN & \textbf{0.1428} & \underline{0.1987} & \textbf{0.4028} & \underline{0.3538} \\
  CORE & \underline{0.1708} & \textbf{0.1650} & \underline{0.4924} & \textbf{0.3049} \\
  \bottomrule
\end{tabular}
\end{table}

\subsubsection{Training and Inference Efficiency}

\cref{tab:training_and_inference_efficiency} compares the training and inference time of CORE against several competitive baselines. Overall, CORE achieves a favorable trade-off between predictive performance and computational cost. Its training time is slightly longer than GREEN's but remains moderate and competitive, while at inference, CORE is the most efficient among all compared methods. In contrast, Path-LLM incurs higher training and inference costs because its pipeline introduces an LLM backbone, and TRACK suffers from larger inference latency because it relies on historical traffic states for each trajectory.

\subsubsection{Overhead of LLM-Based Environment Perception}

\begin{table}[!t]
\caption{Computational overhead of the LLM-based environment perception.}
\label{tab:llm_overhead}
\centering
% \resizebox{1.0\linewidth}{!}{
\begin{tabular}{l cccc}
  \toprule
  \textbf{Dataset} & \textbf{Input Tokens} & \textbf{Output Tokens} & \textbf{Cost} & \textbf{Time} \\
  \midrule
  Beijing & \num{9581977} & \num{12912556} & \$2.12 & \qty{2}{\hour} \qty{9}{\minute} \\
  Chengdu & \num{2106161} & \num{2309139} & \$0.39 & \qty{23}{\minute} \\
  Xi'an & \num{1683295} & \num{2158924} & \$0.36 & \qty{28}{\minute} \\
  Porto & \num{948514} & \num{3098161} & \$0.46 & \qty{44}{\minute} \\
  \bottomrule
\end{tabular}
% }
\end{table}

We quantify the preprocessing overhead of the LLM-based Environment Perception Module. \cref{tab:llm_overhead} reports the API cost and time for generating semantic descriptions using Qwen3-8B via OpenRouter\footnote{\url{https://openrouter.ai}} with a concurrency of 32. Notably, even for a megacity such as Beijing with a complex road network, the process incurs a total cost of approximately \$2 and takes about 2 hours. We argue that this one-time cost of constructing a high-quality, semantically rich digital map is computationally and economically feasible.

\subsection{Visualization Analysis (RQ4)}

In this subsection, we visually validate CORE from three perspectives: (1) \emph{Road Segment Representations}, which verifies the capture of environmental semantics; (2) \emph{MoE Expert Specialization}, which interprets the learned routing patterns; and (3) \emph{Trajectory Embeddings}, which demonstrates the discriminability of the final representations.

\subsubsection{Visualization of Road Segment Representations}

\begin{figure}[!t]
  \centering
  \includegraphics[width=0.7\linewidth]{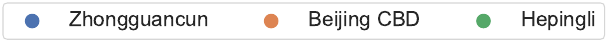}\\[-0.8em]
  \subfloat[CORE]{\includegraphics[width=0.175\linewidth]{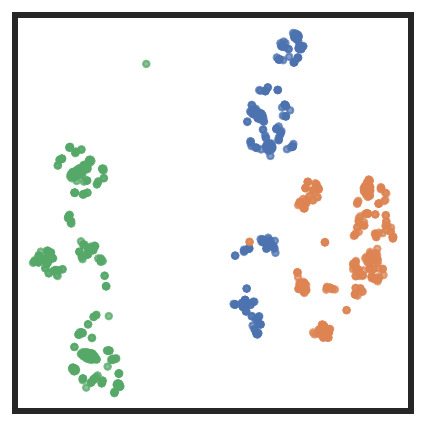}}\hfil
  \subfloat[JGRM]{\includegraphics[width=0.175\linewidth]{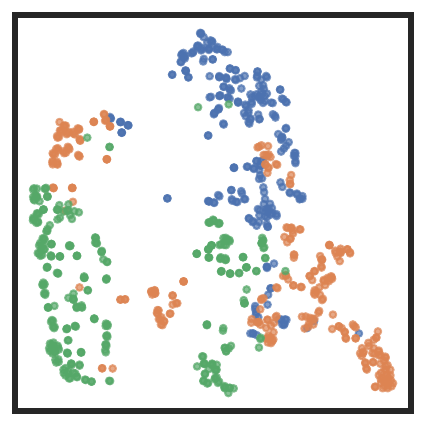}}\hfil
  \subfloat[TRACK]{\includegraphics[width=0.175\linewidth]{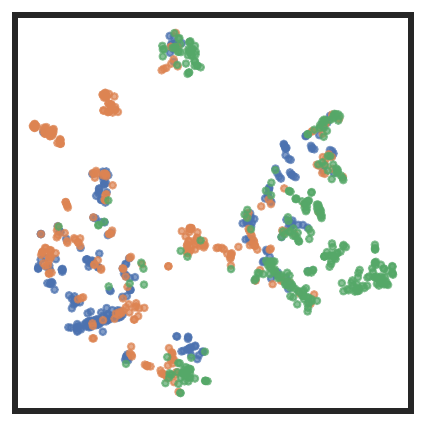}}\hfil
  \subfloat[GREEN]{\includegraphics[width=0.175\linewidth]{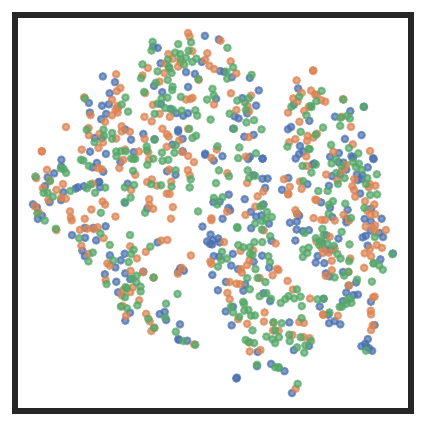}}
  \caption{t-SNE visualization of road segment representations in three representative Beijing areas (Zhongguancun, Beijing CBD, and Hepingli).}
  \label{fig:vis_road_emb}
\end{figure}

To assess whether CORE effectively captures urban environmental semantics, we visualize the learned road segment embeddings in three representative areas of Beijing: Zhongguancun (commercial district), Beijing CBD (commercial district), and Hepingli (residential district). We apply t-SNE~\cite{maaten2008visualizing} to the road segment embeddings of each method. As shown in \cref{fig:vis_road_emb}, the embeddings learned by CORE place Zhongguancun and Beijing CBD close to each other in the t-SNE space while maintaining a clear boundary between them, and both are well separated from Hepingli. This indicates that CORE effectively captures both the semantic similarities and the functional differences across urban environments. In contrast, baselines that rely solely on road network topology or trajectory information, without explicitly modeling urban environmental semantics, fail to distinguish these functional areas, leading to substantial overlap in their representations. Among these baselines, GREEN exhibits less coherent clustering than the others, likely because its grid partition strategy may split a large functional area into multiple cells, causing roads from the same semantic region to be associated with different grid-level contextual signals.

\subsubsection{Visualization of MoE Expert Specialization}

\begin{figure}[!t]
  \centering
  \subfloat[Average expert gating weight across different regions in Beijing.\label{fig:vis_moe_different_regions}]{\includegraphics[width=1.0\linewidth]{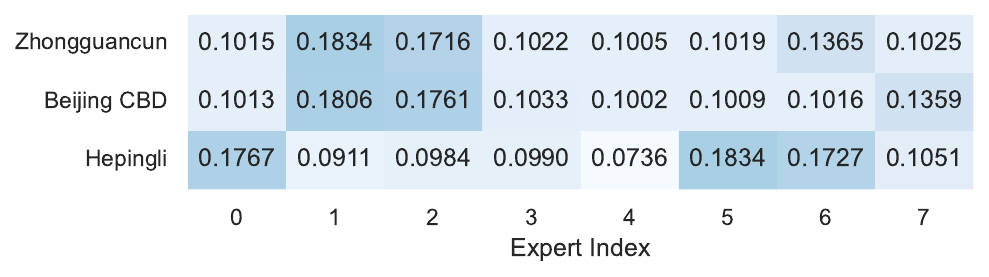}}\\
  \subfloat[Average expert gating weight across different road types in Beijing.\label{fig:vis_moe_different_road_types}]{\includegraphics[width=1.0\linewidth]{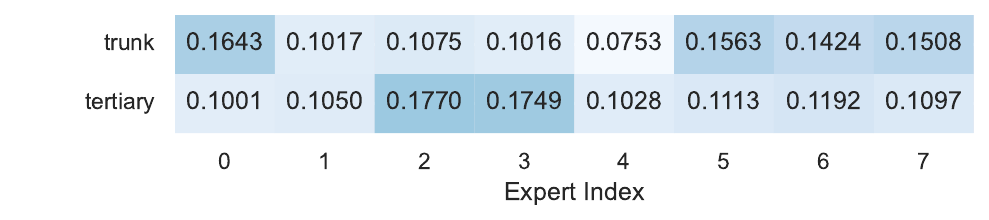}}\\
  \subfloat[Gating weights of different road segments along a trajectory in Beijing.\label{fig:vis_moe_traj}]{
  \begin{minipage}{\linewidth}
    \centering
    \includegraphics[width=0.975\linewidth]{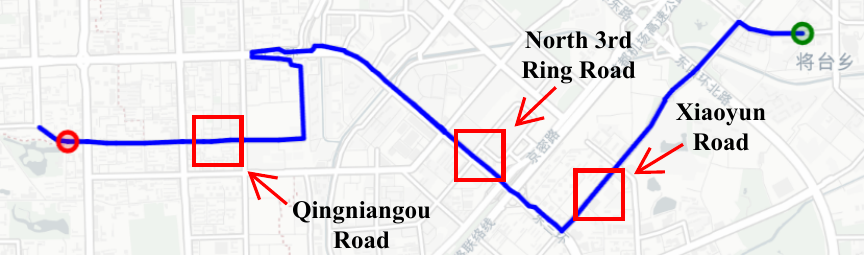}\\
    \includegraphics[width=\linewidth]{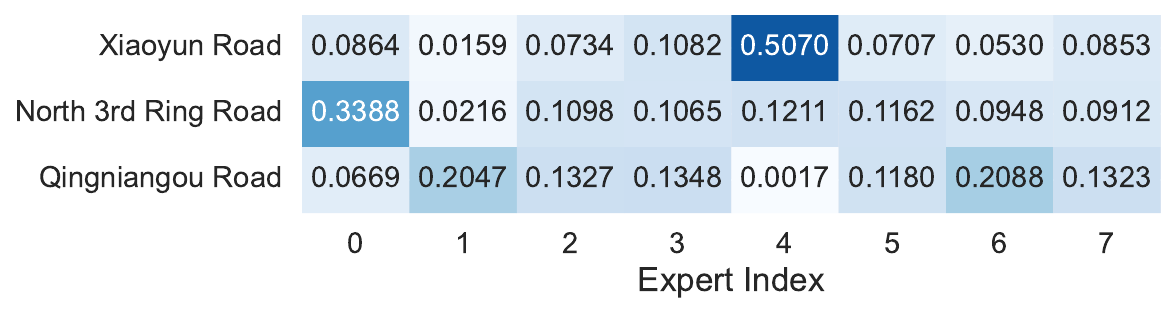}
  \end{minipage}
  }
  \caption{Visualization of MoE gating distributions.}
  \label{fig:vis_moe}
\end{figure}

To investigate the interpretability of the Route Choice Encoder, we visualize the MoE gating weight distributions in \cref{fig:vis_moe}. The results reveal distinct context-aware expert specialization across multiple granularities. At the regional level (\cref{fig:vis_moe_different_regions}), functionally analogous commercial districts (Zhongguancun and Beijing CBD) exhibit consistent expert preferences, which diverge markedly from those of Hepingli, a typical residential district. At the road-type level (\cref{fig:vis_moe_different_road_types}), major arterials (road segments whose \emph{highway} tag in OpenStreetMap is \emph{trunk}) and local streets (road segments whose \emph{highway} tag is \emph{tertiary}) are associated with substantially different expert routing patterns. In \cref{fig:vis_moe_traj}, we further examine a representative trajectory and report gating weights on three road segments with distinct environments: Xiaoyun Road, an urban arterial adjacent to high-density residential areas; the North 3rd Ring Road, a major urban traffic corridor with very high volume; and Qingniangou Road, a local street in the historic city center. The MoE activates markedly different experts across these segments. Together, these observations indicate that the MoE captures meaningful, context-aware expert specialization that is well aligned with urban semantics.

\subsubsection{Visualization of Learned Trajectory Embeddings}

\begin{figure}[!t]
  \centering
  \subfloat[Trajectories between the same origin (green box) and destination (red box) regions in Beijing, illustrating diverse route choice behaviors.]{\includegraphics[width=0.9\linewidth]{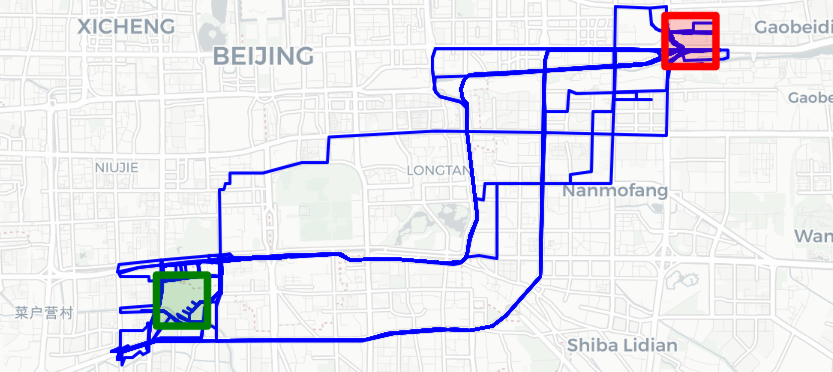}}\\
  \subfloat[CORE]{\includegraphics[width=0.18\linewidth]{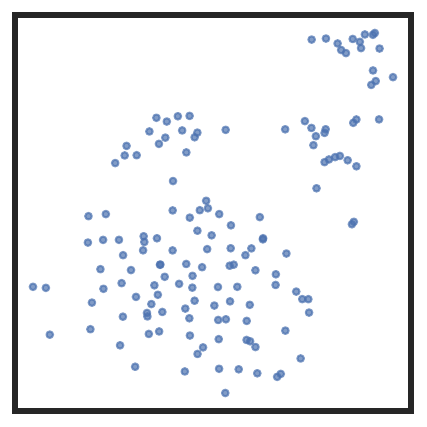}}\hfil
  \subfloat[JGRM]{\includegraphics[width=0.18\linewidth]{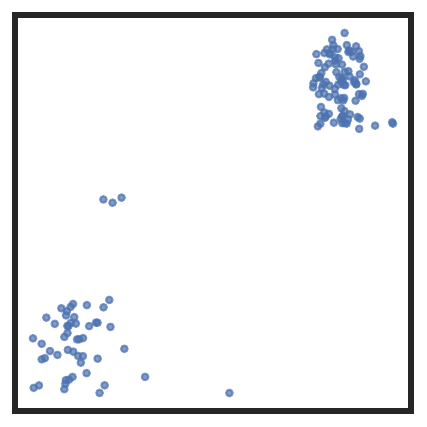}}\hfil
  \subfloat[TRACK]{\includegraphics[width=0.18\linewidth]{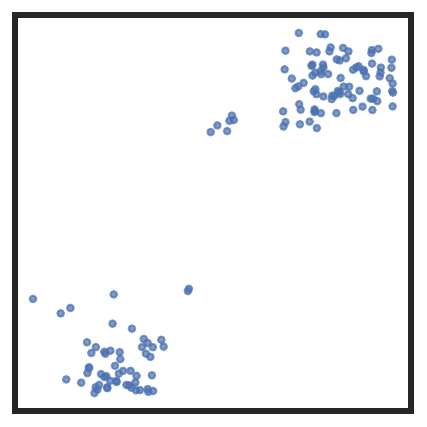}}\hfil
  \subfloat[GREEN]{\includegraphics[width=0.18\linewidth]{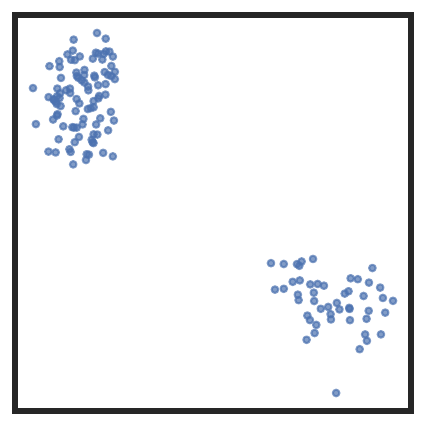}}
  \caption{t-SNE visualization of trajectory embeddings for trajectories sharing the same origin and destination areas in Beijing.}
  \label{fig:vis_traj_emb}
\end{figure}

Finally, we investigate whether CORE yields behaviorally informative trajectory embeddings. We select real trajectories exhibiting diverse route patterns between the same origin and destination areas in Beijing, and visualize the trajectory embeddings produced by each method using t-SNE~\cite{maaten2008visualizing}, as shown in \cref{fig:vis_traj_emb}. The embeddings generated by the baseline models exhibit substantial overlap and strong entanglement in the t-SNE space, which obscures differences in route choices. In contrast, the embeddings learned by CORE show markedly clearer separation among distinct route choice behaviors. These results suggest that, by explicitly capturing context-aware route choice semantics, CORE substantially enhances the discriminability of fine-grained route choice differences and produces trajectory representations that are more informative for downstream tasks.

\subsection{Sensitivity Analyses (RQ5)}

In this subsection, we systematically examine the robustness of CORE from four perspectives: (1) \emph{Environment Perception Hyperparameter Sensitivity}, which studies how spatial perception scales and the \emph{critical segment} selection ratio affect model performance; (2) \emph{Transformer Capacity Sensitivity}, which investigates how the depth and hidden dimensionality of the Transformer encoder influence downstream performance; (3) \emph{LLM Backbone Sensitivity}, which investigates how different LLMs used in the Environment Perception Module influence downstream results; and (4) \emph{POI Completeness Sensitivity}, which evaluates how CORE behaves under varying degrees of POI sparsity in practical settings.

\subsubsection{Environment Perception Hyperparameter Sensitivity}
\label{subsec:hyperparameter_sensitivity}

\begin{figure}[!t]
  \centering
  \includegraphics[width=1.0\linewidth]{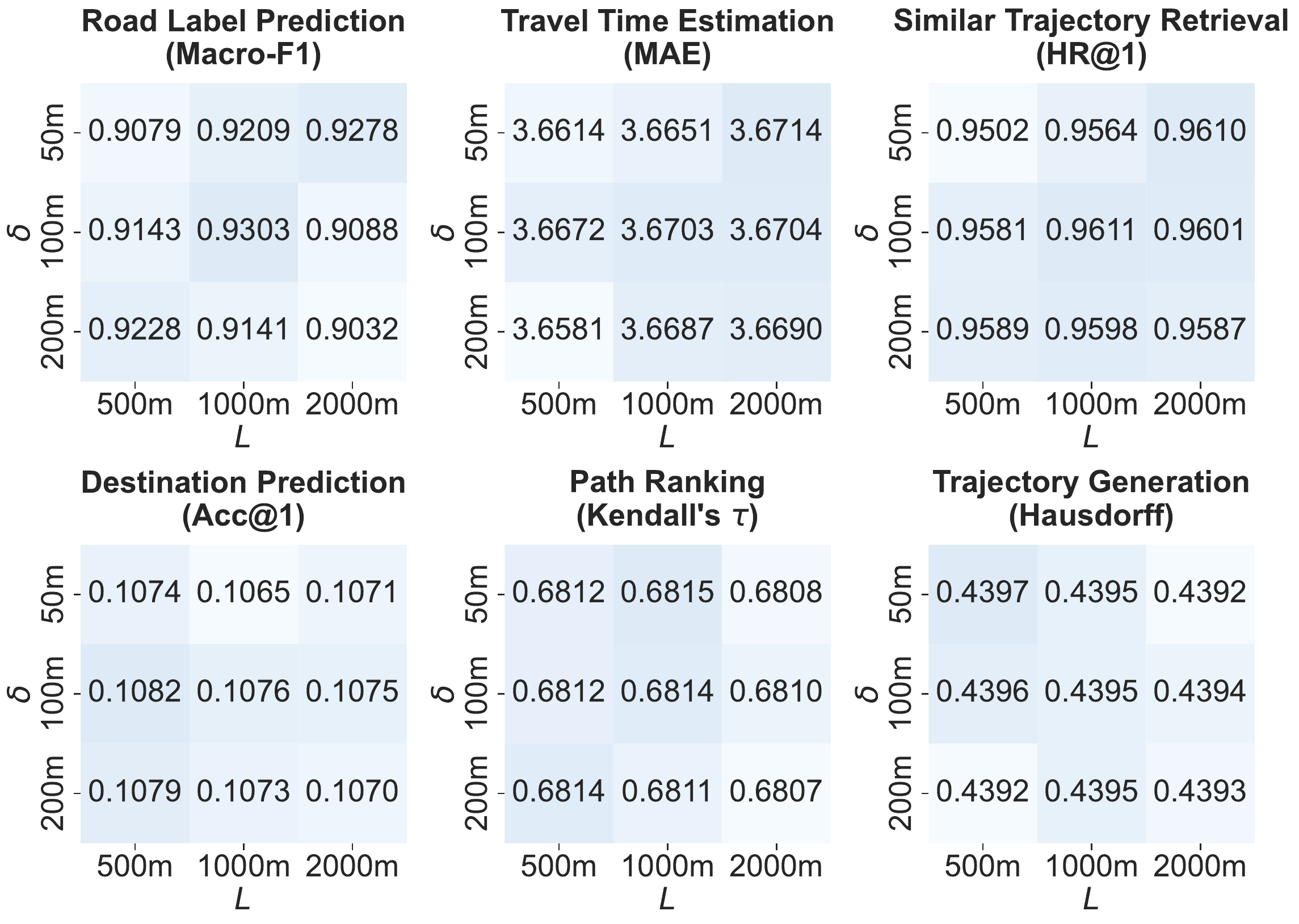}
  \caption{Sensitivity of CORE to spatial perception scale hyperparameters ($\delta$ and $L$) on the Beijing dataset (refer to \cref{subsec:appendix_hyperparameter_sensitivity} for other datasets).}
  \label{fig:hyperparameter_beijing}
\end{figure}

\begin{figure}[!t]
  \centering
  \includegraphics[width=1.0\linewidth]{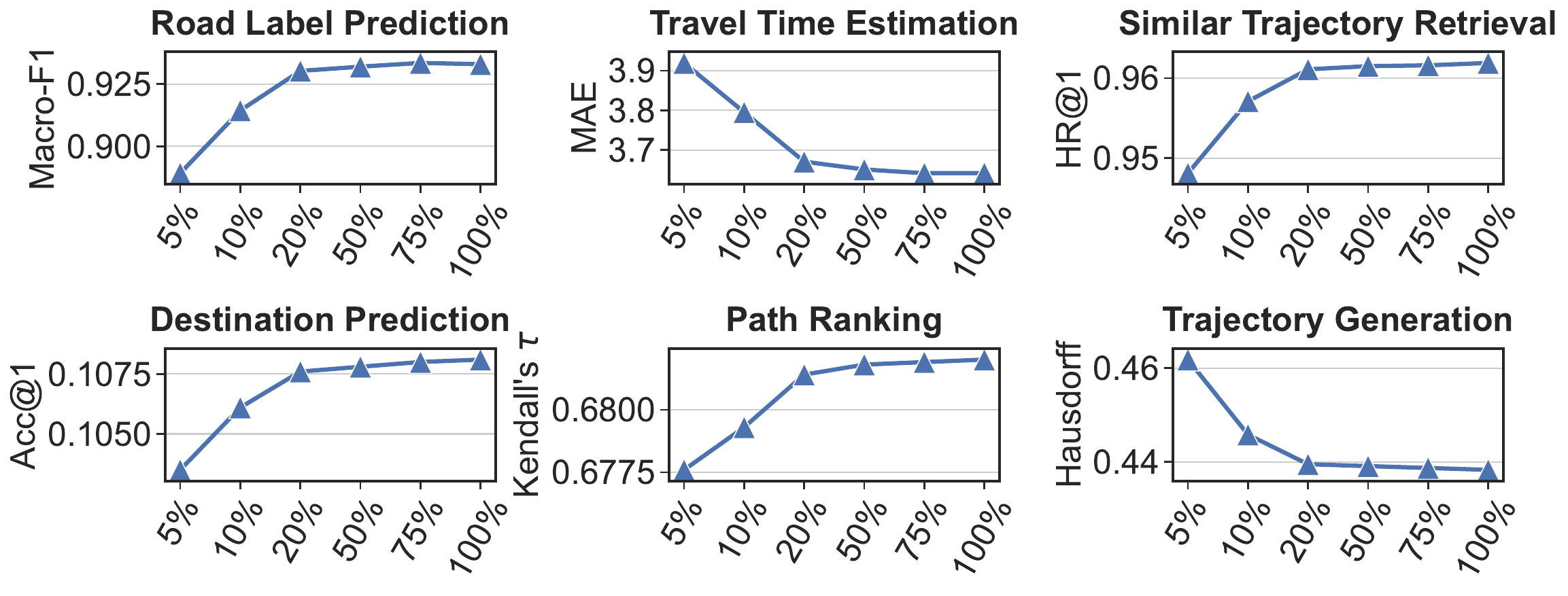}
  \caption{Impact of the critical segment selection ratio $\eta$ on the Beijing dataset (refer to \cref{subsec:appendix_hyperparameter_sensitivity} for other datasets).}
  \label{fig:llm_ratio_beijing}
\end{figure}

In our Environment Perception Module, two hyperparameters jointly determine the spatial scale of semantic perception: (1) the POI perception radius $\delta$ (default \SI{100}{m}) for fine-grained semantic modeling, and (2) the grid side length $L$ (default \SI{1000}{m}) for coarse-grained semantic modeling. Although these hyperparameters modulate the perception granularity, a robust model should retain stable performance over a reasonable range of their values. To assess this robustness, we vary $\delta$ within \{\SI{50}{m}, \SI{100}{m}, \SI{200}{m}\} and $L$ within \{\SI{500}{m}, \SI{1000}{m}, \SI{2000}{m}\}. As shown in \cref{fig:hyperparameter_beijing}, CORE consistently exhibits strong performance across all configurations, indicating that it is not sensitive to these hyperparameter choices.

We further evaluate the effectiveness of restricting fine-grained LLM-based semantic modeling to high-traffic \emph{critical segments}. We conduct a sensitivity analysis on the selection ratio $\eta \in \{5\%, 10\%, 20\%, 50\%, 75\%, 100\%\}$. As shown in \cref{fig:llm_ratio_beijing}, increasing $\eta$ from $5\%$ to $20\%$ yields substantial gains across all downstream tasks, indicating that fine-grained environmental semantic modeling on frequently visited segments is crucial for capturing key behavioral patterns. However, further increasing $\eta$ beyond $20\%$ leads to saturated improvements. This observation is consistent with the long-tailed distribution of urban road-network visitation, where a small subset of core segments accounts for most traffic interactions and semantic information, suggesting that designating the top $20\%$ of segments as \emph{critical segments} offers a favorable trade-off between performance and computational cost.

\subsubsection{Transformer Capacity Sensitivity}

\cref{tab:transformer_capacity_sensitivity} reports one representative metric per task to assess CORE's sensitivity to Transformer encoder capacity. We vary the encoder depth over $\{2,4,6,8\}$ with the hidden dimensionality fixed at 128, and the hidden dimensionality over $\{64,128,256\}$ with the encoder depth fixed at 6. Overall, CORE maintains comparable performance across the tested configurations, suggesting that it is not strongly tied to a specific encoder capacity. Nevertheless, the moderate setting generally provides a more favorable balance: smaller configurations may not offer sufficient representational capacity, whereas excessive capacity yields only marginal or inconsistent gains due to overfitting risk.

\begin{table}[!t]
\caption{Transformer capacity sensitivity on the Beijing dataset (refer to \cref{subsec:appendix_transformer_capacity_sensitivity} for other datasets).}
\label{tab:transformer_capacity_sensitivity}
\centering
\resizebox{1.0\linewidth}{!}{
\begin{tabular}{lccccccc}
  \toprule
  \textbf{Factor} & \textbf{Setting} & \textbf{Macro-F1}$\uparrow$ & \textbf{MAE}$\downarrow$ & \textbf{HR@1}$\uparrow$ & \textbf{Acc@1}$\uparrow$ & \textbf{Kendall's $\tau$}$\uparrow$ & \textbf{Hausdorff}$\downarrow$ \\
  \midrule
  \multirow{4}{*}{Encoder Layers} & 2 & \textbf{0.9315} & 3.7621 & 0.9516 & 0.1028 & 0.6745 & 0.4410 \\
  & 4 & 0.9287 & 3.7010 & 0.9594 & 0.1016 & 0.6784 & 0.4399 \\
  & 6 & \underline{0.9303} & \textbf{3.6703} & \textbf{0.9611} & \textbf{0.1076} & \underline{0.6814} & \textbf{0.4395} \\
  & 8 & 0.9156 & \underline{3.6957} & \underline{0.9606} & \underline{0.1073} & \textbf{0.6822} & \underline{0.4397} \\
  \cmidrule(lr){1-8}
  \multirow{3}{*}{Hidden Dimension} & 64 & 0.8897 & 3.6851 & 0.9495 & 0.1039 & 0.6732 & 0.4403 \\
  & 128 & \underline{0.9303} & \textbf{3.6703} & \textbf{0.9611} & \textbf{0.1076} & \textbf{0.6814} & \textbf{0.4395} \\
  & 256 & \textbf{0.9350} & \underline{3.6711} & \underline{0.9587} & \underline{0.1062} & \underline{0.6809} & \underline{0.4400} \\
  \bottomrule
\end{tabular}
}
\end{table}

\subsubsection{LLM Backbone Sensitivity}

\begin{table}[!t]
\caption{Performance of CORE with different LLM backbones on the Beijing dataset (refer to \cref{subsec:appendix_llm_backbone_sensitivity} for other datasets).}
\label{tab:llm_sensitivity}
\centering
\resizebox{1.0\linewidth}{!}{
\begin{tabular}{l cccccc}
  \toprule
  \textbf{LLM Backbone} & \textbf{Macro-F1}$\uparrow$ & \textbf{MAE}$\downarrow$ & \textbf{HR@1}$\uparrow$ & \textbf{Acc@1}$\uparrow$ & \textbf{Kendall's $\tau$}$\uparrow$ & \textbf{Hausdorff}$\downarrow$ \\
  \midrule
  CORE w/ Qwen3-0.6B & 0.8872 & 3.7881 & 0.9498 & 0.1039 & 0.6781 & 0.4563 \\
  CORE w/ Qwen3-8B & \underline{0.9303} & \textbf{3.6703} & \underline{0.9611} & \textbf{0.1076} & \textbf{0.6814} & \underline{0.4395} \\
  CORE w/ DeepSeek-V3.2-Exp & \textbf{0.9312} & \underline{3.6706} & \textbf{0.9614} & \underline{0.1075} & \underline{0.6813} & \textbf{0.4393} \\
  \bottomrule
\end{tabular}
}
\end{table}

We further examine the sensitivity of CORE to the choice of LLM backbone. Following \cref{tab:transformer_capacity_sensitivity}, we report one representative metric for each downstream task in \cref{tab:llm_sensitivity}. When we scale down the backbone from Qwen3-8B to Qwen3-0.6B, the performance consistently decreases across all metrics, indicating that an overly small LLM is insufficient to capture the required semantic reasoning over POI distributions. In contrast, replacing Qwen3-8B with DeepSeek-V3.2-Exp, a substantially larger model with 671B parameters, brings almost no performance improvement across all metrics, suggesting that a medium-sized LLM is sufficient and strikes a favorable balance between accuracy and computational cost in our setting.

\subsubsection{POI Completeness Sensitivity}

\begin{figure}[!t]
  \centering
  \includegraphics[width=1.0\linewidth]{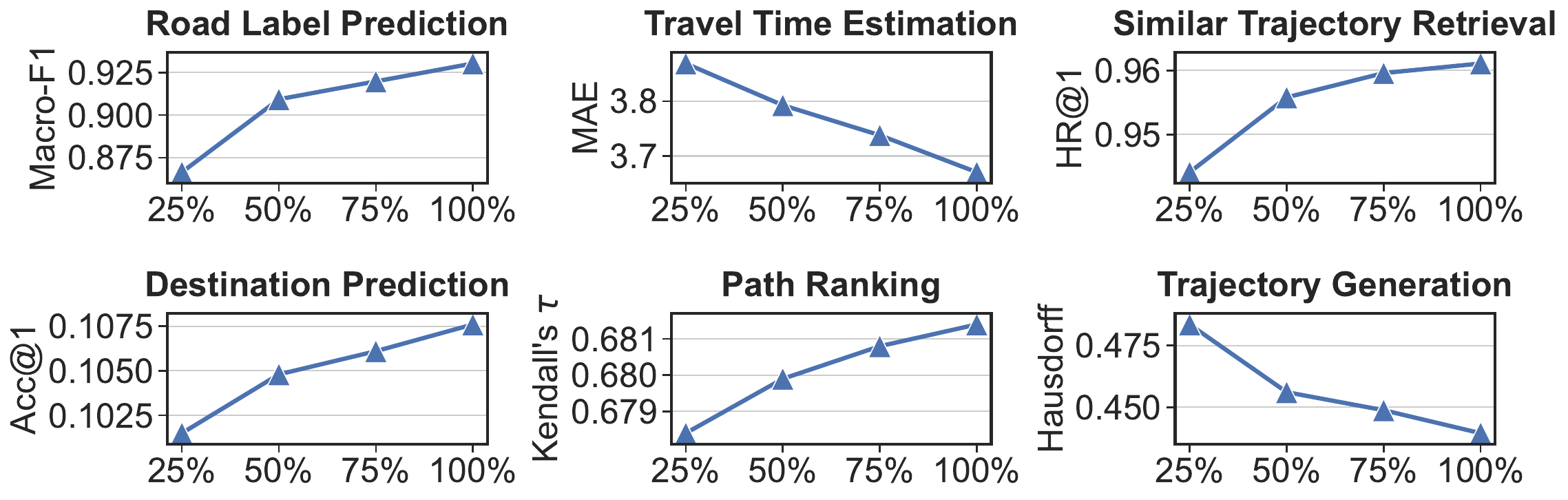}
  \caption{Sensitivity of CORE to POI completeness on the Beijing dataset (refer to \cref{subsec:appendix_poi_ratio} for other datasets).}
  \label{fig:poi_ratio_beijing}
\end{figure}

In practical applications, limited POI sources and collection constraints often lead to incomplete or sparse POI data. To assess the robustness of CORE under such conditions, we randomly subsample the original POI database with $\gamma \in \{25\%, 50\%, 75\%, 100\%\}$ and retrain the model. As shown in \cref{fig:poi_ratio_beijing}, CORE’s performance decreases as POI completeness declines, confirming its dependence on environmental data richness. Nevertheless, CORE does not undergo catastrophic failure even under severe sparsity (e.g., with only 25\% of POIs). This indicates that although missing POI data hinders the precise extraction of environmental semantics, the LLM-based module leverages its broad world knowledge to infer reasonable semantic approximations from sparse clues, thereby partially mitigating the impact of data sparsity on performance.

\section{Conclusion}

In this paper, we propose CORE, a TRL framework that injects context-aware route choice semantics into trajectory representations. CORE leverages an LLM-based Environment Perception Module to construct a context-enriched road network and employs an MoE-based Route Choice Encoder to capture complex routing patterns. Extensive experiments on four real-world datasets demonstrate that CORE consistently outperforms \emph{state-of-the-art} baselines while exhibiting superior data and computational efficiency.

Despite these strengths, CORE is currently limited to single-city modeling and therefore cannot directly transfer the learned semantics to other urban areas without retraining. In future work, we plan to develop a trajectory foundation model. By distilling universal knowledge about the interactions between environment and behavior from large-scale data, we aim to achieve robust generalization and enable zero-shot transfer to previously unseen urban environments.

\bibliographystyle{IEEEtran}
\bibliography{ref}

@string(ICLR = "{International Conference on Learning Representations}")

@string(NeurIPS = "{Annual Conference on Neural Information Processing Systems}")

@string(ICML = "{International Conference on Machine Learning}")

@string(AAAI = "{AAAI Conference on Artificial Intelligence}")

@string(IJCAI = "{International Joint Conference on Artificial Intelligence}")

@string(SIGMOD = "{ACM SIGMOD Conference}")

@string(VLDB = "{International Conference on Very Large Data Bases}")

@string(ICDE = "{IEEE International Conference on Data Engineering}")

@string(SIGKDD = "{ACM SIGKDD Conference on Knowledge Discovery and Data Mining}")

@string(CIKM = "{ACM International Conference on Information and Knowledge Management}")

@string(ACL = "{Annual Meeting of the Association for Computational Linguistics}")

@string(EMNLP = "{Conference on Empirical Methods in Natural Language Processing}")

@string(NAACL = "{North American Chapter of the Association for Computational Linguistics}")

@string(SIGIR = "{International ACM SIGIR Conference on Research and Development in Information Retrieval}")

@string(WWW = "{International World Wide Web Conference}")

@string(SIGSPATIAL = "{ACM Special Interest Group on Spatial Information}")

@string(EDBT = "{International Conference on Extending Database Technology}")

@string(ICLR = "{ICLR}")

@string(NeurIPS = "{NeurIPS}")

@string(ICML = "{ICML}")

@string(AAAI = "{AAAI}")

@string(IJCAI = "{IJCAI}")

@string(SIGMOD = "{SIGMOD}")

@string(VLDB = "{VLDB}")

@string(ICDE = "{ICDE}")

@string(SIGKDD = "{SIGKDD}")

@string(CIKM = "{CIKM}")

@string(ACL = "{ACL}")

@string(EMNLP = "{EMNLP}")

@string(NAACL = "{NAACL}")

@string(SIGIR = "{SIGIR}")

@string(WWW = "{WWW}")

@string(SIGSPATIAL = "{SIGSPATIAL}")

@string(EDBT = "{EDBT}")

@string(CSUR = "{ACM Computing Surveys}")

@string(TKDE = "{IEEE Transactions on Knowledge and Data Engineering}")

@string(TMC = "{IEEE Transactions on Mobile Computing}")

@string(TITS = "{IEEE Transactions on Intelligent Transportation Systems}")

@string(TNNLS = "{IEEE Transactions on Neural Networks and learning systems}")

@string(TIST = "{ACM Transactions on Intelligent Systems and Technology}")

@string(IJGIS = "{International Journal of Geographical Information Science}")

@string(TKDD = "{ACM Transactions on Knowledge Discovery from Data}")

@string(INFFUS = "{Information Fusion}")

@string(JMLR = "{Journal of Machine Learning Research}")

@string(CSUR = "{ACM Comput. Surv.}")

@string(TKDE = "{IEEE Trans. Knowl. Data Eng.}")

@string(TMC = "{IEEE Trans. Mob. Comput.}")

@string(TITS = "{IEEE Trans. Intell. Transp. Syst.}")

@string(TNNLS = "{IEEE Trans. Neural Netw. Learn. Syst.}")

@string(TIST = "{ACM Trans. Intell. Syst. Technol.}")

@string(IJGIS = "{Int. J. Geogr. Inf. Sci.}")

@string(TKDD = "{ACM Trans. Knowl. Discov. Data}")

@string(INFFUS = "{Inf. Fusion}")

@string(JMLR = "{J. Mach. Learn. Res.}")

@article{wang2021survey,
  author = {Wang, Sheng and Bao, Zhifeng and Culpepper, J Shane and Cong, Gao},
  journal = CSUR,
  title = {A survey on trajectory data management, analytics, and learning},
  year = {2021},
  volume = {54},
  number = {2},
  pages = {1--36},
}

@article{wang2022deep,
  author = {Wang, Senzhang and Cao, Jiannong and Yu, Philip S.},
  journal = TKDE,
  title = {Deep learning for spatio-temporal data mining: A survey},
  year = {2022},
  volume = {34},
  number = {8},
  pages = {3681--3700},
}

@article{zheng2015trajectory,
  author = {Zheng, Yu},
  journal = TIST,
  title = {Trajectory data mining: an overview},
  year = {2015},
  volume = {6},
  number = {3},
  pages = {1--41},
}

@article{hu2024spatio,
  author = {Hu, Danlei and Chen, Lu and Fang, Hanxi and Fang, Ziquan and Li, Tianyi and Gao, Yunjun},
  journal = TKDE,
  title = {Spatio-temporal trajectory similarity measures: A comprehensive survey and quantitative study},
  year = {2024},
  volume = {36},
  number = {5},
  pages = {2191--2212},
}

@inproceedings{zhou2025grid,
  author = {Zhou, Silin and Shang, Shuo and Chen, Lisi and Han, Peng and Jensen, Christian S.},
  booktitle = SIGKDD,
  title = {Grid and road expressions are complementary for trajectory representation learning},
  year = {2025}
}

@inproceedings{jiang2023self,
  author = {Jiang, Jiawei and Pan, Dayan and Ren, Houxing and Jiang, Xiaohan and Li, Chao and Wang, Jingyuan},
  booktitle = ICDE,
  title = {Self-supervised trajectory representation learning with temporal regularities and travel semantics},
  year = {2023}
}

@inproceedings{yang2023lightpath,
  author = {Yang, Sean Bin and Hu, Jilin and Guo, Chenjuan and Yang, Bin and Jensen, Christian S.},
  booktitle = SIGKDD,
  title = {{LightPath}: Lightweight and scalable path representation learning},
  year = {2023}
}

@inproceedings{mao2022jointly,
  author = {Mao, Zhenyu and Li, Ziyue and Li, Dedong and Bai, Lei and Zhao, Rui},
  booktitle = CIKM,
  title = {Jointly contrastive representation learning on road network and trajectory},
  year = {2022}
}

@inproceedings{yang2021unsupervised,
  author = {Yang, Sean Bin and Guo, Chenjuan and Hu, Jilin and Tang, Jian and Yang, Bin},
  booktitle = IJCAI,
  title = {Unsupervised path representation learning with curriculum negative sampling},
  year = {2021}
}

@inproceedings{yao2019computing,
  author = {Yao, Di and Cong, Gao and Zhang, Chao and Bi, Jingping},
  booktitle = ICDE,
  title = {Computing trajectory similarity in linear time: A generic seed-guided neural metric learning approach},
  year = {2019}
}

@inproceedings{li2018deep,
  author = {Li, Xiucheng and Zhao, Kaiqi and Cong, Gao and Jensen, Christian S. and Wei, Wei},
  booktitle = ICDE,
  title = {Deep representation learning for trajectory similarity computation},
  year = {2018}
}

@article{fu2020trembr,
  author = {Fu, Tao-Yang and Lee, Wang-Chien},
  journal = TIST,
  title = {{Trembr}: Exploring road networks for trajectory representation learning},
  year = {2020},
  volume = {11},
  number = {1},
  pages = {1--25},
}

@inproceedings{chen2021robust,
  author = {Chen, Yile and others},
  booktitle = CIKM,
  title = {Robust road network representation learning: When traffic patterns meet traveling semantics},
  year = {2021}
}

@inproceedings{devlin2019bert,
  author = {Devlin, Jacob and Chang, Ming-Wei and Lee, Kenton and Toutanova, Kristina},
  booktitle = NAACL,
  title = {{BERT}: Pre-training of deep bidirectional transformers for language understanding},
  year = {2019}
}

@inproceedings{han2025bridging,
  author = {Han, Chengkai and others},
  booktitle = AAAI,
  title = {Bridging traffic state and trajectory for dynamic road network and trajectory representation learning},
  year = {2025}

}

@article{prato2009route,
  author = {Prato, Carlo Giacomo},
  journal = {Journal of Choice Modelling},
  title = {Route choice modeling: past, present and future research directions},
  year = {2009},
  volume = {2},
  number = {1},
  pages = {65--100},
}

@inproceedings{wu2020learning,
  author = {Wu, Ning and Zhao, Xin Wayne and Wang, Jingyuan and Pan, Dayan},
  booktitle = SIGKDD,
  title = {Learning effective road network representation with hierarchical graph neural networks},
  year = {2020}
}

@inproceedings{chen2020simple,
  author = {Chen, Ting and Kornblith, Simon and Norouzi, Mohammad and Hinton, Geoffrey},
  booktitle = ICML,
  title = {A simple framework for contrastive learning of visual representations},
  year = {2020}
}

@article{chen2025semantic,
  author = {Chen, Yile and Li, Xiucheng and Cong, Gao and Bao, Zhifeng and Long, Cheng},
  journal = TMC,
  title = {Semantic-enhanced representation learning for road networks with temporal dynamics},
  year = {2025},
  volume = {24},
  number = {10},
  pages = {9413-9427},
}

@article{yang2025qwen3,
  author = {Yang, An and others},
  journal = {arXiv preprint arXiv:2505.09388},
  title = {{Qwen3} technical report},
  year = {2025}
}

@article{zhang2025qwen3,
  author = {Zhang, Yanzhao and others},
  journal = {arXiv preprint arXiv:2506.05176},
  title = {{Qwen3 Embedding}: Advancing Text Embedding and Reranking Through Foundation Models},
  year = {2025}
}

@article{boeing2025modeling,
  author = {Boeing, Geoff},
  journal = {Geogr. Anal.},
  title = {Modeling and analyzing urban networks and amenities with {OSMnx}},
  year = {2025},
  volume = {57},
  number = {4},
  pages = {567--577},
}

@article{yang2018fast,
  author = {Yang, Can and Gidofalvi, Gyozo},
  journal = IJGIS,
  title = {Fast map matching, an algorithm integrating hidden {Markov} model with precomputation},
  year = {2018},
  volume = {32},
  number = {3},
  pages = {547--570},
}

@inproceedings{velivckovic2018graph,
  author = {Veli{\v{c}}kovi{\'c}, Petar and Cucurull, Guillem and Casanova, Arantxa and Romero, Adriana and Lio, Pietro and Bengio, Yoshua},
  booktitle = ICLR,
  title = {Graph attention networks},
  year = {2018}
}

@inproceedings{han2021graph,
  author = {Han, Peng and Wang, Jin and Yao, Di and Shang, Shuo and Zhang, Xiangliang},
  booktitle = SIGKDD,
  title = {A graph-based approach for trajectory similarity computation in spatial networks},
  year = {2021}
}

@inproceedings{grover2016node2vec,
  author = {Grover, Aditya and Leskovec, Jure},
  booktitle = SIGKDD,
  title = {{node2vec}: Scalable feature learning for networks},
  year = {2016}
}

@inproceedings{vaswani2017attention,
  author = {Vaswani, Ashish and others},
  booktitle = NeurIPS,
  title = {Attention is all you need},
  year = {2017}
}

@inproceedings{ouyang2022training,
  author = {Ouyang, Long and others},
  booktitle = NeurIPS,
  title = {Training language models to follow instructions with human feedback},
  year = {2022}
}

@inproceedings{jiawei2024large,
  author = {Wang, Jiawei and others},
  booktitle = NeurIPS,
  title = {Large language models as urban residents: An {LLM} agent framework for personal mobility generation},
  year = {2024}
}

@article{shao2024chain,
  author = {Shao, Chenyang and others},
  journal = {arXiv preprint arXiv:2402.09836},
  title = {Chain-of-planned-behaviour workflow elicits few-shot mobility generation in {LLMs}},
  year = {2024}
}

@inproceedings{feng2025agentmove,
  author = {Feng, Jie and Du, Yuwei and Zhao, Jie and Li, Yong},
  booktitle = NAACL,
  title = {{AgentMove}: A large language model based agentic framework for zero-shot next location prediction},
  year={2025}
}

@inproceedings{gong2024mobility,
  author = {Gong, Letian and others},
  booktitle = NeurIPS,
  title = {{Mobility-LLM}: Learning visiting intentions and travel preference from human mobility data with large language models},
  year = {2024}
}

@inproceedings{li2024large,
  author = {Li, Peibo and de Rijke, Maarten and Xue, Hao and Ao, Shuang and Song, Yang and Salim, Flora D},
  booktitle = SIGIR,
  title = {Large language models for next point-of-interest recommendation},
  year = {2024}
}

@inproceedings{cheng2025poi,
  author = {Cheng, Jiawei and others},
  booktitle = AAAI,
  title = {{POI-Enhancer}: An {LLM}-based semantic enhancement framework for {POI} representation learning},
  year = {2025}
}

@inproceedings{zhang2023promptst,
  author = {Zhang, Zijian and others},
  booktitle = CIKM,
  title = {{PromptST}: Prompt-enhanced spatio-temporal multi-attribute prediction},
  year = {2023}
}

@inproceedings{li2024urbangpt,
  author = {Li, Zhonghang and others},
  booktitle = SIGKDD,
  title = {{UrbanGPT}: Spatio-Temporal Large Language Models},
  year = {2024}
}

@inproceedings{yuan2024unist,
  author = {Yuan, Yuan and Ding, Jingtao and Feng, Jie and Jin, Depeng and Li, Yong},
  booktitle = SIGKDD,
  title = {{UniST}: {A} Prompt-Empowered Universal Model for Urban Spatio-Temporal Prediction},
  year = {2024}
}

@inproceedings{wang2018learning,
  author = {Wang, Zheng and Fu, Kun and Ye, Jieping},
  booktitle = SIGKDD,
  title = {Learning to estimate the travel time},
  year = {2018}
}

@inproceedings{fang2020constgat,
  author = {Fang, Xiaomin and Huang, Jizhou and Wang, Fan and Zeng, Lingke and Liang, Haijin and Wang, Haifeng},
  booktitle = SIGKDD,
  title = {{ConSTGAT}: Contextual spatial-temporal graph attention network for travel time estimation at {Baidu Maps}},
  year = {2020}
}

@article{yang2022context,
  author = {Yang, Sean Bin and Guo, Chenjuan and Yang, Bin},
  journal = TKDE,
  title = {Context-aware path ranking in road networks},
  year = {2022},
  volume = {34},
  number = {7},
  pages = {3153--3168},
}

@article{kendall1938new,
  title = {A new measure of rank correlation},
  author = {Kendall, Maurice G},
  journal = {Biometrika},
  year = {1938},
  volume = {30},
  number = {1-2},
  pages = {81-93},
}

@article{spearman1961proof,
  title = {The proof and measurement of association between two things},
  author = {Spearman, Charles},
  journal = {Am. J. Psychol},
  year = {1904},
  volume = {15},
  number = {1},
  pages = {72--101},
}

@article{hochreiter1997long,
  title = {Long short-term memory},
  author = {Hochreiter, Sepp and Schmidhuber, J{\"u}rgen},
  journal = {Neural Comput.},
  year = {1997},
  volume = {9},
  number = {8},
  pages = {1735--1780},
}

@inproceedings{cho2014learning,
  author = {Cho, Kyunghyun and others},
  booktitle = EMNLP,
  title = {Learning phrase representations using {RNN} encoder-decoder for statistical machine translation},
  year = {2014}
}

@inproceedings{dai2024deepseekmoe,
  author = {Dai, Damai and others},
  booktitle = ACL,
  title = {{DeepSeekMoE}: Towards ultimate expert specialization in mixture-of-experts language models},
  year = {2024}
}

@article{james2023citywide,
  author = {Yu, James J.Q.},
  journal = TKDE,
  title = {Citywide estimation of travel time distributions with Bayesian deep graph learning},
  year = {2023},
  volume = {35},
  number = {3},
  pages = {2366--2378},
}

@article{zhang2023beyond,
  author = {Zhang, Chao and Zhao, Kai and Chen, Meng},
  journal = TKDE,
  title = {Beyond the limits of predictability in human mobility prediction: Context-transition predictability},
  year = {2023},
  volume = {35},
  number = {5},
  pages = {4514--4526},
}

@inproceedings{wang2018will,
  author = {Wang, Dong and Zhang, Junbo and Cao, Wei and Li, Jian and Zheng, Yu},
  booktitle = AAAI,
  title = {When will you arrive? Estimating travel time based on deep neural networks},
  year = {2018}
}

@inproceedings{zhou2025red,
  author = {Zhou, Silin and Shang, Shuo and Chen, Lisi and Jensen, Christian S and Kalnis, Panos},
  booktitle = VLDB,
  title = {{RED}: Effective trajectory representation learning with comprehensive information},
  year = {2025}
}

@inproceedings{ma2024more,
  author = {Ma, Zhipeng and others},
  booktitle = WWW,
  title = {More than routing: Joint {GPS} and route modeling for refine trajectory representation learning},
  year = {2024}
}

@book{hillier1989social,
  author = {Hillier, Bill and Hanson, Julienne},
  title = {The social logic of space},
  publisher = {Cambridge University Press},
  year = {1989}
}

@inproceedings{wang2019learning,
  author = {Wang, Meng-xiang and Lee, Wang-Chien and Fu, Tao-yang and Yu, Ge},
  booktitle = SIGSPATIAL,
  title = {Learning embeddings of intersections on road networks},
  year = {2019}
}

@article{wang2020representation,
  author = {Wang, Meng-Xiang and Lee, Wang-Chien and Fu, Tao-Yang and Yu, Ge},
  journal = TIST,
  title = {On representation learning for road networks},
  year = {2020},
  volume = {12},
  number = {1},
  pages = {1--27},
}

@article{wu2020comprehensive,
  author = {Wu, Zonghan and Pan, Shirui and Chen, Fengwen and Long, Guodong and Zhang, Chengqi and Yu, Philip S.},
  journal = TNNLS,
  title = {A comprehensive survey on graph neural networks},
  year = {2021},
  volume = {32},
  number = {1},
  pages = {4--24},
}

@article{jepsen2020relational,
  author = {Jepsen, Tobias Skovgaard and Jensen, Christian S and Nielsen, Thomas Dyhre},
  journal = TITS,
  title = {Relational fusion networks: Graph convolutional networks for road networks},
  year = {2022},
  volume = {23},
  number = {1},
  pages = {418--429},
}

@inproceedings{kipf2017semi,
  author = {Kipf, Thomas N. and Welling, Max},
  booktitle = ICLR,
  title = {Semi-supervised classification with graph convolutional networks},
  year = {2017}
}

@inproceedings{chang2023spatial,
  author = {Chang, Yanchuan and Tanin, Egemen and Cao, Xin and Qi, Jianzhong},
  booktitle = {EDBT},
  title = {Spatial Structure-Aware Road Network Embedding via Graph Contrastive Learning},
  year = {2023}
}

@article{zhang2023road,
  author = {Zhang, Liang and Long, Cheng},
  journal = TKDD,
  title = {Road network representation learning: A dual graph-based approach},
  year = {2023},
  volume = {17},
  number = {9},
  pages={1--25}
}

@article{antelmi2023survey,
  author = {Antelmi, Alessia and Cordasco, Gennaro and Polato, Mirko and Scarano, Vittorio and Spagnuolo, Carmine and Yang, Dingqi},
  journal = CSUR,
  title = {A survey on hypergraph representation learning},
  year = {2023},
  volume = {56},
  number = {1},
  pages={1--38}
}

@inproceedings{zhu2021graph,
  author = {Zhu, Yanqiao and Xu, Yichen and Yu, Feng and Liu, Qiang and Wu, Shu and Wang, Liang},
  booktitle = WWW,
  title = {Graph contrastive learning with adaptive augmentation},
  year = {2021}
}

@article{harris1945nature,
  author = {Harris, Chauncy D and Ullman, Edward L},
  journal = {The annals of the American academy of political and social science},
  title = {The nature of cities},
  year = {1945},
  volume = {242},
  number = {1},
  pages = {7--17},
}

@inproceedings{yuan2012discovering,
  author = {Yuan, Jing and Zheng, Yu and Xie, Xing},
  booktitle = SIGKDD,
  title = {Discovering regions of different functions in a city using human mobility and {POIs}},
  year = {2012}
}

@inproceedings{jiang2023continuous,
  author = {Jiang, Wenjun and Zhao, Wayne Xin and Wang, Jingyuan and Jiang, Jiawei},
  booktitle = AAAI,
  title = {Continuous trajectory generation based on two-stage {GAN}},
  year = {2023}
}

@article{wang2024spatiotemporal,
  author = {Wang, Yu and Cao, Ji and Huang, Wenjie and Liu, Zhihua and Zheng, Tongya and Song, Mingli},
  journal = INFFUS,
  title = {Spatiotemporal gated traffic trajectory simulation with semantic-aware graph learning},
  year = {2024},
  volume = {108},
  pages = {102404},
}

@inproceedings{cao2025holistic,
  author = {Cao, Ji and others},
  booktitle = AAAI,
  title = {Holistic Semantic Representation for Navigational Trajectory Generation},
  year = {2025}
}

@inproceedings{xie2017distributed,
  author = {Xie, Dong and Li, Feifei and Phillips, Jeff M.},
  booktitle = VLDB,
  title = {Distributed trajectory similarity search},
  year = {2017}
}

@inproceedings{keogh2002exact,
  author = {Keogh, Eamonn},
  booktitle = VLDB,
  title = {Exact Indexing of Dynamic Time Warping},
  year = {2002},
}

@inproceedings{chen2005robust,
  author = {Chen, Lei and {\"O}zsu, M Tamer and Oria, Vincent},
  booktitle = SIGMOD,
  title = {Robust and fast similarity search for moving object trajectories},
  year = {2005}
}

@inproceedings{loshchilov2019decoupled,
  author = {Loshchilov, Ilya and Hutter, Frank},
  booktitle = ICLR,
  title = {Decoupled weight decay regularization},
  year = {2019}
}

@article{wang2024auxiliary,
  author = {Wang, Lean and Gao, Huazuo and Zhao, Chenggang and Sun, Xu and Dai, Damai},
  journal = {arXiv preprint arXiv:2408.15664},
  title = {Auxiliary-loss-free load balancing strategy for mixture-of-experts},
  year = {2024}
}

@article{maaten2008visualizing,
  author = {Maaten, Laurens van der and Hinton, Geoffrey},
  journal = JMLR,
  title = {Visualizing data using {t-SNE}},
  year = {2008},
  volume = {9},
  number = {86},
  pages = {2579--2605},
}

@inproceedings{zhou2025trajcogn,
  author = {Zhou, Zeyu and others},
  booktitle = IJCAI,
  title = {{TrajCogn}: Leveraging {LLMs} for Cognizing Movement Patterns and Travel Purposes from Trajectories},
  year = {2025}
}

@article{lima2016understanding,
  author = {Lima, Antonio and Stanojevic, Rade and Papagiannaki, Dina and Rodriguez, Pablo and Gonz{\'a}lez, Marta C},
  journal = {Journal of The Royal Society Interface},
  title = {Understanding individual routing behaviour},
  year = {2016},
  volume = {13},
  number = {116},
  pages = {20160021},
}

@inproceedings{wei2025path,
  author = {Wei, Yongfu and Lin, Yan and Gao, Hongfan and Xu, Ronghui and Yang, Sean Bin and Hu, Jilin},
  booktitle = WWW,
  title = {{Path-LLM}: A multi-modal path representation learning by aligning and fusing with large language models},
  year = {2025}
}

@inproceedings{liu2025trajmamba,
  author = {Liu, Yichen and Lin, Yan and Guo, Shengnan and Zhou, Zeyu and Lin, Youfang and Wan, Huaiyu},
  booktitle = NeurIPS,
  title = {{TrajMamba}: An Efficient and Semantic-rich Vehicle Trajectory Pre-training Model},
  year = {2025}
}

@inproceedings{dao2024transformers,
  author = {Dao, Tri and Gu, Albert},
  booktitle = ICML,
  title = {Transformers are {SSMs}: Generalized models and efficient algorithms through structured state space duality},
  year = {2024}
}

@inproceedings{maynez2020faithfulness,
  author = {Maynez, Joshua and Narayan, Shashi and Bohnet, Bernd and McDonald, Ryan},
  booktitle = ACL,
  title = {On faithfulness and factuality in abstractive summarization},
  year = {2020}
}

@inproceedings{kryscinski2020evaluating,
  author = {Kry{\'s}ci{\'n}ski, Wojciech and McCann, Bryan and Xiong, Caiming and Socher, Richard},
  booktitle = EMNLP,
  title = {Evaluating the factual consistency of abstractive text summarization},
  year = {2020}
}

@inproceedings{pagnoni2021understanding,
  author = {Pagnoni, Artidoro and Balachandran, Vidhisha and Tsvetkov, Yulia},
  booktitle = NAACL,
  title = {Understanding factuality in abstractive summarization with {FRANK}: A benchmark for factuality metrics},
  year = {2021}
}

@article{dai2025learning,
  author = {Dai, Junshu and Wang, Yu and Zheng, Tongya and Ji, Wei and Guo, Qinghong and Cao, Ji and Song, Jie and Jin, Canghong and Song, Mingli},
  journal = {arXiv preprint arXiv:2512.22605},
  title = {Learning Multi-Modal Mobility Dynamics for Generalized Next Location Recommendation},
  year = {2025}
}

@inproceedings{wang2026adaptive,
  author = {Wang, Yu and Dai, Junshu and Ying, Yuchen and Yuan, Hanyang and Feng, Zunlei and Zheng, Tongya and Song, Mingli},
  booktitle = WWW,
  title = {Adaptive location hierarchy learning for long-tailed mobility prediction},
  year = {2026}
}

@inproceedings{guo2026dual,
  author = {Guo, Qinghong and Wang, Yu and Cao, Ji and Zheng, Tongya and Dai, Junshu and Hu, Bingde and Liu, Shunyu and Jin, Canghong},
  booktitle = AAAI,
  title = {Dual-branch Spatial-Temporal Self-supervised Representation for Enhanced Road Network Learning},
  year = {2026}
}

@inproceedings{feng2020learning,
  author = {Feng, Jie and Yang, Zeyu and Xu, Fengli and Yu, Haisu and Wang, Mudan and Li, Yong},
  booktitle = SIGKDD,
  title = {Learning to simulate human mobility},
  year = {2020}
}

@article{wang2024star,
  author = {Wang, Yu and Zheng, Tongya and Liu, Shunyu and Feng, Zunlei and Chen, Kaixuan and Hao, Yunzhi and Song, Mingli},
  journal = TKDE,
  title = {Spatiotemporal-augmented graph neural networks for human mobility simulation},
  year = {2024},
  volume = {36},
  number = {11},
  pages = {7074--7086}
}

\clearpage

\appendices
\onecolumn
\crefalias{section}{appendix}
\crefalias{subsection}{appendix}
\crefalias{subsubsection}{appendix}
{\centering
  \Huge{\scshape Supplementary Material\par}
}

\vspace{1em}

{\centering
  \large{\scshape Capturing Context-Aware Route Choice Semantics\\for Trajectory Representation Learning\par}
}

\vspace{1em}

\section{Notation}

For clarity, \cref{tab:notation_basic_definitions,tab:notation_environment_perception,tab:notation_route_choice_encoder,tab:notation_trajectory_embedding} summarize the main notations used throughout the paper.

\begin{table}[H]
\caption{Notation for basic definitions and problem setup.}
\label{tab:notation_basic_definitions}
\centering
\renewcommand{\arraystretch}{1.15}
\begin{tabular}{ll}
  \toprule
  \textbf{Symbol} & \textbf{Meaning} \\
  \midrule
  $\mathcal{G}=\langle\mathcal{V},\mathcal{E}\rangle$ & Directed road network with road segments $\mathcal{V}$ and edges $\mathcal{E}$ \\
  $\mathcal{N}(r)$ & Outgoing neighbor set of segment $r$ \\
  $\mathcal{T}^{\text{GPS}}$ & Raw GPS trajectory \\
  $\boldsymbol{\tau}^{\text{GPS}}_i=(\text{lat}_i, \text{lon}_i, t_i)$ & $i$-th GPS point in $\mathcal{T}^{\text{GPS}}$, with latitude $\text{lat}_i$, longitude $\text{lon}_i$, and timestamp $t_i$ \\
  $\mathcal{T}$ & Road-network-constrained trajectory \\
  $\boldsymbol{\tau}_i=(r_i, t_i)$ & $i$-th point in $\mathcal{T}$, where $r_i\in\mathcal{V}$ is the matched road segment and $t_i$ is the timestamp \\
  $\mathbf{z}$ & Trajectory embedding \\
  $d$ & Trajectory embedding dimension \\
  \bottomrule
\end{tabular}
\end{table}

\begin{table}[H]
\caption{Notation for environment perception module.}
\label{tab:notation_environment_perception}
\centering
\renewcommand{\arraystretch}{1.15}
\begin{tabular}{ll}
  \toprule
  \textbf{Symbol} & \textbf{Meaning} \\
  \midrule
  $\delta$ & POI collection radius for fine-grained semantic modeling \\
  $\mathbf{e}_i^{\text{fine}}$ & Initial fine-grained environmental semantic embedding of segment $i$ \\
  $\tilde{\mathbf{e}}_i^{\text{fine}}$ & Propagated fine-grained environmental semantic embedding of segment $i$ \\
  $\mathbf{h}_i^{(\ell)}$ & Hidden state of segment $i$ at layer $\ell$ in fine-grained semantic propagation \\
  $g_{(i,j)}$ & Spatial grid cell $(i,j)$ \\
  $\mathcal{H}_c$ & Functional hotspot grids for POI category $c$ \\
  $\mathbf{e}_{(i,j),c}^{\text{coarse}}$ & Coarse-grained semantic embedding of POI category $c$ in grid $g_{(i,j)}$ before spillover modeling \\
  $\tilde{\mathbf{e}}_{(i,j),c}^{\text{coarse}}$ & Coarse-grained semantic embedding of POI category $c$ in grid $g_{(i,j)}$ after spillover modeling \\
  $\tilde{\mathbf{e}}_{(i,j)}^{\text{coarse}}$ & Coarse-grained semantic embedding of grid $g_{(i,j)}$ after category-wise average pooling \\
  $\mathrm{grid}(i)$ & Grid assigned to segment $i$ \\
  \bottomrule
\end{tabular}
\end{table}

\begin{table}[H]
\caption{Notation for route choice encoder module.}
\label{tab:notation_route_choice_encoder}
\centering
\renewcommand{\arraystretch}{1.15}
\begin{tabular}{ll}
  \toprule
  \textbf{Symbol} & \textbf{Meaning} \\
  \midrule
  $\mathbf{r}_i$ & Basic road attribute embedding of segment $i$ \\
  $\tilde{\mathbf{r}}_i$ & Fused context-aware representation of segment $i$ \\
  $r_c$ & Candidate next road segment in the outgoing neighbor set $\mathcal{N}(r_i)$ \\
  $r_n$ & Terminal road segment of trajectory $\mathcal{T}$ \\
  $\rho_i$ & Journey progression ratio up to step $i$ \\
  $P(r_c\mid r_i)$ & Historical transition likelihood from $r_i$ to candidate segment $r_c$ \\
  $\Delta\theta_{r_c}$ & Destination-oriented directional deviation of candidate segment $r_c$ \\
  $\mathbf{s}_{\text{current}}$ & Current-state feature at step $i$ \\
  $\mathbf{s}_{r_c}$ & Adjacent-route context feature of candidate segment $r_c$ \\
  $\tilde{\mathbf{s}}_{r_c}$ & MoE-transformed adjacent-route context feature of candidate segment $r_c$ \\
  $\mathbf{o}_i$ & Route-choice behavior representation at step $i$ \\
  \bottomrule
\end{tabular}
\end{table}

\begin{table}[H]
\caption{Notation for Trajectory Embedding and pretraining.}
\label{tab:notation_trajectory_embedding}
\centering
\renewcommand{\arraystretch}{1.15}
\begin{tabular}{ll}
  \toprule
  \textbf{Symbol} & \textbf{Meaning} \\
  \midrule
  $\mathbf{x}_i$ & Input token at position $i$, combining route choice behavior and temporal context \\
  $\mathbf{t}_{m(t_i)}^{\mathrm{m}}$ & Learnable temporal embedding indexed by the minute-of-day of $t_i$ \\
  $\mathbf{t}_{d(t_i)}^{\mathrm{d}}$ & Learnable temporal embedding indexed by the day-of-week of $t_i$ \\
  $\tilde{\mathbf{x}}_i$ & Transformer output token at position $i$ \\
  $\mathcal{B}$ & Mini-batch of trajectories \\
  $\mathcal{L}$ & NT-Xent contrastive loss \\
  $\tau$ & Temperature in the NT-Xent loss \\
  $\text{pos}(i)$ & Positive-pair index for $\mathbf{z}_i$ \\
  \bottomrule
\end{tabular}
\end{table}

\section{Details of Environment Perception Module}

\subsection{Prompts for Fine-Grained Semantic Modeling}\label{subsec:appendix_prompt_fine}

At the fine-grained level, the objective is to generate detailed descriptions of individual road segments by summarizing the POI context surrounding them. The prompt template fed into the LLM consists of the following components:
\begin{itemize}
  \item \emph{Role Definition}: The LLM is prompted to assume the role of a resident, encouraging it to produce contextually grounded and locally nuanced descriptions.

  \item \emph{Structured POI Data Input}: A structured list of nearby POIs is provided as factual grounding for the description.

  \item \emph{Task Instruction}: The LLM is explicitly instructed to generate a description of the road segment based on the given information.
\end{itemize}

\begin{tcolorbox}[
  breakable,
  colback=white,
  colframe=black!38!gray,
  arc=2mm, auto outer arc,
  title=\textbf{LLM Prompt Template (Fine-Grained Semantic Modeling).}
]
\scriptsize
\ttfamily
You are a resident living in \textcolor[RGB]{166, 33, 0}{<city\_name>}, familiar with the local transportation network and surrounding POI information.\\

There is a road segment with the following POIs located within a 100-meter radius:\\

\textcolor[RGB]{166, 33, 0}{<poi\_count>} POIs categorized under [\textcolor[RGB]{166, 33, 0}{<poi\_category>}], further subdivided as:
\begin{itemize}
  \item \textcolor[RGB]{166, 33, 0}{<poi\_count>} [\textcolor[RGB]{166, 33, 0}{<poi\_subcategory>}]: \textcolor[RGB]{166, 33, 0}{<poi\_name\_list>}

  \item \textcolor[RGB]{166, 33, 0}{<poi\_count>} [\textcolor[RGB]{166, 33, 0}{<poi\_subcategory>}]: \textcolor[RGB]{166, 33, 0}{<poi\_name\_list>}

  \item \textcolor[RGB]{84, 130, 53}{$\cdots$ (Can be further expanded here)}\\
\end{itemize}

\textcolor[RGB]{84, 130, 53}{$\cdots$ (Additional primary categories may be added in the same format)}\\

Please describe the relevant characteristics of this road section based on the POI information within a 100-meter radius around it.
\end{tcolorbox}

To illustrate the applicability of the proposed template, we present a case study of a road segment on Qianmen East Street, which is located in a historically and commercially significant area of Beijing frequented by both residents and tourists.

\begin{tcolorbox}[
  breakable,
  colback=white,
  colframe=black!38!gray,
  arc=2mm, auto outer arc,
  title=\textbf{Illustrative Prompt Instance (Fine-Grained Semantic Modeling).}
]
\scriptsize
\ttfamily
You are a resident living in \textcolor[RGB]{166, 33, 0}{Beijing}, familiar with the local transportation network and surrounding POI information.\\

There is a road segment with the following POIs located within a 100-meter radius:\\

\textcolor[RGB]{166, 33, 0}{6} POIs categorized under [\textcolor[RGB]{166, 33, 0}{Shopping Services}], further subdivided as:
\begin{itemize}
  \item \textcolor[RGB]{166, 33, 0}{2} [\textcolor[RGB]{166, 33, 0}{Clothing \& Shoes Stores}]: \textcolor[RGB]{166, 33, 0}{Fashion Frontline (Xinjiekou Street), Old Beijing Cloth Shoes (Qianmen West Houheyan Street)}.
  
  \item \textcolor[RGB]{166, 33, 0}{2} [\textcolor[RGB]{166, 33, 0}{Specialty Stores}]: \textcolor[RGB]{166, 33, 0}{Gaojiazhai Grocery, Beijing Specialty Store (Qianmen West Heyan Community)}.
  
  \item \textcolor[RGB]{166, 33, 0}{2} [\textcolor[RGB]{166, 33, 0}{Convenience Stores}]: \textcolor[RGB]{166, 33, 0}{Lehafu Supermarket, Cigarettes \& Cold Drinks}.\\
\end{itemize}

\textcolor[RGB]{166, 33, 0}{4} POIs categorized under [\textcolor[RGB]{166, 33, 0}{Catering Services}], further subdivided as:
\begin{itemize}
  \item \textcolor[RGB]{166, 33, 0}{1} [\textcolor[RGB]{166, 33, 0}{Food-Related Establishment}]: \textcolor[RGB]{166, 33, 0}{Juewei Taiwanese Salt \& Pepper Chicken}.
  
  \item \textcolor[RGB]{166, 33, 0}{2} [\textcolor[RGB]{166, 33, 0}{Chinese Restaurants}]: \textcolor[RGB]{166, 33, 0}{Mian Ai Mian (Qianmen Branch), Qipin Xiang Tofu}.
  
  \item \textcolor[RGB]{166, 33, 0}{1} [\textcolor[RGB]{166, 33, 0}{Fast Food Restaurant}]: \textcolor[RGB]{166, 33, 0}{KFC (Qianmen Branch)}.\\
\end{itemize}

\textcolor[RGB]{166, 33, 0}{1} POIs categorized under [\textcolor[RGB]{166, 33, 0}{Transportation Facilities}], further subdivided as:
\begin{itemize}
  \item \textcolor[RGB]{166, 33, 0}{1} [\textcolor[RGB]{166, 33, 0}{Subway Station}]: \textcolor[RGB]{166, 33, 0}{Qianmen Subway Station (Southwest Exit C)}.\\
\end{itemize}

\textcolor[RGB]{166, 33, 0}{1} POIs categorized under [\textcolor[RGB]{166, 33, 0}{Government \& Social Organizations}], further subdivided as:
\begin{itemize}
  \item \textcolor[RGB]{166, 33, 0}{1} [\textcolor[RGB]{166, 33, 0}{Public Security Institution}]: \textcolor[RGB]{166, 33, 0}{Dashilan Patrol Police Station}.\\
\end{itemize}

\textcolor[RGB]{166, 33, 0}{1} POIs categorized under [\textcolor[RGB]{166, 33, 0}{Healthcare Services}], further subdivided as:
\begin{itemize}
  \item \textcolor[RGB]{166, 33, 0}{1} [\textcolor[RGB]{166, 33, 0}{Healthcare Facility}]: \textcolor[RGB]{166, 33, 0}{Blood Donation Cabin}.\\
\end{itemize}

\textcolor[RGB]{166, 33, 0}{2} POIs categorized under [\textcolor[RGB]{166, 33, 0}{Accommodation Services}], further subdivided as:
\begin{itemize}
  \item \textcolor[RGB]{166, 33, 0}{2} [\textcolor[RGB]{166, 33, 0}{Hotel}]: \textcolor[RGB]{166, 33, 0}{Qianmen Guanqi Hotel (Qianmen East Branch), Shunchaolong Hotel}.\\
\end{itemize}

Please describe the relevant characteristics of this road section based on the POI information within a 100-meter radius around it.
\end{tcolorbox}

\begin{tcolorbox}[
  breakable,
  colback=white,
  colframe=black!38!gray,
  arc=2mm, auto outer arc,
  title=\textbf{LLM-Generated Description (Fine-Grained Semantic Modeling).}
]
\scriptsize
\ttfamily
Based on POI information within 100 meters of this road section, the area has the following characteristics:\\

\textbf{1. Vibrant Commerce and Diverse Consumption Scenes}
\begin{itemize}
  \item \textbf{Dense Shopping Services}: Six shopping-related POIs form a commercial belt focused on clothing, shoes, specialty retail, and convenience stores.

  \item \textbf{Blend of Fashion and Tradition}: Stores such as ``Fashion Frontline (Xinjiekou Street)'' and ``Old Beijing Cloth Shoes (Qianmen West Houheyan Street)'' combine modern trends with local culture, attracting young people and tourists.

  \item \textbf{Convenience Meets Specialty}: Convenience stores like ``Lehafu Supermarket'' and ``Cigarettes \& Cold Drinks'' meet daily needs, while specialty shops like ``Gaojiazhai Grocery'' and ``Beijing Specialty Store (Qianmen West Heyan Community)'' offer local products favored by tourists seeking souvenirs.

  \item \textbf{Varied Dining Options}: Four food-related POIs cover fast food, Chinese cuisine, and specialty snacks, creating an all-day dining scene.

  \item \textbf{Local and International Flavors}: Restaurants like ``Mian Ai Mian (Qianmen Branch)'' and ``Qipin Xiang Tofu'' specialize in Beijing cuisine, while ``KFC (Qianmen Branch)'' and ``Juewei Taiwanese Salt \& Pepper Chicken'' offer quick and convenient meals, catering to both locals and tourists. \\
\end{itemize}

\textbf{2. Transportation Hub and High Foot Traffic}
\begin{itemize}
 \item \textbf{Direct Metro Access}: ``Qianmen Subway Station (Southwest Exit C)'' links subway and ground transport, generating heavy flows during rush hours and serving both commercial and commuting needs.

 \item \textbf{Public Security}: The ``Dashilan Police Patrol Station'' reflects strict public safety management, enhancing residents' and visitors' sense of security. \\
\end{itemize}

\textbf{3. Prominent Tourism and Accommodation Functions}
\begin{itemize}
  \item \textbf{Comprehensive Lodging Options}: Hotels such as ``Qianmen Guanqi Hotel (Qianmen East Branch)'' and ``Shunchaolong Hotel'' confirm the area as a tourist hotspot, likely near historical and cultural attractions like Qianmen Street.

  \item \textbf{Cultural Experience Extension}: POIs like ``Old Beijing Cloth Shoes (Qianmen West Houheyan Street)'' and ``Beijing Specialty Store'' complement the Qianmen historical–cultural district and enhance its tourism appeal. \\
\end{itemize}

\textbf{4. Healthcare and Community Services}
\begin{itemize}
  \item \textbf{Blood Donation Cabin}: The blood donation cabin provides basic medical support and public welfare services for residents and tourists, strengthening community functions. \\
\end{itemize}

\textbf{Summary}: This road section lies in the \textbf{core Qianmen business district}, integrating \textbf{commerce, transportation, tourism, and residential} functions, and exemplifying the fusion of Beijing's historical culture with modern lifestyle. Its dense POIs and diverse services make it a key node for tourists, residents, and commuters, well suited for walking, shopping, leisure, and short-term stays.
\end{tcolorbox}

As can be seen, given a list of POIs around a road segment (e.g., 6 shopping-related POIs, 4 dining-related POIs), the LLM does more than simply restate this information. Instead, it uses commonsense reasoning to derive abstract functional concepts such as ``vibrant commerce and diverse consumption scenes'' and ``transportation hub and high foot traffic''. This process constitutes a conceptual leap from discrete, low-level features to a high-level, synthesized semantic representation. Notably, these abstract concepts are not present in the raw input features; they are inferred by the LLM using its world knowledge.

\subsection{Prompts for Coarse-Grained Semantic Modeling}\label{subsec:appendix_prompt_coarse}

At the coarse-grained level, we aim to generate textual descriptions for \emph{functional hotspots} $\mathcal{H}_c$ of POI type $c$. Similar to the prompt designed for fine-grained POI context described above, this prompt also comprises three components: \emph{Role Definition}, \emph{Structured POI Data Input}, and \emph{Task Instruction}.

\begin{tcolorbox}[
  breakable,
  colback=white,
  colframe=black!38!gray,
  arc=2mm, auto outer arc,
  title=\textbf{LLM Prompt Template (Coarse-Grained Semantic Modeling).}
]
\scriptsize
\ttfamily
You are a resident living in \textcolor[RGB]{166, 33, 0}{<city\_name>}, familiar with the local transportation network and surrounding POI information.\\

In a 1000m $\times$ 1000m area of \textcolor[RGB]{166, 33, 0}{<city\_name>}, POIs of the type [\textcolor[RGB]{166, 33, 0}{<poi\_category>}] exhibit significant clustering characteristics. Data analysis shows that the number of [\textcolor[RGB]{166, 33, 0}{<poi\_category>}] POIs in this area ranks within the top 10\% in \textcolor[RGB]{166, 33, 0}{<city\_name>}. Further subdividing these [\textcolor[RGB]{166, 33, 0}{<poi\_category>}] POIs, they include:\\

\begin{itemize}
  \item \textcolor[RGB]{166, 33, 0}{<poi\_count>} [\textcolor[RGB]{166, 33, 0}{<poi\_subcategory>}]: \textcolor[RGB]{166, 33, 0}{<poi\_name\_list>}

  \item \textcolor[RGB]{166, 33, 0}{<poi\_count>} [\textcolor[RGB]{166, 33, 0}{<poi\_subcategory>}]: \textcolor[RGB]{166, 33, 0}{<poi\_name\_list>}

  \item \textcolor[RGB]{84, 130, 53}{$\cdots$ (Can be further expanded here)}\\
\end{itemize}

Please describe the characteristics of this 1000m $\times$ 1000m area where a large number of [\textcolor[RGB]{166, 33, 0}{<poi\_category>}] POIs are clustered.
\end{tcolorbox}

Below we present an example of a functional hotspot of type ``Shopping Services'' near Beijing CBD.

\begin{tcolorbox}[
  breakable,
  colback=white,
  colframe=black!38!gray,
  arc=2mm, auto outer arc,
  title=\textbf{Illustrative Prompt Instance (Coarse-Grained Semantic Modeling).}
]
\scriptsize
\ttfamily
You are a resident living in \textcolor[RGB]{166, 33, 0}{Beijing}, familiar with the local transportation network and surrounding POI information.\\

In a 1000m $\times$ 1000m area of \textcolor[RGB]{166, 33, 0}{Beijing}, POIs of the type [\textcolor[RGB]{166, 33, 0}{Shopping Services}] exhibit significant clustering characteristics. Data analysis shows that the number of [\textcolor[RGB]{166, 33, 0}{Shopping Services}] POIs in this area ranks within the top 10\% in \textcolor[RGB]{166, 33, 0}{Beijing}. Further subdividing these [\textcolor[RGB]{166, 33, 0}{Shopping Services}] POIs, they include:\\

\begin{itemize}
  \item \textcolor[RGB]{166, 33, 0}{790} [\textcolor[RGB]{166, 33, 0}{Specialty Stores}]: \textcolor[RGB]{166, 33, 0}{Li-Ning (Fuli Plaza Store), Dionysus Wine Cellar, Youngor, etc.}

  \item \textcolor[RGB]{166, 33, 0}{80} [\textcolor[RGB]{166, 33, 0}{Personal Care \& Cosmetics Stores}]: \textcolor[RGB]{166, 33, 0}{Qingzhuang Cosmetics, Scent Library, PIPI FACE Perfume, etc.}

  \item \textcolor[RGB]{166, 33, 0}{24} [\textcolor[RGB]{166, 33, 0}{Sports Goods Stores}]: \textcolor[RGB]{166, 33, 0}{Adidas (China World Trade Center South Zone), Reebok Treadmill Store (Guomao Store), Kingsmith Treadmill Store (Guomao Store), etc.}

  \item \textcolor[RGB]{166, 33, 0}{52} [\textcolor[RGB]{166, 33, 0}{Convenience Stores}]: \textcolor[RGB]{166, 33, 0}{7-Eleven (Huamao Store), Quick (Jianwai Branch 2), Wankeyuan Supermarket (Jingheng Street), etc.}

  \item \textcolor[RGB]{166, 33, 0}{40} [\textcolor[RGB]{166, 33, 0}{Shopping Malls}]: \textcolor[RGB]{166, 33, 0}{Beijing Yintai Center in01, China World Trade Center South Zone, China World Trade Center North Zone, etc.}

  \item \textcolor[RGB]{166, 33, 0}{46} [\textcolor[RGB]{166, 33, 0}{Home \& Building Material Markets}]: \textcolor[RGB]{166, 33, 0}{Dakang Home Furnishing, Risheng Furniture \& Hardware, Bairuisi Furniture, etc.}

  \item \textcolor[RGB]{166, 33, 0}{46} [\textcolor[RGB]{166, 33, 0}{Electronics \& Appliance Stores}]: \textcolor[RGB]{166, 33, 0}{Changhong Air Conditioning, ThinKpad Laptop Store, BOSE (China World Trade Center North Zone), etc.}

  \item \textcolor[RGB]{166, 33, 0}{8} [\textcolor[RGB]{166, 33, 0}{Stationery Stores}]: \textcolor[RGB]{166, 33, 0}{JimmyBlack, 7G Office, Montblanc (China World Trade Center South Zone), etc.}

  \item \textcolor[RGB]{166, 33, 0}{708} [\textcolor[RGB]{166, 33, 0}{Clothing \& Shoes Stores}]: \textcolor[RGB]{166, 33, 0}{Yaliange, Xiu Clothing, Xiaojin Wardrobe (Jianwai SOHO Store), etc.}

  \item \textcolor[RGB]{166, 33, 0}{6} [\textcolor[RGB]{166, 33, 0}{Auction Houses}]: \textcolor[RGB]{166, 33, 0}{Beijing Jianya Century Auction Co., Ltd., Beijing Zhonghan Auction Co., Ltd., Beijing Zhongzhao International Auction Co., Ltd. (Jintong East Road), etc.}

  \item \textcolor[RGB]{166, 33, 0}{62} [\textcolor[RGB]{166, 33, 0}{Flower \& Pet Markets}]: \textcolor[RGB]{166, 33, 0}{No. 18 Flower Shop, Love Flowers, LOVE Flower Studio, etc.}

  \item \textcolor[RGB]{166, 33, 0}{28} [\textcolor[RGB]{166, 33, 0}{Supermarkets}]: \textcolor[RGB]{166, 33, 0}{Jingkelong Supermarket, Mannings (Yintai Center Store), CR Vanguard (Guomao Store), etc.}

  Please describe the characteristics of this 1000m $\times$ 1000m area where a large number of [\textcolor[RGB]{166, 33, 0}{Shopping Services}] POIs are clustered.
\end{itemize}
\end{tcolorbox}

\begin{tcolorbox}[
  breakable,
  colback=white,
  colframe=black!38!gray,
  arc=2mm, auto outer arc,
  title=\textbf{LLM-Generated Description (Coarse-Grained Semantic Modeling).}
]
\scriptsize
\ttfamily
Based on the data, the 1000m $\times$ 1000m area shows a highly concentrated commercial profile, summarized as follows:

\textbf{1. Highly Concentrated, Diverse Business Mix}
\begin{itemize}
  \item \textbf{Core Business Types}: Mainly [specialty stores] (790) and [clothing \& shoes stores] (708), indicating strong fashion and personalized shopping, likely with brand flagships and boutiques.

  \item \textbf{Supporting Daily Services}: [convenience stores] (52) and [supermarkets] (28) enable one-stop shopping for everyday needs.

  \item \textbf{High-End and Niche Segments}: [electronics \& appliance stores] (46), [home \& building material markets] (46), and [stationery stores] (8) show that the area also serves professionals and affluent residents. \\
\end{itemize}

\textbf{2. Likely Location: Beijing's Core Commercial Districts}
\begin{itemize}
  \item \textbf{POI Name Clues}: Frequent ``Guomao'' in POI names suggests proximity to the Guomao CBD or a similar prime business district.

   \item \textbf{Exceptionally High POI Density}: Nearly 2,000 shopping-related POIs within one square kilometer, far exceeding typical commercial zones, points to a central retail hub such as Guomao, Sanlitun, or Wangjing. \\
\end{itemize}

\textbf{3. Multi-Level Consumption Structure}
\begin{itemize}
  \item \textbf{Mass Market Consumption}: Supermarkets and convenience stores address the daily needs of nearby residents.

  \item \textbf{High-End and Niche Consumption}: Stores like ``Dionysus Wine Cellar'', ``Montblanc'', and ``Beijing Jianya Century Auction Co., Ltd.'' reflect luxury brands and specialized services targeting affluent and interest-driven customers. \\
\end{itemize}

\textbf{4. Commercial Vitality Driven by Transit and Foot Traffic}
\begin{itemize}
  \item \textbf{Density Backed by Accessibility}: Such POI concentration implies strong transport links (e.g., metro, bus hubs) and high population density (e.g., office towers, housing). ``Guomao'' further hints at a metro hub or major retail cluster.

  \item \textbf{Competitive and Complementary Dynamics}: Coexisting specialty, fashion, and general markets suggest differentiated positioning (e.g., brand vs. value), while services like convenience stores provide complementary functions that enhance overall vitality. \\
\end{itemize}

\textbf{5. Potential Challenges}
\begin{itemize}
  \item \textbf{Oversaturation Risk}: High POI density may cause homogeneous competition and pressure on smaller retailers.

   \item \textbf{Traffic Pressure}: Intensive commercial activity can strain surrounding roads, requiring effective traffic planning and management. \\
\end{itemize}

\textbf{Conclusion:}
The area is most likely a \textbf{core commercial district in Beijing} (e.g., Guomao, Sanlitun) or a \textbf{large emerging commercial complex}, characterized by:
\begin{itemize}
  \item \textbf{Diverse Business Types}: Covering everyday retail and luxury brands.

  \item \textbf{Broad Consumption Spectrum}: Serving mass, middle-class, and affluent groups.

  \item \textbf{Transit-Oriented Clustering}: High commercial density supported by transit and strong population flows.

  \item \textbf{Potential Functional Zoning}: Distinct commercial ecosystems indicated by specialized POIs.
\end{itemize}
\end{tcolorbox}

These results demonstrate that, in coarse-grained semantic modeling, LLMs also utilize their encoded world knowledge to perform powerful spatial pattern recognition. For example, by detecting dense clusters of specialty stores and shopping malls, the model not only identifies the area as a core commercial district of Beijing but also reveals deeper characteristics, such as its multi-level consumption structure and commercial vitality driven by transit and foot traffic.

\subsection{Faithfulness Check of LLM-Generated Descriptions}\label{subsec:appendix_faithfulness_check}

To verify the faithfulness of LLM-generated descriptions, we adopt a source-grounded factuality evaluation protocol commonly used in abstractive summarization~\cite{maynez2020faithfulness, kryscinski2020evaluating, pagnoni2021understanding}. In this protocol, structured POI records serve as source evidence, and each generated description is labeled as supported by, unsupported by, or inconsistent with its corresponding records. We randomly sample 100 Beijing descriptions (50 fine-grained road segments and 50 coarse-grained functional hotspots) and manually assess them according to the criteria below.

\begin{itemize}
  \item \emph{Supported}: The main semantic characterization is grounded in the input POI records, or can be reasonably inferred from their category, subcategory, count, name, and spatial scope.

  \item \emph{Unsupported}: The description introduces entities, functions, traffic states, population flows, historical status, consumption levels, or other attributes that are absent from and not inferable from the POI records.

  \item \emph{Factual Deviation}: The description conflicts with the input POI records, such as by reporting an incorrect POI category, count, dominant function, or spatial scope.

  \item \emph{Severe Error}: The unsupported or factually deviated content would mislead downstream road-segment or region-level semantic representation by changing the dominant urban semantics.
\end{itemize}

\begin{table}[H]
\caption{Faithfulness check of LLM-generated descriptions.}
\label{tab:llm_faithfulness}
\centering
\scriptsize
\begin{tabular}{lcccc}
\toprule
\textbf{Granularity} & \textbf{Fully Supported}$\uparrow$ & \textbf{Unsupported}$\downarrow$ & \textbf{Factual Deviation}$\downarrow$ & \textbf{Severe Error}$\downarrow$ \\
\midrule
Fine-grained & 92\% & 6\% & 2\% & 0\% \\
Coarse-grained & 94\% & 4\% & 2\% & 0\% \\
\bottomrule
\end{tabular}
\end{table}

As shown in \cref{tab:llm_faithfulness}, the generated descriptions are largely faithful: 92\% of fine-grained and 94\% of coarse-grained descriptions are fully supported, with only 6\%/4\% unsupported and 2\%/2\% showing factual deviations. No severe errors are observed, indicating that minor unsupported or deviated content does not alter the dominant semantics used for downstream representation learning. This low error rate is expected because the LLM summarizes structured POI records rather than making open-ended decisions, which constrains generation and reduces task-critical hallucinations.

\section{Additional Experimental Results}

\subsection{Ablation Studies}\label{subsec:appendix_ablation}

\begin{figure}[H]
    \centering
    \includegraphics[width=1.0\linewidth]{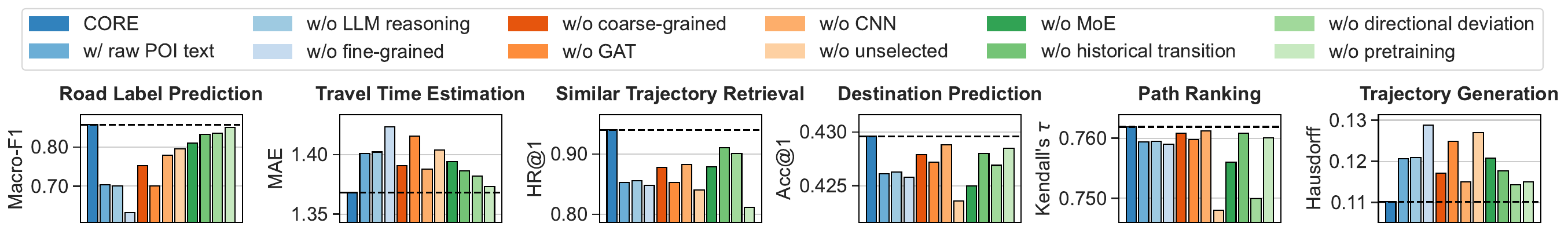}
    \caption{Performance comparison of CORE and its ablated variants on the Chengdu dataset.}
    \label{fig:ablation_chengdu}
\end{figure}

\begin{figure}[H]
    \centering
    \includegraphics[width=1.0\linewidth]{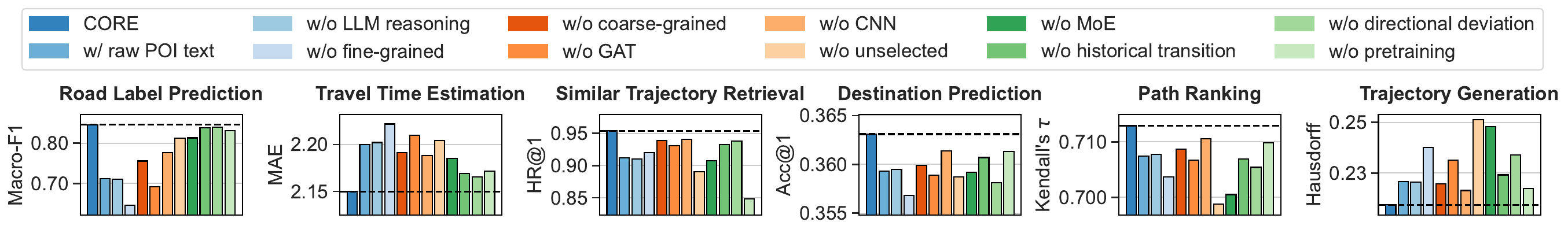}
    \caption{Performance comparison of CORE and its ablated variants on the Xi'an dataset.}
    \label{fig:ablation_xian}
\end{figure}

\begin{figure}[H]
    \centering
    \includegraphics[width=1.0\linewidth]{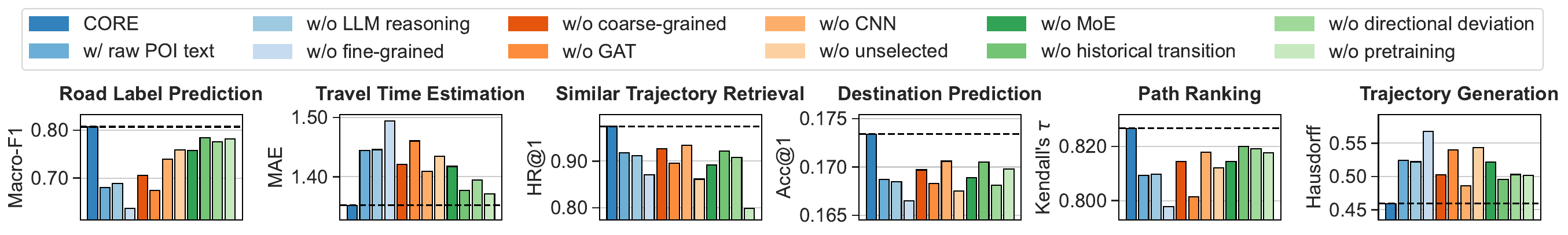}
    \caption{Performance comparison of CORE and its ablated variants on the Porto dataset.}
    \label{fig:ablation_porto}
\end{figure}

\cref{fig:ablation_chengdu,fig:ablation_xian,fig:ablation_porto} report additional ablation results on the Chengdu, Xi'an, and Porto datasets. These results exhibit trends that are highly consistent with those observed on the Beijing dataset in the main text, further corroborating the robustness and effectiveness of CORE's core components under different urban contexts.

\subsection{Data Efficiency}\label{subsec:appendix_data_efficiency}

\begin{figure}[H]
    \centering
    \includegraphics[width=1.0\linewidth]{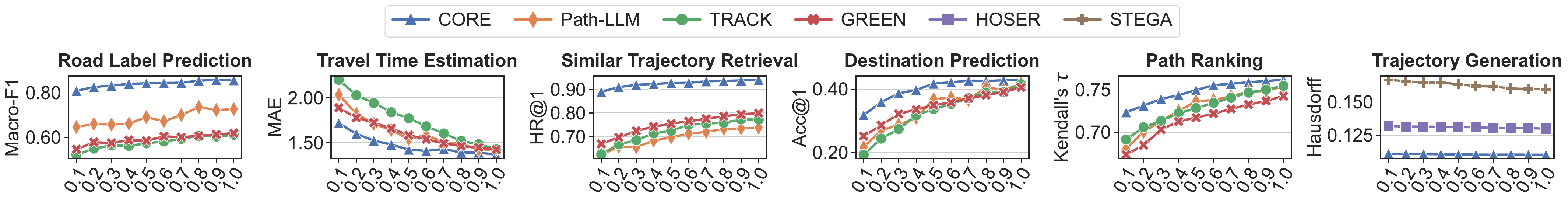}
    \caption{Results of using varying proportions of training data on the Chengdu dataset.}
    \label{fig:data_efficiency_chengdu}
\end{figure}

\begin{figure}[H]
    \centering
    \includegraphics[width=1.0\linewidth]{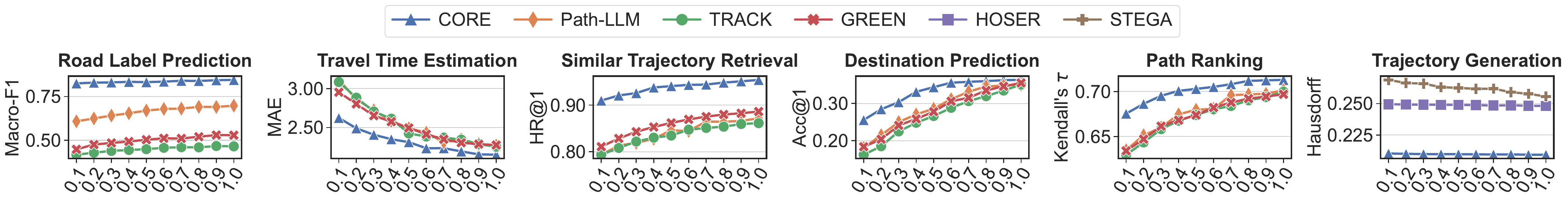}
    \caption{Results of using varying proportions of training data on the Xi'an dataset.}
    \label{fig:data_efficiency_xian}
\end{figure}

\begin{figure}[H]
    \centering
    \includegraphics[width=1.0\linewidth]{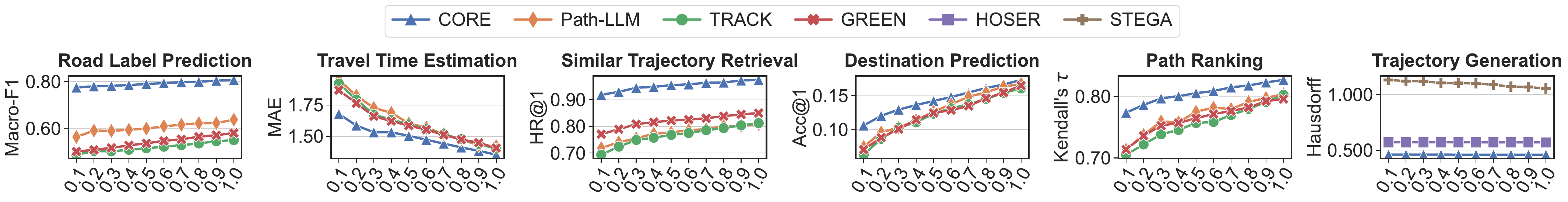}
    \caption{Results of using varying proportions of training data on the Porto dataset.}
    \label{fig:data_efficiency_porto}
\end{figure}

\cref{fig:data_efficiency_chengdu,fig:data_efficiency_xian,fig:data_efficiency_porto} present the data-efficiency results on the Chengdu, Xi'an, and Porto datasets. The observed trends are highly consistent with the results on the Beijing dataset presented in the main text, further demonstrating that CORE reliably maintains strong data efficiency across different cities and data scales.

\subsection{Sensitivity Analyses}

\subsubsection{Hyperparameter Sensitivity}
\label{subsec:appendix_hyperparameter_sensitivity}

\begin{figure}[H]
  \centering
  \includegraphics[width=1.0\linewidth]{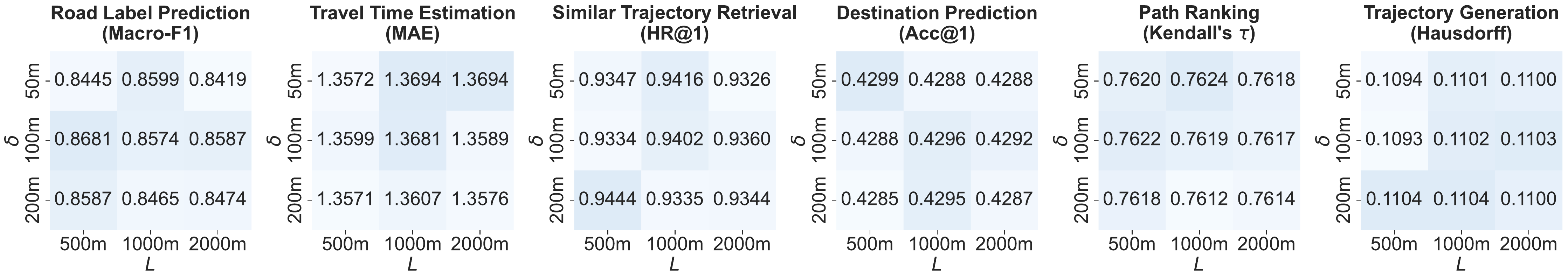}
  \caption{Sensitivity of CORE to spatial perception scale hyperparameters ($\delta$ and $L$) on the Chengdu dataset.}
  \label{fig:hyperparameter_chengdu}
\end{figure}

\begin{figure}[H]
  \centering
  \includegraphics[width=1.0\linewidth]{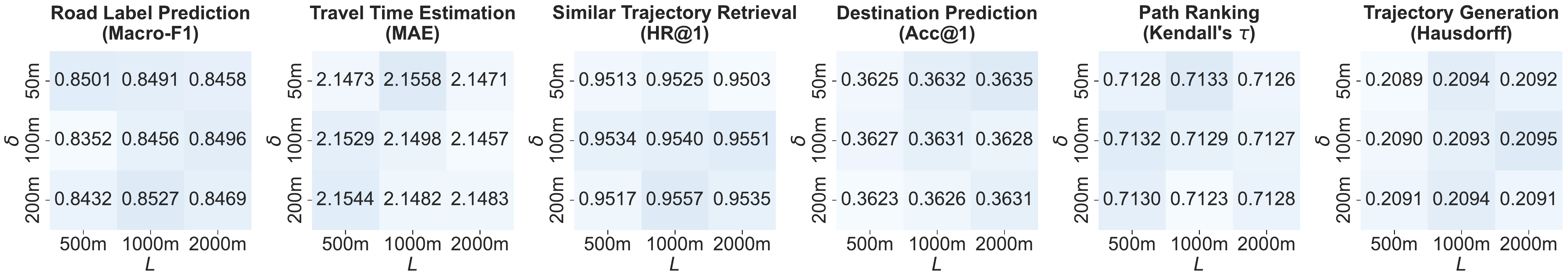}
  \caption{Sensitivity of CORE to spatial perception scale hyperparameters ($\delta$ and $L$) on the Xi'an dataset.}
  \label{fig:hyperparameter_xian}
\end{figure}

\begin{figure}[H]
  \centering
  \includegraphics[width=1.0\linewidth]{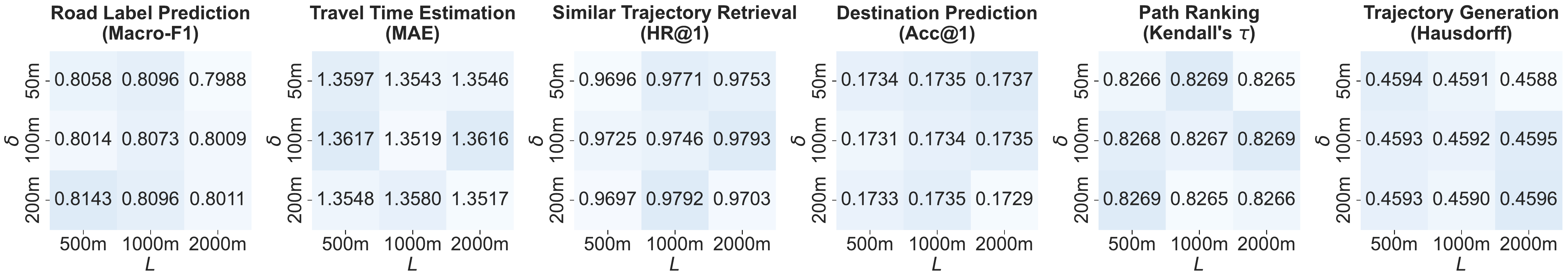}
  \caption{Sensitivity of CORE to spatial perception scale hyperparameters ($\delta$ and $L$) on the Porto dataset.}
  \label{fig:hyperparameter_porto}
\end{figure}

\cref{fig:hyperparameter_chengdu,fig:hyperparameter_xian,fig:hyperparameter_porto} show the sensitivity of CORE to the spatial perception hyperparameters (i.e., the POI perception radius $\delta$ and the grid side length $L$) on the Chengdu, Xi'an, and Porto datasets. Within the examined ranges, performance remains stable, suggesting that CORE does not critically depend on a specific choice of spatial scale.

\begin{figure}[H]
  \centering
  \includegraphics[width=1.0\linewidth]{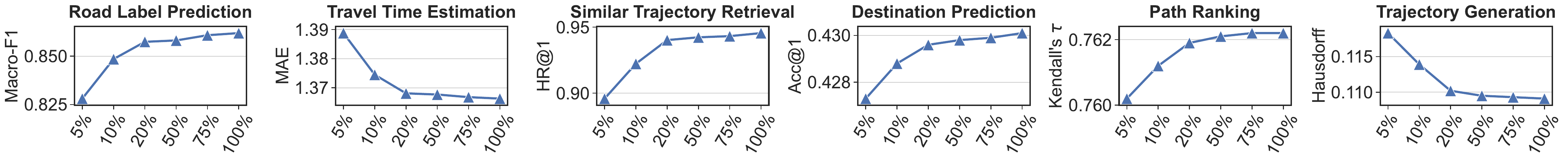}
  \caption{Impact of the critical segment selection ratio $\eta$ on the Chengdu dataset.}
  \label{fig:llm_ratio_chengdu}
\end{figure}

\begin{figure}[H]
  \centering
  \includegraphics[width=1.0\linewidth]{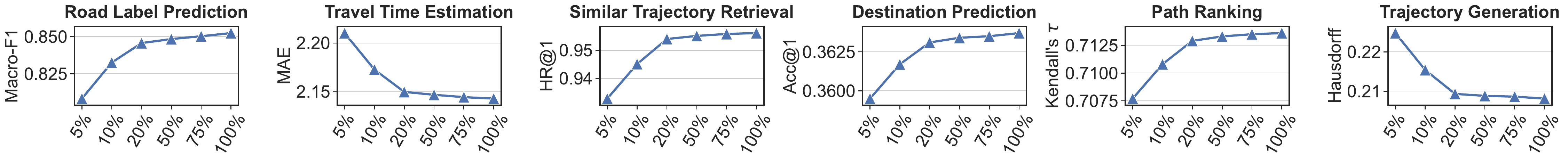}
  \caption{Impact of the critical segment selection ratio $\eta$ on the Xi'an dataset.}
  \label{fig:llm_ratio_xian}
\end{figure}

\begin{figure}[H]
  \centering
  \includegraphics[width=1.0\linewidth]{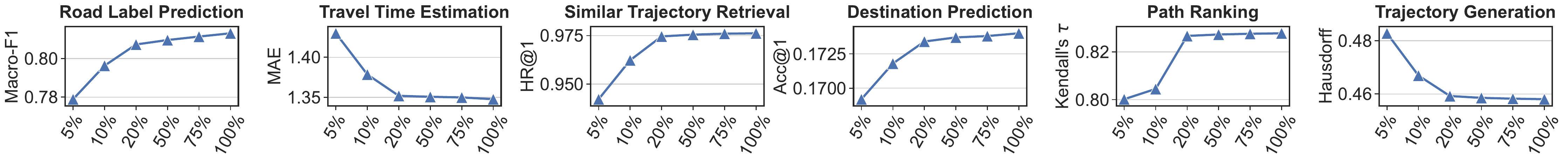}
  \caption{Impact of the critical segment selection ratio $\eta$ on the Porto dataset.}
  \label{fig:llm_ratio_porto}
\end{figure}

In addition, \cref{fig:llm_ratio_chengdu,fig:llm_ratio_xian,fig:llm_ratio_porto} illustrate the impact of the \emph{critical segment} selection ratio $\eta$ on the Chengdu, Xi'an, and Porto datasets. The observed trends align with those in the main text, confirming that the top 20\% selection strategy consistently offers a favorable trade-off between performance and efficiency across diverse urban environments.

\subsubsection{Transformer Capacity Sensitivity}
\label{subsec:appendix_transformer_capacity_sensitivity}

As shown in \cref{tab:transformer_capacity_sensitivity_chengdu_xian_porto}, CORE exhibits broadly stable performance on Chengdu, Xi'an, and Porto under different encoder depths and hidden dimensions, further indicating that its effectiveness does not depend on a specific Transformer capacity. Consistent with the main-text observation, moderate capacity provides a better trade-off, while overly small or large configurations yield limited benefits.

\begin{table}[H]
\caption{Transformer capacity sensitivity on the Chengdu, Xi'an, and Porto datasets.}
\label{tab:transformer_capacity_sensitivity_chengdu_xian_porto}
\centering
\scriptsize
\begin{tabular}{lcccccccc}
\toprule
\textbf{City} & \textbf{Factor} & \textbf{Setting}
& \textbf{Macro-F1}$\uparrow$
& \textbf{MAE}$\downarrow$
& \textbf{HR@1}$\uparrow$
& \textbf{Acc@1}$\uparrow$
& \textbf{Kendall's $\tau$}$\uparrow$
& \textbf{Hausdorff}$\downarrow$ \\
\midrule

\multirow{7}{*}[\multirowoffset]{\textbf{Chengdu}} & \multirow{4}{*}{Encoder Layers}
& 2 & \textbf{0.8685} & 1.3843 & 0.9289 & 0.4278 & 0.7558 & 0.1108 \\
& & 4 & \underline{0.8630} & 1.3772 & \underline{0.9384} & \underline{0.4291} & \textbf{0.7630} & \underline{0.1105} \\
& & 6 & 0.8574 & \textbf{1.3681} & \textbf{0.9402} & \textbf{0.4296} & \underline{0.7619} & \textbf{0.1102} \\
& & 8 & 0.8485 & \underline{1.3709} & 0.9379 & 0.4287 & 0.7606 & 0.1109 \\
\cmidrule(lr){2-9}
& \multirow{3}{*}{Hidden Dimension}
& 64  & 0.8143 & 1.3764 & 0.9329 & 0.4279 & \underline{0.7605} & 0.1106 \\
& & 128 & \underline{0.8574} & \textbf{1.3681} & \textbf{0.9402} & \textbf{0.4296} & \textbf{0.7619} & \underline{0.1102} \\
& & 256 & \textbf{0.8732} & \underline{1.3732} & \underline{0.9388} & \underline{0.4292} & 0.7581 & \textbf{0.1099} \\

\cmidrule(lr){1-9}
\multirow{7}{*}[\multirowoffset]{\textbf{Xi'an}} & \multirow{4}{*}{Encoder Layers}
& 2 & 0.8270 & 2.2067 & 0.9357 & 0.3584 & 0.7060 & \underline{0.2097} \\
& & 4 & \textbf{0.8575} & 2.1639 & 0.9508 & 0.3599 & 0.7103 & 0.2101 \\
& & 6 & \underline{0.8456} & \textbf{2.1498} & \underline{0.9540} & \textbf{0.3631} & \underline{0.7129} & \textbf{0.2093} \\
& & 8 & 0.8363 & \underline{2.1585} & \textbf{0.9557} & \underline{0.3611} & \textbf{0.7132} & 0.2098 \\
\cmidrule(lr){2-9}
& \multirow{3}{*}{Hidden Dimension}
& 64  & 0.7992 & 2.1659 & 0.9489 & 0.3605 & \underline{0.7113} & 0.2104 \\
& & 128 & \underline{0.8456} & \textbf{2.1498} & \textbf{0.9540} & \underline{0.3631} & \textbf{0.7129} & \textbf{0.2093} \\
& & 256 & \textbf{0.8663} & \underline{2.1547} & \underline{0.9520} & \textbf{0.3635} & 0.7098 & \underline{0.2102} \\

\cmidrule(lr){1-9}
\multirow{7}{*}[\multirowoffset]{\textbf{Porto}} & \multirow{4}{*}{Encoder Layers}
& 2 & \textbf{0.8339} & 1.3702 & 0.9350 & 0.1641 & 0.8212 & \textbf{0.4589} \\
& & 4 & 0.7887 & 1.3722 & 0.9663 & 0.1721 & 0.8244 & 0.4594 \\
& & 6 & \underline{0.8073} & \underline{1.3519} & \textbf{0.9746} & \underline{0.1734} & \textbf{0.8267} & \underline{0.4592} \\
& & 8 & 0.7791 & \textbf{1.3491} & \underline{0.9701} & \textbf{0.1741} & \underline{0.8253} & 0.4597 \\
\cmidrule(lr){2-9}
& \multirow{3}{*}{Hidden Dimension}
& 64  & 0.7386 & 1.3688 & 0.9593 & 0.1708 & 0.8192 & \textbf{0.4571} \\
& & 128 & \underline{0.8073} & \textbf{1.3519} & \textbf{0.9746} & \textbf{0.1734} & \textbf{0.8267} & 0.4592 \\
& & 256 & \textbf{0.8411} & \underline{1.3543} & \underline{0.9710} & \underline{0.1715} & \underline{0.8261} & \underline{0.4582} \\

\bottomrule
\end{tabular}
\end{table}

\subsubsection{LLM Backbone Sensitivity}
\label{subsec:appendix_llm_backbone_sensitivity}

As shown in \cref{tab:llm_sensitivity_chengdu_xian_porto}, the results on Chengdu, Xi'an, and Porto follow the same trend as the Beijing results in the main text: Qwen3-0.6B consistently underperforms, while DeepSeek-V3.2-Exp brings only marginal gains over Qwen3-8B. This further confirms that CORE is robust to LLM backbone choices once sufficient semantic capacity is provided, with Qwen3-8B offering an effective accuracy-cost trade-off across datasets.

\begin{table}[H]
\caption{Performance of CORE with different LLM backbones on the Chengdu, Xi'an, and Porto datasets.}
\label{tab:llm_sensitivity_chengdu_xian_porto}
\centering
\scriptsize
\begin{tabular}{ll cccccc}
  \toprule
  \textbf{City} & \textbf{LLM Backbone} & \textbf{Macro-F1}$\uparrow$ & \textbf{MAE}$\downarrow$ & \textbf{HR@1}$\uparrow$ & \textbf{Acc@1}$\uparrow$ & \textbf{Kendall's $\tau$}$\uparrow$ & \textbf{Hausdorff}$\downarrow$ \\
  \midrule
  \multirow{3}{*}{\textbf{Chengdu}} & CORE w/ Qwen3-0.6B & 0.8182 & 1.3837 & 0.9218 & 0.4267 & 0.7601 & 0.1201 \\
  & CORE w/ Qwen3-8B & \textbf{0.8574} & \underline{1.3681} & \underline{0.9402} & \textbf{0.4296} & \underline{0.7619} & \underline{0.1102} \\
  & CORE w/ DeepSeek-V3.2-Exp & \underline{0.8541} & \textbf{1.3675} & \textbf{0.9408} & \underline{0.4294} & \textbf{0.7622} & \textbf{0.1098} \\
  \cmidrule(lr){1-8}
  \multirow{3}{*}{\textbf{Xi'an}} & CORE w/ Qwen3-0.6B & 0.8163 & 2.1934 & 0.9399 & 0.3589 & 0.7067 & 0.2285 \\
  & CORE w/ Qwen3-8B & \underline{0.8456} & \underline{2.1498} & \textbf{0.9540} & \textbf{0.3631} & \underline{0.7129} & \textbf{0.2093} \\
  & CORE w/ DeepSeek-V3.2-Exp & \textbf{0.8472} & \textbf{2.1475} & \underline{0.9536} & \underline{0.3627} & \textbf{0.7131} & \underline{0.2097} \\
  \cmidrule(lr){1-8}
  \multirow{3}{*}{\textbf{Porto}} & CORE w/ Qwen3-0.6B & 0.7693 & 1.4029 & 0.9587 & 0.1706 & 0.8056 & 0.4793 \\
  & CORE w/ Qwen3-8B & \underline{0.8073} & \underline{1.3519} & \textbf{0.9746} & \underline{0.1734} & \textbf{0.8267} & \underline{0.4592} \\
  & CORE w/ DeepSeek-V3.2-Exp & \textbf{0.8089} & \textbf{1.3493} & \underline{0.9738} & \textbf{0.1739} & \underline{0.8265} & \textbf{0.4587} \\
  \bottomrule
\end{tabular}
\end{table}

\subsubsection{POI Completeness Sensitivity}
\label{subsec:appendix_poi_ratio}

\begin{figure}[H]
  \centering
  \includegraphics[width=1.0\linewidth]{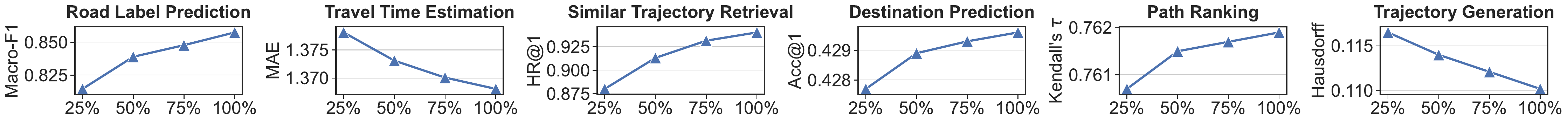}
  \caption{Sensitivity of CORE to POI completeness on the Chengdu dataset.}
  \label{fig:poi_ratio_chengdu}
\end{figure}

\begin{figure}[H]
  \centering
  \includegraphics[width=1.0\linewidth]{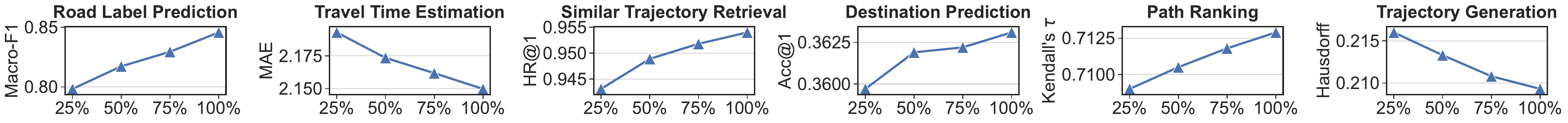}
  \caption{Sensitivity of CORE to POI completeness on the Xi'an dataset.}
  \label{fig:poi_ratio_xian}
\end{figure}

\begin{figure}[H]
  \centering
  \includegraphics[width=1.0\linewidth]{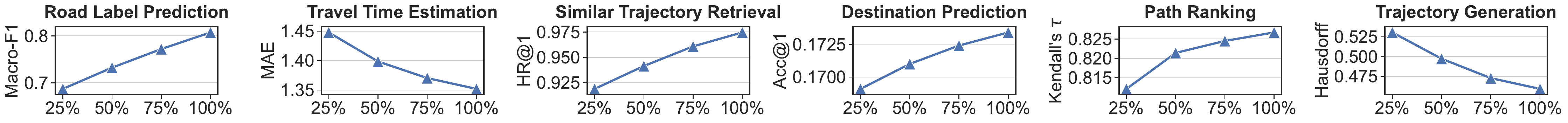}
  \caption{Sensitivity of CORE to POI completeness on the Porto dataset.}
  \label{fig:poi_ratio_porto}
\end{figure}

\cref{fig:poi_ratio_chengdu,fig:poi_ratio_xian,fig:poi_ratio_porto} report the impact of POI completeness on CORE across the Chengdu, Xi'an, and Porto datasets. The observed trends align perfectly with those in the main text, confirming that CORE consistently maintains robustness against sparse environmental data across diverse urban scenarios.

\subsection{Critical Segment Selection Criterion}
\label{subsec:appendix_critical_segment_selection}

To examine whether traffic volume is the most suitable criterion for selecting \emph{critical segments}, we compare three strategies under the same LLM inference budget, with $\eta=20\%$ of road segments selected for fine-grained semantic modeling: (1) \emph{Traffic volume}, which selects segments with the largest visit counts $v_i$; (2) \emph{POI density}, which selects segments with the largest number of nearby POIs $d_i^{\mathrm{poi}}$ within radius $\delta$; and (3) \emph{Traffic-POI rank fusion}, which jointly considers both signals.

For fusion, we first convert both criteria into percentile ranks to avoid scale mismatch:
\begin{equation}
  \rho_i^{\mathrm{vol}} =
  \frac{\operatorname{rank}(v_i)-1}{|\mathcal{V}|-1}, \quad
  \rho_i^{\mathrm{poi}} =
  \frac{\operatorname{rank}(d_i^{\mathrm{poi}})-1}{|\mathcal{V}|-1},
\end{equation}
where $\operatorname{rank}(\cdot)$ assigns 1 to the smallest value and $|\mathcal{V}|$ to the largest value. The fused score is the average rank:
\begin{equation}
  s_i^{\mathrm{fusion}} =
  \frac{1}{2}\left(\rho_i^{\mathrm{vol}} + \rho_i^{\mathrm{poi}}\right).
\end{equation}
We select the top $\eta$ segments according to $s_i^{\mathrm{fusion}}$.

\begin{table}[H]
\caption{Performance comparison of different critical segment selection criteria.}
\label{tab:critical_segment_selection_criterion}
\centering
\scriptsize
\begin{tabular}{llcccccc}
  \toprule
  \textbf{City} & \textbf{Selection Criterion} & \textbf{Macro-F1}$\uparrow$ & \textbf{MAE}$\downarrow$ & \textbf{HR@1}$\uparrow$ & \textbf{Acc@1}$\uparrow$ & \textbf{Kendall's $\tau$}$\uparrow$ & \textbf{Hausdorff}$\downarrow$ \\
  \midrule
  \multirow{3}{*}{\textbf{Beijing}} & Traffic volume & \textbf{0.9303} & \textbf{3.6703} & \textbf{0.9611} & \textbf{0.1076} & \underline{0.6814} & \textbf{0.4395} \\
  & POI density & 0.8511 & 3.8583 & \underline{0.9562} & 0.1029 & 0.6799 & 0.4597 \\
  & Traffic-POI rank fusion & \underline{0.9015} & \underline{3.7990} & 0.9557 & \underline{0.1044} & \textbf{0.6820} & \underline{0.4518} \\
  \cmidrule(lr){1-8}
  \multirow{3}{*}{\textbf{Chengdu}} & Traffic volume & \textbf{0.8574} & \textbf{1.3681} & \textbf{0.9402} & \underline{0.4296} & \textbf{0.7619} & \textbf{0.1102} \\
  & POI density & 0.8049 & 1.3856 & 0.9080 & 0.4272 & \underline{0.7611} & 0.1145 \\
  & Traffic-POI rank fusion & \underline{0.8103} & \underline{1.3788} & \underline{0.9227} & \textbf{0.4301} & 0.7607 & \underline{0.1132} \\
  \cmidrule(lr){1-8}
  \multirow{3}{*}{\textbf{Xi'an}} & Traffic volume & \textbf{0.8456} & \underline{2.1498} & \textbf{0.9540} & \textbf{0.3631} & \textbf{0.7129} & \textbf{0.2093} \\
  & POI density & 0.7786 & 2.1611 & \underline{0.9386} & 0.3585 & 0.7080 & 0.2253 \\
  & Traffic-POI rank fusion & \underline{0.7947} & \textbf{2.1479} & 0.9367 & \underline{0.3608} & \underline{0.7113} & \underline{0.2130} \\
  \cmidrule(lr){1-8}
  \multirow{3}{*}{\textbf{Porto}} & Traffic volume & \textbf{0.8073} & \textbf{1.3519} & \textbf{0.9746} & \textbf{0.1734} & \textbf{0.8267} & \textbf{0.4592} \\
  & POI density & \underline{0.7564} & 1.4281 & 0.9246 & \underline{0.1721} & 0.8078 & 0.4859 \\
  & Traffic-POI rank fusion & 0.7467 & \underline{1.4001} & \underline{0.9429} & 0.1697 & \underline{0.8150} & \underline{0.4727} \\
  \bottomrule
\end{tabular}
\end{table}

As shown in \cref{tab:critical_segment_selection_criterion}, the \emph{Traffic volume} criterion achieves the best overall performance on most metrics, outperforming the \emph{POI density} and \emph{Traffic-POI rank fusion} criteria. This validates our choice of using traffic volume to identify \emph{critical segments}. The reason is that CORE learns trajectory representations, and high-volume segments are precisely where route-choice behaviors are most frequently observed. In contrast, POI density captures environmental richness but is not necessarily aligned with actual mobility demand; for example, commercial districts may contain POI-rich internal roads that are rarely traversed by vehicle trajectories, making detailed LLM descriptions of these segments less beneficial for representation learning.

\end{document}